\documentclass{article}

     \PassOptionsToPackage{numbers, compress, sort}{natbib}
\usepackage[preprint]{neurips_2026}

\usepackage[utf8]{inputenc} % allow utf-8 input
\usepackage[T1]{fontenc}    % use 8-bit T1 fonts
\usepackage{hyperref}       % hyperlinks
\usepackage{url}            % simple URL typesetting
\usepackage{booktabs}       % professional-quality tables
\usepackage{amsfonts}       % blackboard math symbols
\usepackage{nicefrac}       % compact symbols for 1/2, etc.
\usepackage{microtype}      % microtypography
\usepackage[table]{xcolor}  % colors

\usepackage{amsmath}
\usepackage{amssymb}
\usepackage{mathtools}
\usepackage{amsthm}

\usepackage{lipsum}
\usepackage{enumitem}
\usepackage[most]{tcolorbox}

\usepackage{subcaption}
\usepackage{booktabs} 
\usepackage{multirow}
\usepackage{makecell}

\usepackage{fontawesome5}
\definecolor{softpink}{HTML}{F0529C}

\hypersetup{
    colorlinks=true,
    urlcolor=softpink,
    linkcolor=softpink,
    citecolor=softpink
}

\usepackage{float}
\usepackage[capitalize,noabbrev]{cleveref}

\definecolor{bestblue}{RGB}{209, 237, 242}

\definecolor{BlueBox}{HTML}{4dbbd5}
\definecolor{RedBox}{HTML}{e64b35}
\definecolor{GreenBox}{HTML}{00a087}
\definecolor{OrBox}{HTML}{FF8C00}
\definecolor{Bl2Box}{HTML}{3C5488}
\definecolor{BrBox}{HTML}{B09C85}

\definecolor{forestgreen}{HTML}{009B55}
\definecolor{methodblue}{HTML}{2F6F9F}

\colorlet{BlueBoxBg}{BlueBox!6}
\colorlet{RedBoxBg}{RedBox!6}
\colorlet{GreenBoxBg}{GreenBox!6}
\colorlet{OrBoxBg}{OrBox!6}
\colorlet{Bl2BoxBg}{Bl2Box!6}
\colorlet{BrBoxBg}{BrBox!6}

\usepackage[textsize=tiny]{todonotes}

\usepackage{tcolorbox}
\usepackage{enumitem}
\usepackage{pgfplots}
\pgfplotsset{compat=1.18}

\title{LLMs can construct powerful representations and streamline sample-efficient supervised learning}

\author{%
  Ilker Demirel \\
  MIT
  \And
  Lawrence Shi \\
  MIT 
  \And
  Zeshan Hussain \\
  MIT, Harvard Medical School \\
  \And
  David Sontag \\
  MIT 
}
\begin{document}
\maketitle
\vspace{-1.5em}
\begin{center}
\href{https://github.com/demireal/LRRL}{\texttt{\faGithub\ Code}}
\quad
\href{https://lrrlpaper.github.io}{\texttt{\faGlobe\ Website}}
\end{center}
\vspace{0.5em}
\begin{abstract}
    As real-world datasets become more complex and heterogeneous, supervised learning is often bottlenecked by input representation design. Modeling multimodal data, such as time-series, free text, and structured records, often requires non-trivial domain expertise. We propose an agentic pipeline to streamline this process. First, an LLM analyzes a small but diverse subset of {\em text-serialized} input examples {\em in-context} to synthesize a {\em global rubric}, which acts as a programmatic specification for extracting and organizing evidence. This rubric is then used to transform {\em naive} text-serializations of inputs into a more standardized format for downstream models. We also describe {\em local rubrics}, which are task-conditioned interpretive summaries generated by an LLM. Across 15 clinical tasks from the EHRSHOT benchmark, our rubric approaches significantly outperform count-feature models, naive LLM baselines, and a clinical foundation model pretrained on orders of magnitude more data. Beyond performance, rubrics offer operational advantages such as being easy to audit, cost-effectiveness at scale, and facilitating tabular representations.
\end{abstract}
\begin{figure}[ht]
    \centering
    \includegraphics[width=.84\linewidth]{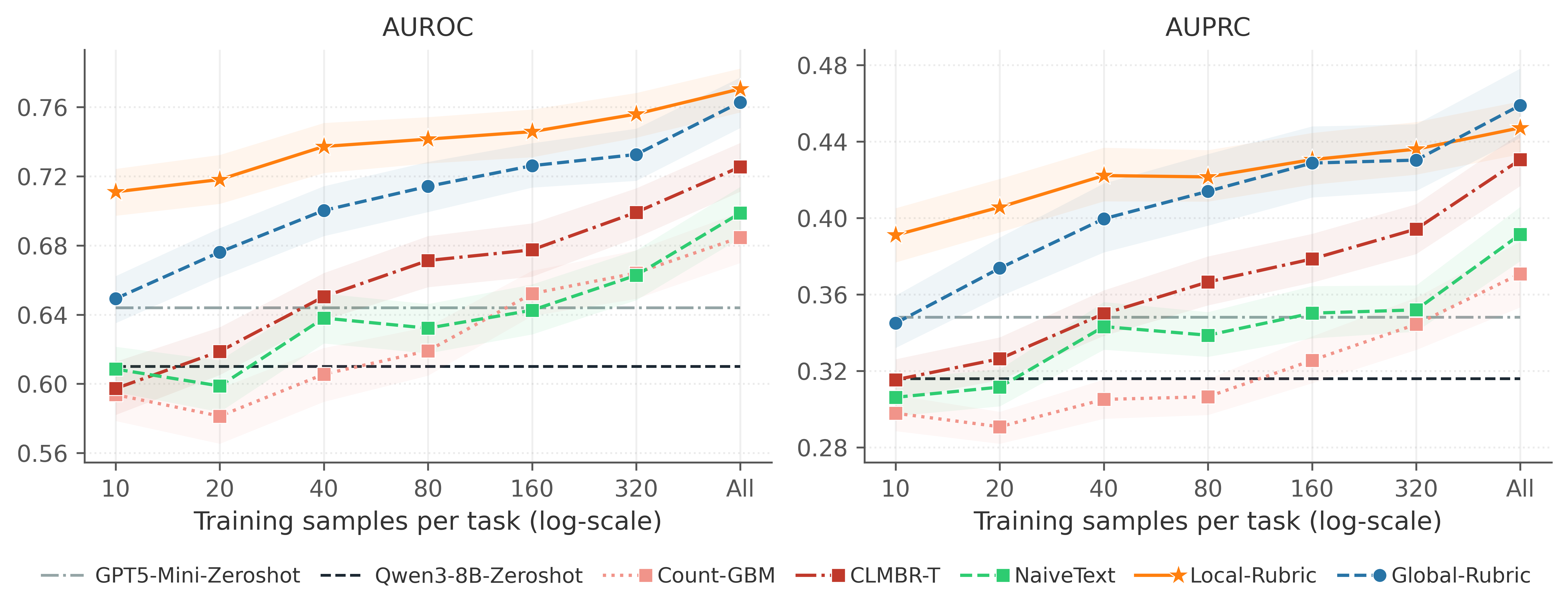}
    \caption{Performance (macro-averaged) over 15 clinical prediction tasks in the EHRSHOT benchmark \citep{wornow2023ehrshot}, swept over training-set size $n$ (per task). Our rubric-style representations constructed by LLMs outperform the {\em naive} text-serialization baseline in \citet{hegselmann2025large}, as well as a clinical foundation model pretrained on 2.57M patient timelines (CLMBR-T, \citep{wornow2023ehrshot}), and a count feature-based gradient boosting machine (Count-GBM, \citep{ke2017lightgbm, wornow2023ehrshot}).}
    \label{fig:fig1}
    \vspace{-10pt}
\end{figure}
\section{Introduction} \label{ssec:cont}
Supervised learning underpins a wide range of applications across domains. In medicine, deep neural networks achieve specialist-level performance in pneumonia detection and diabetic retinopathy screening \citep{rajpurkar2017chexnet, gulshan2016development}. In finance, credit risk assessment models outperform legacy scorecards \citep{lessmann2015benchmarking}. In environmental science, supervised learning enables weather forecasting from radar observations \citep{ravuri2021skilful}.

The success of supervised learning depends on the availability of input representations that can be easily processed by off-the-shelf models. Real-world datasets, however, are increasingly more complex and heterogeneous. They combine structured fields with unstructured text, time-series, and images. In healthcare, clinical prediction may benefit from longitudinal labs, coded events ({\em e.g.}, diagnoses), free-text notes, and medical images. In finance, stock-price forecasting and risk modeling may involve price and volume time-series, news reports, and structured events such as analyst ratings.

In domains with complex data, representation design requires bespoke engineering and domain expertise. Even then, resulting representations are not necessarily optimal: they may discard critical signal or bury it in noise. We show how large language models (LLMs) can build {\em agentic} pipelines that automate designing powerful input representations and enable sample-efficient supervised learning.

LLMs offer a practical interface to heterogeneous data through text-serializations. \citet{song2024omnipred} and \citet{akhauri2025performance} serialize diverse configurations and logs into text sequences to predict system performance metrics. \citet{hegselmann2025large} serialize longitudinal electronic health records (EHR) into Markdown and train linear heads over text-embeddings for clinical prediction tasks (see \Cref{fig:serialization_vs_rubric}, {\em left}). \citet{demirel2025using} use LLMs to predict daily user activities by combining multimodal time-series data from wearables after transforming them into textual descriptions. 

These works show LLMs' potential to streamline supervised learning with complex datasets, but they treat text-serialization of the input as fixed and leave the bulk of {\em calibration} to data for the downstream model. In contrast, we take the text-serialized input as a starting point and show how LLMs can automate constructing better representations to directly improve downstream performance.
\definecolor{BlueBox}{HTML}{4dbbd5}
\definecolor{RedBox}{HTML}{e64b35}
\definecolor{GreenBox}{HTML}{00a087}

% \definecolor{BlueBox}{HTML}{2874A6}
% \definecolor{RedBox}{HTML}{EC407A}
% \definecolor{GreenBox}{HTML}{27AE60}

\colorlet{BlueBoxBg}{BlueBox!6}
\colorlet{RedBoxBg}{RedBox!6}
\colorlet{GreenBoxBg}{GreenBox!6}

\begin{figure*}[t]
    \centering

    \newlength{\ExampleBoxHeight}
    \setlength{\ExampleBoxHeight}{0.255\textheight}

    \setlength{\tabcolsep}{2pt}

    \begin{tabular}{@{}p{0.32\textwidth}@{\hspace{0.4em}} p{0.32\textwidth}@{\hspace{0.4em}} p{0.32\textwidth}@{}}

    \begin{tcolorbox}[
        colback=RedBoxBg,
        colframe=RedBox,
        coltext=black,
        boxrule=1pt,
        arc=5pt,
        width=\linewidth,
        top=8pt, bottom=8pt, left=8pt, right=8pt
    ]
    \scriptsize\setlength{\parskip}{0.3em}
    \begin{minipage}[t][\ExampleBoxHeight][t]{\linewidth}
        \vspace{-1em}
        \textbf{Naive Text Representation}
        \vspace{0.4em}
        \hrule
        \vspace{0.6em}

        \textbf{\# Patient Demographics}

        \vspace{0.4em}
        
        - Patient age: 78, Female {[...]} 

        \vspace{0.4em}
        
        \textbf{\# Detailed Past Medical Visits}

        \vspace{0.4em}

        \textbf{\#\# Inpatient Visit (14 days to pred. time, current visit)}

        \vspace{0.4em}

        \textnormal{\#\#\# Conditions}\\
        - Acute posthemorrhagic anemia\\
        - pH meas., venous: 7.25, 7.31 {[...]}

        \vspace{0.4em}

        \textnormal{\#\#\# Medications}\\
        - furosemide 20 MG Oral Tablet {[...]}

        \vspace{0.4em}

        \textnormal{\#\#\# Procedures}\\
        - Chest x-ray\\
        - Electrocardiogram report {[...]}

        \vspace{0.5em}

        \textbf{\#\# Emergency Room Visit (87 days before prediction time)} 
        
        \vspace{0.4em}
        
        \textnormal{\#\#\# Conditions}\\
        - Benign essential hypertension \\
        - Chest pain {[...]}
    \end{minipage}
    \end{tcolorbox}

    &
    % ---------------- Middle: Summary representation ----------------
    \begin{tcolorbox}[
        colback=BlueBoxBg,
        colframe=BlueBox,
        coltext=black,
        boxrule=1pt,
        arc=5pt,
        width=\linewidth,
        top=8pt, bottom=8pt, left=8pt, right=8pt
    ]
    \scriptsize\setlength{\parskip}{0.3em}
    \begin{minipage}[t][\ExampleBoxHeight][t]{\linewidth}
        \vspace{-1em}
        \textbf{Local Rubric Representation}
        \vspace{0.4em}
        \hrule
        \vspace{0.6em}

        \textbf{1. Patient Snapshot} 
        \vspace{0.3em}
        
        - 27 yo hispanic male. \\
        - Recurrent cardiology visits for congenital anomaly of coronary artery {[...]} 

        \vspace{0.6em}
        
        \textbf{2. Main Risk Factors} 
        \vspace{0.3em}
        
        - Congenital coronary artery anomaly (established structural predisposition to myocardial ischemia/infarction). \\ 
        - Tobacco exposure (smokeless tobacco reported) {[...]} 
        
        \vspace{0.6em}
        
        \textbf{3. Protective Factors} 
        \vspace{0.3em}
        
        - Young age (27) — lower baseline atherosclerotic burden relative to older adults. \\
        - Normal BMI (21-22). \\ 
        - No documented diabetes (glucose in normal range) or chronic renal impairment {[...]}

    \end{minipage}
    \end{tcolorbox}

    &
    % ---------------- RIGHT: Rubricified representation ----------------
    \begin{tcolorbox}[
        colback=GreenBoxBg,
        colframe=GreenBox,
        coltext=black,
        boxrule=1pt,
        arc=5pt,
        width=\linewidth,
        top=8pt, bottom=8pt, left=8pt, right=8pt
    ]
    \scriptsize\setlength{\parskip}{0.3em}
    \begin{minipage}[t][\ExampleBoxHeight][t]{\linewidth}
        \vspace{-1em}
        \textbf{Global Rubric Representation}
        \vspace{0.4em}
        \hrule
        \vspace{0.6em}
        
        \textbf{3. Demographics} 

        \vspace{0.5em}
        
        - 55 | Female | {[...]}
        
        \vspace{0.5em}
        
        \textbf{6. Recent Cardiac Symptoms (last 365 days)} 

        \vspace{0.4em}
        
        - Chest pain/angina: No  \\
        - Dyspnea/shortness of breath: Yes (date unknown)  {[...]}
        
         \vspace{0.5em}
        
        \textbf{12. Other Relevant Labs} 

        \vspace{0.4em}
        
        - Creatinine: 1.12 (2023-12-02)  \\
        - eGFR: No data  {[...]}
        
         \vspace{0.5em}
        
        \textbf{17. Known Risk Factors} 

        \vspace{0.4em}
        
        - Diabetes: No (A1c date unknown)  \\
        - Family history of premature CAD: Unknown [...]
        
         \vspace{0.5em}
        
        \textbf{20. Non-cardiac Serious Illness That May Mimic or Alter MI Risk} 

        \vspace{0.4em}
        
        - Active malignancy: No {[...]}

    \end{minipage}
    \end{tcolorbox}

    \end{tabular}

    \caption{Synthetic electronic health record representation examples. {\em Left.} Naive text-serialization adopted from \citet{hegselmann2025large}. {\em Middle.} Local rubric representation (task-conditioned summary of the naive text-serialization). {\em Right.} Global rubric transformed version of the naive-text serialization.}
    \label{fig:serialization_vs_rubric}
    \vspace{-13pt}
\end{figure*}

Beyond modeling convenience, LLMs' pretraining knowledge can enable effective regularization, which is key to sample-efficiency. Our work aligns with literature on injecting knowledge into statistical models: LMPriors uses language descriptions as task-specific priors \citep{choi2022lmpriors}, and TabLLM shows effective few-shot tabular learning \citep{hegselmann2023tabllm}. LLM-Select and LLM-Lasso guide feature selection and regularization \citep{jeong2025llmselect, zhang2025llm}, and \citet{kim25knowledge} use task metadata to construct inductive biases.

These methods mostly use LLMs to augment traditional models working on clean datasets. In contrast, we focus on representation design: how complex inputs should be organized \emph{prior} to downstream learning. In that sense, our work also aligns with recent literature on learning in the language space, such as GEPA by \citet{agrawal2025gepa}. An extended related work section is included in Appendix~\ref{app:related_work}.

\textbf{Our contribution.}\hspace{11pt}We propose {\em rubrics}, which are used to process complex text inputs into a standardized and information-rich format that can be easily and efficiently digested by downstream learners. We assume {\em naive} text-serializations of inputs are available, or they can be constructed straightforwardly (see \Cref{fig:serialization_vs_rubric}, left for an example). We develop two types of rubrics, which are given below and detailed in \Cref{sec:rubric}.

\textbf{Global rubrics.}\hspace{11pt}A global rubric is a task-level specification that defines what information should be extracted from the input and how. It is {\em generated} by prompting an LLM with a diverse set of examples (see Figure~\ref{fig:rubric_creation_pipeline}, Panels A--C). 

\textbf{Local rubrics.}\hspace{11pt}We ask an LLM to produce a {\em task-conditioned} interpretive summary with structured sections (see \Cref{fig:sum_rub_creation_prompts}, left), similar to recent work on explainable clinical prediction models \citep{petridis2026holistic}.

\textbf{Advantages of global rubrics.}\hspace{11pt}Both rubrics achieve similar downstream performance and outperform the baselines. However, local rubrics do not have the same level of standardization as global rubrics, which endows the latter with several practical desiderata lacking in the former. 
\begin{itemize}[leftmargin=1.5em]
\setlength{\itemsep}{1pt} 
    \item \textit{Auditable and improvable:} Global rubrics are more amenable to inspections by domain experts, such as for analyzing subgroup bias risk and iterative refinement.
    \item \textit{More operationally useful:} Global rubric representations can be transformed into tabular features (\Cref{fig:rubric_creation_pipeline}, Panel F), immediately unlocking a suite of machine learning techniques.
    \item \textit{Cheaper to deploy:} Global rubric transformation at inference time can be automated (see \Cref{fig:rubric_creation_pipeline}, Panels E and F), whereas summarization requires an LLM forward pass per example. This makes global rubrics ``free'' compared to local rubrics, which incur ${\cal O}(N)$ time and compute cost.
\end{itemize}
We evaluate on 15 binary clinical prediction tasks from the EHRSHOT benchmark \citep{wornow2023ehrshot}, spanning operational outcomes, new diagnoses, lab results, and chest X-ray findings. We compare against a gradient boosting machine with count-based features (Count-GBM, \citep{ke2017lightgbm}), a clinical foundation model pretrained on 2.57M patients (CLMBR-T, \citep{wornow2023ehrshot}), zero-shot chain-of-thought prompting (CoT) with Qwen3-8B and GPT5-Mini\footnote{We used GPT5-Mini and GPT-5.2 via the HIPAA-compliant Microsoft Azure OpenAI Service.} \citep{wei2022chain, qwen3technicalreport, openai2025gpt5_system_card}, and the LLM baseline of \citet{hegselmann2025large}, which uses the naive EHR text-serializations we build on top of. 

We preview the main findings here and provide detailed discussion in \Cref{sec:quant}. Rubrics substantially outperform baselines on average across sample sizes $n$, with gaps largest for small $n$. First, the LLM's {\em interpretation} of evidence acts as a sample-efficiency lever. Stripping it from local rubrics costs noticeable performance at small $n$ and almost none at $n{=}\text{All}$, indicating that pretrained world knowledge supplies a prior the downstream classifier increasingly does without as data accumulate. Second, {\em standardized} global rubric templates are themselves strong representations even without an interpretive layer, beating all baselines across all sample sizes. Global rubrics trail local rubrics at small $n$, where the LLM-injected statistical prior in the language space matters most, but the gap closes by $n{=}\text{All}$ as labels accumulate.
\section{Rubric representation learning with LLMs} \label{sec:rubric}
\subsection{Global rubrics} \label{sec:global_rubs}
We introduce global rubrics for converting heterogeneous, weakly structured inputs into standardized, task-aligned representations. While we focus on electronic health records (EHR), the procedure applies wherever inputs can be text-serialized. Throughout, we use GPT5-Mini for natural-language steps (rubric synthesis and application, local-rubric generation) and GPT-5.2 for code-generation, as it produced more reliable scripts in our pilots (\Cref{fig:rubric_creation_pipeline}, Panels E, F).
\paragraph{Setup and notation.}
We describe the global rubric learning procedure for a single prediction task. Let $\mathcal{D}=\{(x_i,y_i)\}_{i=1}^n$ denote labeled training data, where $x$ is a \emph{raw input} and $y\in\{0,1\}$ is the task label. Let $s(\cdot)$ be some serialization procedure that maps an input $x$ into its textual representation, and define $x^{\text{text}} = s(x)$. A rubric specifies a task-specific transformation
\[
\mathcal{R}:\; x^{\text{text}} \mapsto x^{\text{rubric}},
\]
where $x^{\text{rubric}}$ is a more structured representation of the same underlying input $x$, and it can be used with downstream predictors instead of $x^{\text{text}}$. We describe downstream training in \Cref{sec:downstream_training}.
\definecolor{ColA}{HTML}{D44020}
\definecolor{ColB}{HTML}{5B2D8E}
\definecolor{ColC}{HTML}{C8960A}
\definecolor{ColD}{HTML}{2874A6}
\definecolor{ColE}{HTML}{EC407A}
\definecolor{ColF}{HTML}{27AE60}

\colorlet{ColABg}{ColA!6}
\colorlet{ColBBg}{ColB!6}
\colorlet{ColCBg}{ColC!6}
\colorlet{ColDBg}{ColD!6}
\colorlet{ColEBg}{ColE!6}
\colorlet{ColFBg}{ColF!6}

\definecolor{InnerGray}{HTML}{9AA8B5}
\definecolor{InnerGrayBg}{HTML}{F7F8F9}

\begin{figure*}[t]
    \centering

    \newlength{\PanelHeight}
    \setlength{\PanelHeight}{0.24\textheight}

    \begin{tabular}{
        @{}p{0.32\textwidth}
        @{\hspace{0.4em}}
        p{0.31\textwidth}
        @{\hspace{0.4em}}
        p{0.32\textwidth}@{}
    }

    %% ── Panel A ──────────────────────────────────────────────────────────
    \begin{tcolorbox}[
        colback=ColABg,
        colframe=ColA,
        colbacktitle=ColABg,
        coltitle=black,
        coltext=black,
        boxrule=1.15pt,
        arc=6pt,
        width=\linewidth,
        top=4pt, bottom=4pt, left=6pt, right=6pt,
        title=\textbf{(A) Diverse Cohort Selection},
        fonttitle=\bfseries\small
    ]
    \begin{minipage}[t][0.81\PanelHeight][t]{\linewidth}

    {\raggedright\bfseries\scriptsize
    \# Label stratified $k$-means in
    text-embedding (\textcolor{red}{$x_{\text{text}}$}) space
    \par}

    \vspace{0.5em}
    \hrule
    \vspace{0.3em}

    \centering

    \begin{tikzpicture}[scale=1.12, x=1.0cm, y=1.0cm]
        \pgfmathsetseed{11}

        \begin{scope}
            \clip (-2.3,-1.78) rectangle (2.00,1.87);

            \newcommand{\DrawCluster}[6]{%
                \def\cx{#1}\def\cy{#2}%
                \draw[#3!80, dotted, thick, opacity=0.55] (\cx,\cy) circle (0.62);
                \foreach \j in {1,...,30} {
                    \pgfmathsetmacro{\dx}{(rand-0.5)*0.78}
                    \pgfmathsetmacro{\dy}{(rand-0.5)*0.78}
                    \fill[#3!75, opacity=0.38] (\cx+0.58*\dx,\cy+0.58*\dy) circle (2.05pt);
                }
                \foreach \j in {1,...,3} {
                    \pgfmathsetmacro{\dx}{(rand-0.5)*0.55}
                    \pgfmathsetmacro{\dy}{(rand-0.5)*0.55}
                }
                \fill[#3!85!black] (\cx-0.18,\cy-0.18) rectangle ++(0.16,0.16);
                \draw[white, very thin] (\cx-0.18,\cy-0.18) rectangle ++(0.16,0.16);
            }

            \DrawCluster{-1}{1.2}{red}{blue}{0}{0}
            \DrawCluster{-1.5}{0.3}{blue}{red}{0}{0}
            \DrawCluster{0.6}{1.25}{blue}{red}{0}{0}
            \DrawCluster{0.3}{-0.1}{red}{blue}{0}{0}

        \end{scope}

        \node[anchor=west, font=\scriptsize] at (-2.35,2.4) {%
            \raisebox{0.2ex}{\tikz{\fill[blue!85!black] (0,0) rectangle (0.18,0.18); \draw[white,thick] (0,0) rectangle (0.18,0.18);}}~
            Y~=~0 medoid\qquad
            \raisebox{0.2ex}{\tikz{\fill[red!85!black] (0,0) rectangle (0.18,0.18); \draw[white,thick] (0,0) rectangle (0.18,0.18);}}~
            Y~=~1 medoid
        };
        \node[anchor=west, font=\scriptsize] at (-2.33,2.1) {%
            \raisebox{0.2ex}{\tikz{\fill[blue!75] (0.09,0.09) circle (0.09);}}~
            Y~=~0 patient\qquad\hspace{0.1em}
            \raisebox{0.2ex}{\tikz{\fill[red!75] (0.09,0.09) circle (0.09);}}~
            Y~=~1 patient
        };
    \end{tikzpicture}

    \vfill
    \end{minipage}
    \end{tcolorbox}

    &

    %% ── Panel B ──────────────────────────────────────────────────────────
    \begin{tcolorbox}[
        colback=ColBBg,
        colframe=ColB,
        colbacktitle=ColBBg,
        coltitle=black,
        coltext=black,
        boxrule=1.15pt,
        arc=6pt,
        width=\linewidth,
        top=4pt, bottom=4pt, left=6pt, right=6pt,
        title={\textbf{(B) Rubric Synthesis}},
        fonttitle=\bfseries\small
    ]
    \scriptsize\setlength{\parskip}{0.16em}
    \begin{minipage}[t][0.81\PanelHeight][t]{\linewidth}

    {\raggedright\bfseries\scriptsize
    \# Ask an LLM to synthesize a task-specific rubric.
    \par}

    \vspace{0.5em}
    \hrule
    \vspace{0.5em}

    Create a rubric for predicting hypertension risk in the next year by analyzing data from 40 patients.

    \vspace{0.5em}

    \textbf{List of EHRs (Medoids, \textcolor{blue}{$x_{\text{text}}$}):}

    \begin{tcolorbox}[
        colback=InnerGrayBg,
        colframe=InnerGray,
        coltext=black,
        boxrule=1.15pt,
        arc=3pt,
        left=4pt, right=4pt, top=3pt, bottom=3pt
    ]
    \hspace{0.1em}Pt$_1$: [78yo, F, HTN meds, SBP=148...]\\
    \hspace{0.1em}...\\
    \hspace{0.1em}Pt$_{40}$: [27yo, M, family hx...]
    \end{tcolorbox}

    Output a rubric.\\
    \hspace{0.5em}$\bullet$ Be structured and consistent \\
    \hspace{0.5em}$\bullet$ Extract facts only {[...]}\\

    \vfill

    \end{minipage}
    \end{tcolorbox}

    &

    %% ── Panel C ──────────────────────────────────────────────────────────
    \begin{tcolorbox}[
        colback=ColCBg,
        colframe=ColC,
        colbacktitle=ColCBg,
        coltitle=black,
        coltext=black,
        boxrule=1.15pt,
        arc=6pt,
        width=\linewidth,
        top=4pt, bottom=4pt, left=6pt, right=6pt,
        title={\textbf{(C) Task-specific Rubric} $\mathcal{R}$},
        fonttitle=\bfseries\small
    ]
    \scriptsize\setlength{\parskip}{0.15em}
    \begin{minipage}[t][.81\PanelHeight][t]{\linewidth}

    {\raggedright\bfseries\scriptsize
    \# LLM-derived rubric ${\cal R}$ for transforming \textcolor{red}{$x_{\text{text}}$} to \textcolor{red}{$x_{\text{rubric}}$}
    \par}

    \vspace{0.5em}
    \hrule
    \vspace{0.5em}

    \vspace{0.20em}

    {\color{black}\textbf{\S1.\ \ Demographics}}\\
    \hspace{0.5em}$\vdash$ Age, sex, BMI

    \vspace{0.18em}

    {\color{black}\textbf{\S2.\ \ CV Risk Factors}}\\
    \hspace{0.5em}$\vdash$ BP readings (SBP/DBP)\\
    \hspace{0.5em}$\vdash$ HTN medications

    \vspace{0.18em}

    {\color{black}\textbf{\S3.\ \ Comorbidities}}\\
    \hspace{0.5em}$\vdash$ Diabetes, CKD status

    \vspace{0.18em}

    {\color{black}\textbf{\S4.\ \ Temporal Trends}}\\
    \hspace{0.5em}$\vdash$ BP trajectory (6-12mo)\\
    \hspace{0.5em}$\vdash$ Weight changes

    \vfill
    \begin{center}
    \begin{tcolorbox}[
        colback=InnerGrayBg,
        colframe=InnerGray,
        boxrule=1.3pt,
        arc=6pt,
        left=10pt, right=10pt, top=2pt, bottom=2pt,
        width=0.95\linewidth
    ]
    \centering\bfseries\small
    \textcolor{black}{${\cal R}: x_{\text{text}} \;\rightarrow\; x_{\text{rubric}}$}
    \end{tcolorbox}
    \end{center}

    \end{minipage}
    \end{tcolorbox}

    \\[0.1em]

    %% ── Panel D ──────────────────────────────────────────────────────────
    \begin{tcolorbox}[
        colback=ColDBg,
        colframe=ColD,
        colbacktitle=ColDBg,
        coltitle=black,
        coltext=black,
        boxrule=1.15pt,
        arc=6pt,
        width=\linewidth,
        top=4pt, bottom=4pt, left=6pt, right=6pt,
        title=\textbf{(D) Rubric Appl. via LLMs},
        fonttitle=\bfseries\small
    ]
    \scriptsize\setlength{\parskip}{0.15em}
    \begin{minipage}[t][0.79\PanelHeight][t]{\linewidth}

    {\raggedright\scriptsize
    \# \textbf{Global-Rubric method} \\
    Ask an LLM to apply the rubric transformation ${\cal R}$ to each input.
    \par}

    \vspace{0.5em}
    \hrule
    \vspace{0.5em}

    \#\# Rubric ${\cal R}$:\\ \textcolor{red}{\{rubric\_instructions\}}

    \vspace{0.5em}

    \#\# Patient EHR:\\ \textcolor{red}{\{ehr\_text ($x_{\textnormal{text}}$)\}}

    \vspace{0.5em}

    Fill in every field of the rubric template above using ONLY information from this patient's EHR. Rules:

    \vspace{0.5em}

    $\bullet$ Follow the exact field order and section structure of the rubric. \\
    \hspace{0.5em}$\bullet$ If data for a field is not present, write "No data". {[...]}
    \end{minipage}
    \end{tcolorbox}

    &

    %% ── Panel E ──────────────────────────────────────────────────────────
    \begin{tcolorbox}[
        colback=ColEBg,
        colframe=ColE,
        colbacktitle=ColEBg,
        coltitle=black,
        coltext=black,
        boxrule=1.15pt,
        arc=6pt,
        width=\linewidth,
        top=4pt, bottom=4pt, left=6pt, right=6pt,
        title={\textbf{(E) Rubric Appl. via Parser}},
        fonttitle=\bfseries\small
    ]
    \scriptsize\setlength{\parskip}{0.16em}
    \begin{minipage}[t][0.79\PanelHeight][t]{\linewidth}

    {\raggedright\scriptsize
    \# \textbf{Global-Rubric-Auto method} \\
    Ask an LLM to generate a parser script to apply the learned rubric transformation ${\cal R}$ to each input.
    \par}

    \vspace{0.5em}
    \hrule
    \vspace{0.5em}

    Write a script that reads patient EHR serializations and fills in a  rubric template using deterministic string/regex parsing only {[...]}

    \vspace{0.5em}

    \#\# Rubric ${\cal R}$: \textcolor{blue}{\{rubric\_instructions\}}

    \vspace{0.5em}

    Example EHR text serializations: \\\textcolor{blue}{\{List of medoid pairs: ($x_{\textnormal{text}}, x_{\textnormal{rubric}})$\}}

    \vspace{0.5em}

    $\bullet$ Use only Python standard libraries such as 're', 'json' {[...]} \\
    \hspace{0.5em}$\bullet$ No LLM API calls {[...]}
    \vfill

    \end{minipage}
    \end{tcolorbox}

    &

    %% ── Panel F ──────────────────────────────────────────────────────────
    \begin{tcolorbox}[
        colback=ColFBg,
        colframe=ColF,
        colbacktitle=ColFBg,
        coltitle=black,
        coltext=black,
        boxrule=1.15pt,
        arc=6pt,
        width=\linewidth,
        top=4pt, bottom=4pt, left=6pt, right=6pt,
        title={\textbf{(F) Rubric Tabularization}},
        fonttitle=\bfseries\small
    ]
    \scriptsize\setlength{\parskip}{0.15em}
    \begin{minipage}[t][0.79\PanelHeight][t]{\linewidth}

    {\raggedright\scriptsize
    \# \textbf{Global-Rubric-Tabular method} \\
    Ask an LLM to generate a script to transform $x_{\textnormal{rubric}}$ to tabular features based on ${\cal R}$.
    \par}

    \vspace{0.5em}
    \hrule
    \vspace{0.5em}

    Write a Python script to convert rubric-formatted patient EHRs into numeric feature vectors {[...]}

    \vspace{0.5em}

    Example rubric-transformed EHR text serializations: \\\textcolor{red}{\{Medoids in $x_{\textnormal{rubric}}$ format, obtained from $x_{\textnormal{text}}$ using parser in Panel (E)\}}

    \vspace{0.25em}

    $\bullet$ General: handle any value the rubric parser could plausibly produce {[...]} \\
    $\bullet$ Robust: gracefully handle missing values {[...]}
    \end{minipage}
    \end{tcolorbox}

    \end{tabular}

    \caption{
        Agentic global-rubric pipeline for EHRSHOT tasks. Full prompts are in Appendix~\ref{app:agent_prompts}.
    }
    \label{fig:rubric_creation_pipeline}
    \vspace{-8pt}
\end{figure*}
\paragraph{Global rubric synthesis.}
Global rubric learning has two stages, shown in Panels A and B of \Cref{fig:rubric_creation_pipeline}. First, we select a small, label-balanced and \emph{diverse} cohort from the training split. Second, an LLM inspects this cohort \emph{in-context} and synthesizes a task-specific rubric by defining predictive features and describing how to extract them. 
\begin{itemize}[leftmargin=1.5em]
    \item 
    \textit{Step 1) Diverse cohort selection:\hspace{11pt}}
    Rubric synthesis is done through a single prompt to an LLM (GPT5-Mini). We first embed each text-serialized input $x_i^{\text{text}}$ into a vector space using a pretrained text-embedding model \citep{qwen3embedding}, and stratify by label, 
    \[
    \mathcal{D}^+ = \{x_i^{\text{text}} : y_i = 1\}, \qquad
    \mathcal{D}^- = \{x_i^{\text{text}} : y_i = 0\}.
    \]
    We perform $k$-means clustering within each stratum, where $k$ is the number of clusters per label-stratum, so the final cohort contains $2k$ examples. Due to context size limitations, we use $k=20$. Finally, we take the cluster medoids to obtain a diverse cohort (see \Cref{fig:rubric_creation_pipeline}, Panel A).  

    \item
    \textit{Step 2) Rubric synthesis:\hspace{11pt}}
    Given the selected cohort, we ask an LLM (GPT5-Mini) to produce a task-specific rubric that (i) defines discriminative, task-relevant signals, and (ii) specifies how each signal should be extracted from a given input, $x^{\text{text}}$ (see \Cref{fig:rubric_creation_pipeline}, Panel B). The full prompt is provided in Appendix~\ref{app:rubric_prompt}, and two full-rubric examples can be found in Appendix~\ref{app:fullrubrics}.

\end{itemize}

\textbf{Global rubric application.} \hspace{11pt} A global rubric ${\cal R}$ is applied to naive text-serializations, $x^{\text{text}}$, to produce $x^{\text{rubric}}$. We propose four different methods at this stage. 
\begin{itemize}[leftmargin=1.5em]
    \item 
    \textcolor{methodblue}{\textbf{Global-Rubric}} \textit{(LLM application; \Cref{fig:rubric_creation_pipeline}, Panel D):\hspace{11pt}} We prompt an LLM (GPT5-Mini) with the global rubric ${\cal R}$, and the naive text-serialization of the input $x^{\text{text}}$, asking it to return $x^{\text{rubric}}$. 

    \item 
    \textcolor{methodblue}{\textbf{Global-Rubric-Auto}} \textit{(parser-script application; \Cref{fig:rubric_creation_pipeline}, Panel E):\hspace{11pt}} We prompt an LLM (GPT-5.2) with ${\cal R}$ and 40 paired $(x^{\text{text}}, x^{\text{rubric}})$ examples, asking it to write a {\em deterministic} parser script that converts $x^{\text{text}}$ to $x^{\text{rubric}}$. The script is then used at deployment time without LLM calls.

    \item 
    \textcolor{methodblue}{\textbf{Global-Rubric-Tabular}} \textit{(tabularization-script; \Cref{fig:rubric_creation_pipeline}, Panel F):\hspace{11pt}} We prompt an LLM (GPT-5.2) with the global rubric ${\cal R}$, parser script for the rubric transformation (see item above), and some examples (40) of {\em parser-generated} rubric-transformations, $x^{\text{rubric}}$. We ask the LLM to write a script to convert $x^{\text{rubric}}$ into a set of tabular features.
    \item \textcolor{methodblue}{\textbf{Global-Rubric-Blind}}: ${\cal R}$ is generated from the task description and the LLM's world knowledge alone (skipping Step 1 in rubric synthesis above), then applied like \textcolor{methodblue}{Global-Rubric}.
\end{itemize}

\subsection{Local rubrics}
\label{sec:local_rubrics}
Global rubrics define a structure shared across {\em all} inputs, once applied. Beyond performance gains, this unlocks several practical advantages, as summarized in \Cref{ssec:cont}. However, it is important to characterize the effect of standardizing the input on statistical performance.

To study this, we introduce \emph{local rubrics}: task-conditioned summaries of $x^{\text{text}}$, generated independently for each example by an LLM (GPT5-Mini). Unlike global rubrics, they impose only a general section-level structure, giving the LLM flexibility to extract and interpret the most relevant evidence per case. We define three variants that progressively ablate the LLM's interpretation of the evidence:
\begin{itemize}[leftmargin=1.5em]
    \item \textcolor{methodblue}{\textbf{Local-Rubric}}: a task-conditioned summary that includes both factual evidence and the LLM's reasoning over it (\Cref{fig:sum_rub_creation_prompts}, left).
    \item \textcolor{methodblue}{\textbf{Local-Rubric-NoInterp}}: starts from \textcolor{methodblue}{Local-Rubric} and strips out explicit interpretive or predictive language while preserving factual evidence and pointers to missing or otherwise notable unobserved information (\Cref{fig:sum_rub_creation_prompts}, middle).
    \item \textcolor{methodblue}{\textbf{Local-Rubric-Basic}}: asks the LLM to extract only task-relevant facts, without weighing evidence, reasoning, or assessing risk (\Cref{fig:sum_rub_creation_prompts}, right). It is the simplest of the three and may omit useful cues retained by \textcolor{methodblue}{Local-Rubric-NoInterp}, such as potential missingness flags.
\end{itemize}
\definecolor{ColA}{HTML}{FF7F0E}
\definecolor{ColB}{HTML}{D4A000}
\definecolor{ColC}{HTML}{8B4010}
\colorlet{ColABg}{ColA!6}
\colorlet{ColBBg}{ColB!6}
\colorlet{ColCBg}{ColC!6}

\begin{figure*}[t]
\centering

\begin{tabular}{
    @{}p{0.325\textwidth}
    @{\hspace{0.4em}}
    p{0.325\textwidth}@{}
    @{\hspace{0.4em}}
    p{0.325\textwidth}@{}
}

\begin{tcolorbox}[
    colback=ColABg,
    colframe=ColA,
    boxrule=1pt,
    arc=5pt,
    width=\linewidth,
    height=0.2\textheight,
    valign=top,
    top=3pt, bottom=3pt, left=5pt, right=5pt,
    boxsep=2pt,
    before skip=0pt,
    after skip=0pt
]
\scriptsize
\setlength{\parskip}{0pt}
\setlength{\baselineskip}{0.92\baselineskip}

\textbf{\# Local-Rubric prompt}\par
\vspace{0.3em}
\hrule
\vspace{0.4em}

Read the EHR and write a reasoning trace that characterizes the patient's risk profile for the task:
\{\textcolor{blue}{task\_query}\}

\vspace{0.5em}
--- START OF EHR DATA ---\par
\{\textcolor{blue}{NaiveText Serialization ($x^{\text{text}}$)}\}\par
--- END OF EHR DATA ---

\vspace{0.3em}
Your output must follow the exact section structure given below:\\ \\
   \hspace{11pt}1. Patient snapshot\\  \hspace{11pt} 2. Main risk factors\\ \hspace{11pt} 3. Protective factors\\ \hspace{11pt}  4. What's unknown \& can swing the risk\\ \hspace{11pt}  5. Weighing \& aggregating the evidence
\end{tcolorbox}
&
\begin{tcolorbox}[
    colback=ColBBg,
    colframe=ColB,
    boxrule=1pt,
    arc=5pt,
    width=\linewidth,
    height=0.2\textheight,
    valign=top,
    top=3pt, bottom=3pt, left=5pt, right=5pt,
    boxsep=2pt,
    before skip=0pt,
    after skip=0pt
]
\scriptsize
\setlength{\parskip}{0pt}
\setlength{\baselineskip}{0.92\baselineskip}

\textbf{\# Local-Rubric-NoInterp prompt}\par
\vspace{0.3em}
\hrule
\vspace{0.4em}
Below is a clinical summary of an EHR. Produce a minimally edited version that removes interpretive language.

\vspace{0.5em}
--- START OF EHR SUMMARY ---\par
\{\textcolor{blue}{Local-Rubric Representation ($x^{\textnormal{rubric}}$)}\} \par
--- END OF EHR SUMMARY ---

\vspace{0.5em}
Remove:

- Interpretive words ({\em e.g., 'suggests'}) \\
- Predictions, risk assessments {[...]}

\vspace{0.35em}

Keep unchanged: 

\vspace{0.35em}
- Facts: demographics, diagnoses, lab values with numbers, vitals, {[...]} \\
- Exact wording of factual statements {[...]}
\end{tcolorbox}
&
\begin{tcolorbox}[
    colback=ColCBg,
    colframe=ColC,
    boxrule=1pt,
    arc=5pt,
    width=\linewidth,
    height=0.2\textheight,
    valign=top,
    top=3pt, bottom=3pt, left=5pt, right=5pt,
    boxsep=2pt,
    before skip=0pt,
    after skip=0pt
]
\scriptsize
\setlength{\parskip}{0pt}
\setlength{\baselineskip}{0.92\baselineskip}

\textbf{\# Local-Rubric-Basic Prompt}\par
\vspace{0.3em}
\hrule
\vspace{0.4em}

Read the EHR below. Extract and summarize the evidence that is relevant to the task: \{\textcolor{blue}{task\_query}\}

\vspace{0.5em}
--- START OF EHR DATA ---\par
\{\textcolor{blue}{NaiveText Serialization ($x^{\text{text}}$)}\}\par
--- END OF EHR DATA ---

\vspace{0.5em}
- ONLY list factual evidence in the EHR. \\
- Do NOT interpret, weigh, or reason about the evidence. \\
- Do NOT assess risk, draw conclusions, or make predictions. \\
- Do NOT use language like "suggests", "indicates", "consistent with", "increases risk", or "protective". \\

\end{tcolorbox}

\end{tabular}

\caption{Prompts for generating local rubric representations. {\em Left.} Task-conditioned interpretive summaries (Local-Rubric). {\em Middle.} Edited summaries with interpretations stripped away (Local-Rubric-NoInterp, full prompt in Appendix~\ref{app:local-interp}). {\em Right.} Fact-listing summaries (Local-Rubric-Basic).}
\label{fig:sum_rub_creation_prompts}
\vspace{-10pt}
\end{figure*}
\section{EHRSHOT benchmark}
\label{sec:ehrshot}

We evaluate on EHRSHOT \citep{wornow2023ehrshot}, a longitudinal EHR benchmark which contains deidentified data from 6{,}739 patients at Stanford Medicine, including demographics, diagnoses, procedures, medications, and labs across full patient timelines with millions of coded events. 

\textbf{Clinical prediction tasks.}\hspace{11pt}
EHRSHOT comprises 15 binary classification tasks across 4 categories:
\begin{itemize}[leftmargin=1.5em]
    \setlength\itemsep{0pt}
    \item \textit{Operational outcomes}: ICU transfer, Long length of stay ($>7$ days), 30-day readmission
    \item \textit{Assignment of new diagnoses}: Acute myocardial infarction (MI), Celiac, Hyperlipidemia, Hypertension, Lupus, Pancreatic cancer
    \item \textit{Anticipating labs}: Anemia, Hyperkalemia, Hypoglycemia, Hyponatremia, Thrombocytopenia
    \item \textit{Chest X-ray}: Abnormal chest x-ray findings
\end{itemize}
Operational tasks predict near-term events in the context of the current visit. Diagnosis tasks predict new diagnoses within one year. Lab tasks predict abnormal upcoming results for the most recent lab order. The chest X-ray task predicts abnormal radiology findings.

\textbf{Subsampling.}\hspace{11pt}
Some methods use an LLM call per example, so we subsample some EHRSHOT tasks to keep training and evaluation budget within a reasonable boundary (full details provided in \Cref{app:ehrshot}). All methods, including the baselines, are trained and evaluated on the same subsampled splits. \Cref{app:grtab-full} reports \textcolor{methodblue}{Global-Rubric-Tabular} performance on the full EHRSHOT dataset, since it does not use LLM calls at deployment.

% \textbf{Evaluation.}\hspace{11pt}
% %
% Models are trained on the training split, with their hyperparameters tuned based on the validation split, and evaluated in test split. We report AUROC and AUPRC, which are more appropriate than threshold-dependent metrics such as F1-score \citep{van2025evaluation}, along with 95\% confidence intervals obtained by bootstrapping in the test split.
%
\section{Methods and downstream training} \label{sec:downstream_training}
\subsection{Our rubric representation-based methods}
In \Cref{sec:rubric}, we introduced six textual rubric variants---\textcolor{methodblue}{Global-Rubric}, \textcolor{methodblue}{Global-Rubric-Blind}, \textcolor{methodblue}{Global-Rubric-Auto}, \textcolor{methodblue}{Local-Rubric}, \textcolor{methodblue}{Local-Rubric-NoInterp}, and \textcolor{methodblue}{Local-Rubric-Basic}---and one tabular variant, \textcolor{methodblue}{Global-Rubric-Tabular}. For all textual variants, the rubric output $x^{\text{rubric}}$ (or $x^{\text{text}}$ for \textcolor{methodblue}{NaiveText}) is wrapped in a unified task-conditioned prompt template (\Cref{fig:task_prompt} in Appendix~\ref{app:task_prompt}) before being encoded into a numerical vector by a frozen pretrained text-embedding model, and an L2-regularized logistic regression classifier is fit on top of those embeddings. We report results with \texttt{Qwen3-Embedding-8B} \citep{qwen3embedding} as the default backbone in the main paper; \texttt{LLaMA-3-8B} \citep{llama3, llm2vec}, \texttt{Mistral-7B} \citep{mistral7b, llm2vec}, and OpenAI's \texttt{text-embedding-3-large} \citep{openai_te3l} are evaluated as additional backbones in \Cref{app:smaller_embedding}. For the tabular variant, the LLM-generated parser/tabularization scripts produce a feature vector that is fed into an XGBoost classifier \citep{chen2016xgboost}. Hyperparameter tuning details are provided in \Cref{app:downstream_hparams}.
\subsection{EHRSHOT baselines} \label{sec:baselines_structure}
We include \textcolor{methodblue}{Count-GBM} following EHRSHOT \citep{wornow2023ehrshot}, where each EHR is converted into a vector of code counts observed prior to the prediction time. Different time-windows are used, such as last 90 days and 90-180 days before. A LightGBM classifier is then trained \citep{ke2017lightgbm}.

We also evaluate \textcolor{methodblue}{CLMBR-T}, a transformer-based autoregressive medical foundation model. It is pretrained with a next-code prediction objective, using longitudinal data from 2.57M patients drawn from the same distribution as the EHRSHOT dataset \citep{wornow2023ehrshot}. For downstream tasks, a logistic regression classifier is fit on top of the vector embeddings extracted from CLMBR-T.

\subsection{LLM-based baselines} \label{sec:llm_baseline}

Our rubric representations, $x^{\text{rubric}}$, are derived from {\em naive} text-serializations of the input, $x^{\text{text}}$. We adopt the serialization introduced by \citet{hegselmann2025large} (see \Cref{fig:serialization_vs_rubric}, left). Each record includes patient demographics, a ``General Medical Events'' section for codes that are not tied to any visit, and a ``Detailed Past Medical Visits'' section listing visits in reverse chronological order. We refer to the baseline that uses $x^{\text{text}}$ directly as \textcolor{methodblue}{NaiveText}. As with the textual rubrics, $x^{\text{text}}$ is embedded using a pretrained embedding model, and a logistic regression classifier is trained on top.

We also evaluate zero-shot chain-of-thought (CoT) prompting \citep{wei2022chain}. For each example, \textcolor{methodblue}{Qwen3-8B} and \textcolor{methodblue}{GPT5-Mini} are prompted to reason over the \textcolor{methodblue}{NaiveText} EHR serialization and give a final \texttt{Yes}/\texttt{No} answer. We sample 10 responses and estimate the probability as the fraction of \texttt{Yes} answers.

\section{Quantitative results} \label{sec:quant}
\textbf{Rubrics outperform the baselines on average.}\hspace{11pt}
We evaluate the 15 EHRSHOT tasks across training-set sizes from $n{=}10$ per
class to $n{=}\text{All}$ (\Cref{fig:sweep_overall,fig:sweep_groups}, with
numerical results in \Cref{tab:embedding_8b} in Appendix~\ref{app:emb_table},
and per-task tables in Appendix~\ref{app:pertask}). We use $1{:}1$
label-balanced training sets except for $n{=}\text{All}$. \textcolor{methodblue}{Global-Rubric} and
\textcolor{methodblue}{Local-Rubric} outperform every baseline for all values of $n$, with the
largest gap at small $n$. At $n{=}10$, \textcolor{methodblue}{Global-Rubric} reaches $0.649$ AUROC
($0.345$ AUPRC) and \textcolor{methodblue}{Local-Rubric} $0.701$ ($0.382$), while \textcolor{methodblue}{NaiveText},
\textcolor{methodblue}{CLMBR-T}, and \textcolor{methodblue}{Count-GBM} all remain near $0.60$ ($0.30$). At
$n{=}\text{All}$, \textcolor{methodblue}{Global-Rubric} obtains $0.763$ ($0.459$) and \textcolor{methodblue}{Local-Rubric}
$0.772$ ($0.452$), with the strongest baseline, \textcolor{methodblue}{CLMBR-T}, trailing at
$0.725$ ($0.430$). On rare-event tasks, rubric methods roughly double the
baselines' AUPRC (e.g., new diagnoses at $n{=}10$).

\textbf{Sample-efficiency: gains are largest at small $n$.}\hspace{11pt}
Rubric methods outpace traditional baselines most strongly when labels are
scarce, and the gap narrows but does not close at $n{=}\text{All}$.
\textcolor{methodblue}{CLMBR-T}, in particular, trails our rubric methods by more than $0.10$ AUROC
and $\sim 0.07$ AUPRC at $n{=}10$. It is pretrained on a domain-specific corpus of $2.57$M
patient records drawn from the same distribution as EHRSHOT, a form of
prior knowledge no rubric method has access to. Rubrics nonetheless surpass it,
drawing on a complementary source of prior knowledge from the
general-purpose, web-scale pretraining of the LLMs used during rubric
synthesis and application.

\begin{figure}[t]
    \centering
    \includegraphics[width=\linewidth]{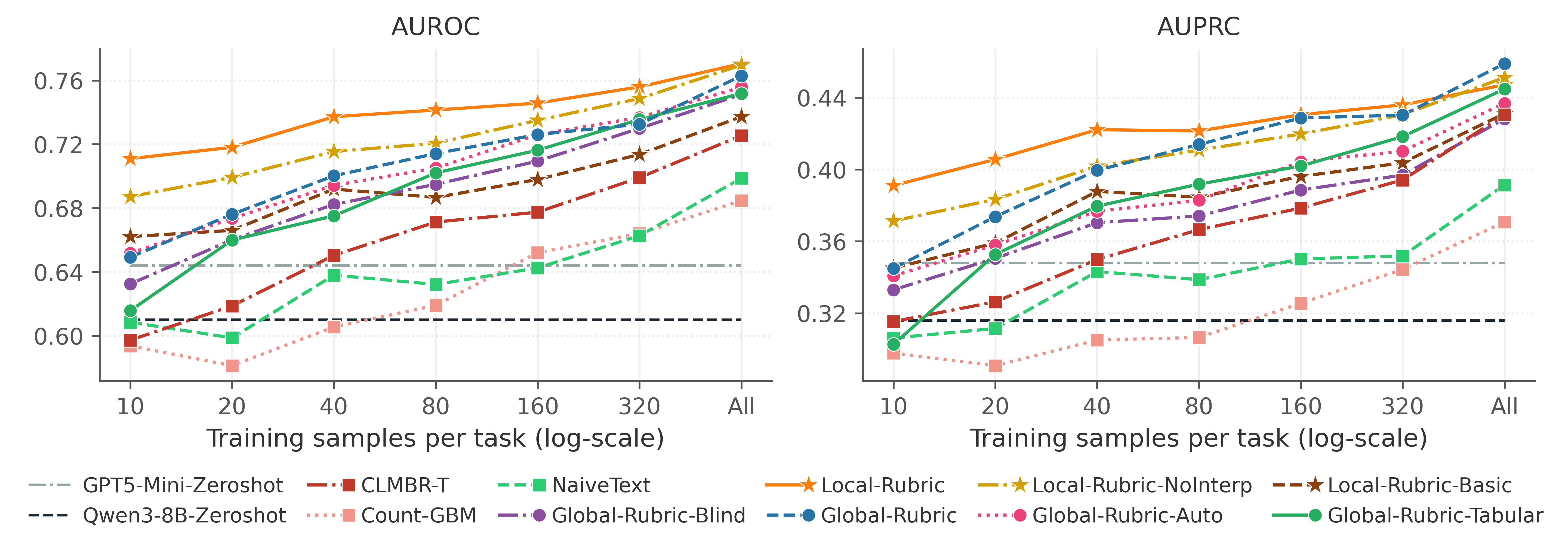}
    \caption{Average performance over all tasks. For confidence bands, see Appendix~\ref{app:figs_with_ci}.}
    \label{fig:sweep_overall}
    \vspace{-5pt}
\end{figure}

\textbf{Input representation design is crucial to performance gains.}\hspace{11pt}
\textcolor{methodblue}{NaiveText} and our rubric methods share the same downstream stack
(a pretrained text-embedding model with a logistic-regression head), so
the gap between them isolates the value of LLM-driven representation
design in the language space. At $n{=}10$/$\text{All}$, \textcolor{methodblue}{Local-Rubric}
improves on \textcolor{methodblue}{NaiveText} by $0.09$/$0.07$ AUROC ($0.08$/$0.06$ AUPRC) and
\textcolor{methodblue}{Global-Rubric} by $0.04$/$0.06$ AUROC ($0.04$/$0.07$ AUPRC).

\definecolor{LocalQualBox}{HTML}{FF7F0E}
\definecolor{NoInterpQualBox}{HTML}{D4A000}
\definecolor{GlobalQualBox}{HTML}{2874A6}
\colorlet{LocalQualBoxBg}{LocalQualBox!6}
\colorlet{NoInterpQualBoxBg}{NoInterpQualBox!6}
\colorlet{GlobalQualBoxBg}{GlobalQualBox!6}

\newlength{\QualBoxHeight}
\setlength{\QualBoxHeight}{0.25\textheight}

\begin{figure}[t]
\centering
\begin{minipage}[t]{0.32\linewidth}
\begin{tcolorbox}[
    colback=LocalQualBoxBg,
    colframe=LocalQualBox,
    boxrule=1pt,
    arc=5pt,
    height=\QualBoxHeight,
    valign=top,
    top=3pt, bottom=3pt, left=5pt, right=5pt,
    boxsep=3pt
]
\scriptsize
\textbf{\textcolor{methodblue}{Local-Rubric} representation}
\vspace{5pt}
\hrule
\vspace{5pt}
72-year-old female. Serum sodium $138 \to 134 \to 131$~mmol/L over 24 days, last value 5 days before prediction. Hydrochlorothiazide 25~mg daily started 12 days before prediction. Thiazide-induced hyponatremia is most pronounced in the first 1--2 weeks of initiation, raising the prior probability of an abnormal sodium. Mild congestive heart failure adds a chronic dilutional pathway. Renal function preserved (Cr 0.9~mg/dL), making a renal-failure mechanism unlikely. Documented fatigue and mild nausea at the most recent visit are consistent with early symptomatic hyponatremia.\\[2pt]
\textbf{Overall: elevated prior risk.}
\end{tcolorbox}
\end{minipage}\hfill
\begin{minipage}[t]{0.32\linewidth}
\begin{tcolorbox}[
    colback=NoInterpQualBoxBg,
    colframe=NoInterpQualBox,
    boxrule=1pt,
    arc=5pt,
    height=\QualBoxHeight,
    valign=top,
    top=3pt, bottom=3pt, left=5pt, right=5pt,
    boxsep=3pt
]
\scriptsize
\textbf{\textcolor{methodblue}{Local-Rubric-NoInterp} \\ representation}
\vspace{5pt}
\hrule
\vspace{5pt}
72-year-old female. Serum sodium $138 \to 134 \to 131$~mmol/L over 24 days, last value 5 days before prediction. Hydrochlorothiazide 25~mg daily started 12 days before prediction. Documented mild congestive heart failure. Renal function preserved (Cr 0.9~mg/dL). Documented fatigue and mild nausea at the most recent visit.\\[2pt]

\end{tcolorbox}
\end{minipage}\hfill
\begin{minipage}[t]{0.32\linewidth}
\begin{tcolorbox}[
    colback=GlobalQualBoxBg,
    colframe=GlobalQualBox,
    boxrule=1pt,
    arc=5pt,
    height=\QualBoxHeight,
    valign=top,
    top=3pt, bottom=3pt, left=5pt, right=5pt,
    boxsep=3pt
]
\scriptsize
\textbf{\textcolor{methodblue}{Global-Rubric} representation}
\vspace{5pt}
\hrule
\vspace{5pt}
\textbf{Patient}: Age 72, Sex FEMALE.

\vspace{2pt}

\textbf{ProblemListFlags}: CKD/ESRD No, Dialysis No, Prior hyponatremia No, Active malignancy No.

\vspace{2pt}

\textbf{SerumSodium\_Last3}: 2024-08-10 (5d) 131; 2024-08-01 (14d) 134; 2024-07-22 (24d) 138 (mmol/L).

\vspace{2pt}

\textbf{SerumSodium\_Min90}: 131~mmol/L (2024-08-10).

\vspace{2pt}

\textbf{Meds affecting sodium}: 2024-08-03 hydrochlorothiazide, thiazide diuretic, oral, 25~mg daily.

\vspace{2pt}

\textbf{RenalFunction}: Cr 0.9~mg/dL, BUN 18~mg/dL (2024-08-10).

\vspace{2pt}

\textbf{Other fields}: NA.
\end{tcolorbox}
\end{minipage}
\caption{Three rubric outputs for the same synthetic hyponatremia case. \textcolor{methodblue}{Local-Rubric} renders the evidence as a task-aligned risk assessment, \textcolor{methodblue}{Local-Rubric-NoInterp} keeps the same evidence with interpretive language stripped, and \textcolor{methodblue}{Global-Rubric} extracts it into standardized fields.}
\label{fig:local_vs_global_qual}
\vspace{-10pt}
\end{figure}

\textbf{Local versus global rubrics: interpretation helps most when labels are scarce.}\hspace{11pt}
\textcolor{methodblue}{Local-Rubric} is strongest at $n=10$ (0.701 AUROC, 0.382 AUPRC), while
\textcolor{methodblue}{Global-Rubric} nearly closes the gap by $n=\mathrm{All}$ and slightly exceeds
\textcolor{methodblue}{Local-Rubric} in overall AUPRC (0.459 vs.\ 0.452). This pattern suggests that the
interpretive language in \textcolor{methodblue}{Local-Rubric} is most useful when the downstream
classifier has few labels. \textcolor{methodblue}{Local-Rubric} outputs often translate raw findings into
task-conditioned clinical and statistical language, whereas \textcolor{methodblue}{Global-Rubric} extracts
similar evidence into standardized fields without explicit risk interpretation
(\Cref{fig:local_vs_global_qual}, left and right). Thus, \textcolor{methodblue}{Local-Rubric} appears to provide a useful
inductive bias/prior in the language representation at small $n$; as labels accumulate,
the downstream classifier can increasingly learn from the structured evidence
itself, and the advantage of explicit interpretation diminishes.

\textbf{Ablations separate the effects of interpretation and standardization.}\hspace{11pt}
\textcolor{methodblue}{Local-Rubric-NoInterp} preserves the patient-level evidence and missingness cues from
\textcolor{methodblue}{Local-Rubric} while removing explicit risk-stratification language. Its drop at small
$n$, followed by convergence with \textcolor{methodblue}{Local-Rubric} at $n=\mathrm{All}$, is consistent with
interpretive LLM-generated language acting as an inductive bias when labels are scarce.
\textcolor{methodblue}{Local-Rubric-Basic} removes both this interpretive layer and some of the richer
missingness/context cues. Compared with \textcolor{methodblue}{Global-Rubric}, which also avoids explicit risk
interpretation but imposes a standardized template, the two are comparable at $n=10$,
while \textcolor{methodblue}{Global-Rubric} pulls ahead at $n=\mathrm{All}$. This suggests that global
standardization becomes increasingly valuable once the downstream classifier has enough
labels to exploit the structured representation.

\textbf{Global rubric variants: cost of automation.}\hspace{11pt}
\textcolor{methodblue}{Global-Rubric-Auto}, which replaces per-example LLM calls with a
deterministic parser, tracks \textcolor{methodblue}{Global-Rubric} closely on AUROC across
sample sizes and gives up about $0.02$ AUPRC at $n{=}\text{All}$ ($0.437$
vs $0.459$), recovering most of the gain over baselines at near-zero marginal LLM cost. \textcolor{methodblue}{Global-Rubric-Tabular} lags visibly at small $n$, which we
attribute to XGBoost's inability to leverage a prior, unlike the
text-embedding models used in other methods. As expected, it nearly catches
up at $n{=}\text{All}$, and on lab tasks it is in fact the single strongest
method overall. Finally, \textcolor{methodblue}{Global-Rubric-Blind} generates the rubric from
the task description alone. It continues to outperform the baselines across
sample sizes but remains inferior to \textcolor{methodblue}{Global-Rubric}, showing the value of
letting the LLM examine some data first (Panel A in
\Cref{fig:rubric_creation_pipeline}).
\begin{figure}[t]
    \centering
    \includegraphics[width=\linewidth]{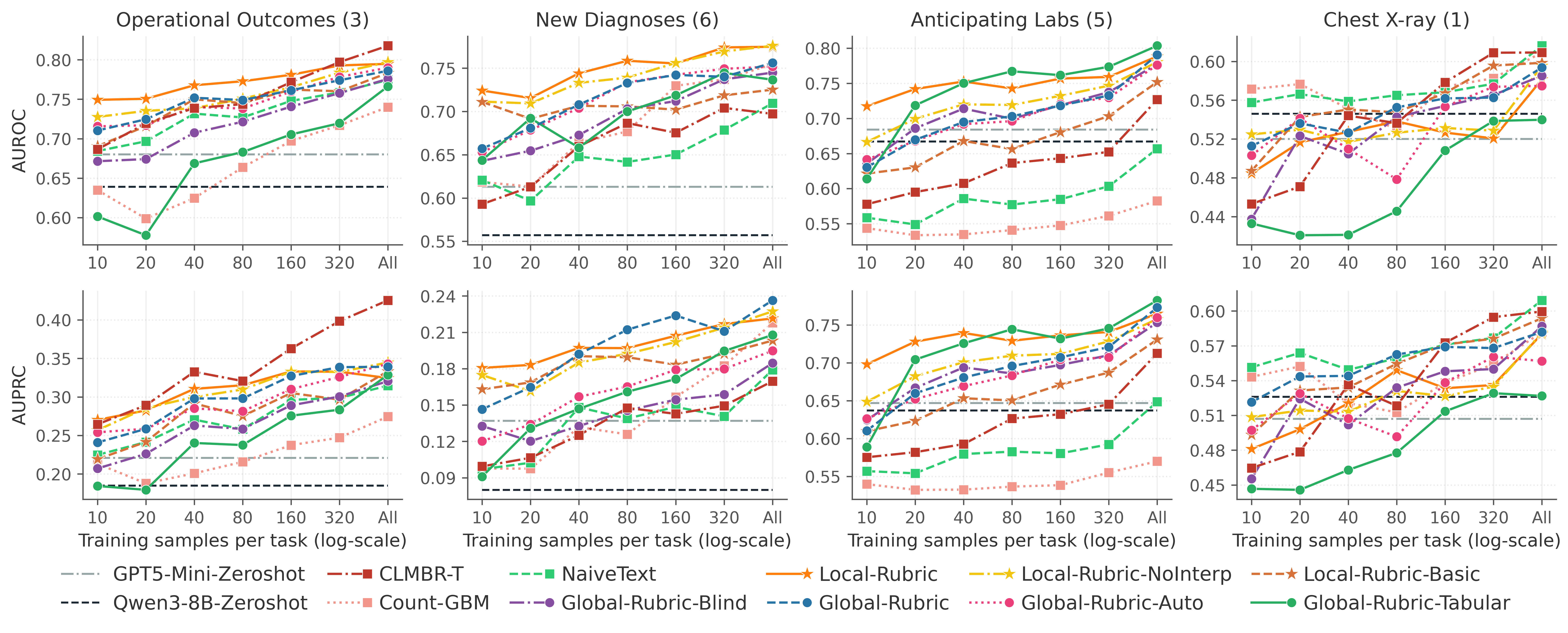}
    \caption{Average performance by task group. For confidence bands, see Appendix~\ref{app:figs_with_ci}.}
    \label{fig:sweep_groups}
    \vspace{-10pt}
\end{figure}

\textbf{Operational outcomes.}\hspace{11pt}
\textcolor{methodblue}{CLMBR-T} is the strongest here at $n{=}\text{All}$
($0.818$ AUROC, $0.425$ AUPRC), edging out \textcolor{methodblue}{Local-Rubric} ($0.802$, $0.341$)
and \textcolor{methodblue}{Global-Rubric} ($0.786$, $0.339$), with a more pronounced lead in
AUPRC. We attribute this to alignment between \textcolor{methodblue}{CLMBR-T}'s next-code
pretraining and operational targets. For instance, every visit in the
pretraining corpus contributes an implicit ``label'' for the ICU admission
task via next-code prediction, which is not the case for other tasks such
as lab results. Even so, in the small-sample regime our rubrics overtake
\textcolor{methodblue}{CLMBR-T} in AUROC (\textcolor{methodblue}{Local-Rubric} $0.750$, \textcolor{methodblue}{Global-Rubric} $0.710$ at
$n{=}10$, vs \textcolor{methodblue}{CLMBR-T} $0.687$) and match it in AUPRC.

\textbf{Lab results.} \hspace{11pt}
Rubric methods deliver large gains here. At $n{=}\text{All}$, \textcolor{methodblue}{Global-Rubric-Tabular} is the best method, and strong small-sample performance of \textcolor{methodblue}{Local-Rubric} prevails. Rubrics align well with this type of tasks, where the predictive signal lives in a compact, recency-aware set of measurements (recent labs, trends, contributing medications). The lab rubrics
(Appendix~\ref{app:fullrubrics}) expose those directly, whereas raw serializations leave them scattered across visit narratives, buried in noise.

\textbf{New diagnoses and chest X-ray.} \hspace{11pt}
Rubrics dominate the new-diagnosis group at small $n$ (\textcolor{methodblue}{Local-Rubric}
$0.713$/$0.170$ AUROC/AUPRC vs \textcolor{methodblue}{Count-GBM} $0.619$/$0.098$ at $n{=}10$),
roughly $1.5$--$2{\times}$ above baseline AUPRC and especially strong on
rare-event tasks
(\Cref{tab:embedding_task_newpancan,tab:embedding_task_newlupus}).
Chest
X-ray is the noisiest task in EHRSHOT. The binary label collapses an
originally $14$-way radiology label, and many patients have their last
documented visit more than a month before the prediction time. All methods
cluster near $0.60$ AUROC at $n{=}\text{All}$, consistent with limited
learnable signal in this label given the available context, rather than a
method-specific failure. We cannot, however, fully rule out room for
method-specific improvement.

\textbf{Robustness across text-embedding backbones.} \hspace{11pt}
The rubric-over-\textcolor{methodblue}{NaiveText} ordering is preserved with Mistral-7B,
LLaMA-3-8B (LLM2Vec), and OpenAI's text-embedding-3-large
(Appendix~\ref{app:smaller_embedding}), in both AUROC and AUPRC.
\section{Qualitative analysis of a global rubric for hypertension onset prediction} 
\label{sec:qual_results}
We examine the learned global rubric for the new hypertension diagnosis prediction task. An excerpt is shown in \Cref{fig:htn_rubric_excerpt}, with the full rubric in Appendix~\ref{app:htn_rubric}.

\textbf{Temporal standardization and BP normalization.} \hspace{11pt}
The rubric first imposes structure on noisy EHR data before extracting features. In the preparation stage (\Cref{fig:htn_rubric_excerpt}, A), it defines explicit time windows: very recent ($\leq 30$ days), recent (31--180 days), and baseline/remote ($>180$ days), and standardizes units and formats, including blood pressure (BP) in mmHg and weight in kg. This is important for hypertension prediction because recency, persistence, and measurement comparability are central to distinguishing sustained hypertension risk from isolated or context-specific elevations.

\definecolor{OrBox}{HTML}{FF8C00}
\colorlet{OrBoxBg}{OrBox!6}

\begin{figure}[t]
\centering
\begin{tcolorbox}[
    colback=OrBoxBg,
    colframe=OrBox,
    boxrule=1pt,
    arc=5pt,
    width=\linewidth,
    top=2pt, bottom=2pt, left=6pt, right=6pt,
    boxsep=3pt
]
\scriptsize
\textbf{A. Preparation (before extracting)}\\
- Define time windows: \textit{Very recent}: last 30 days, \textit{Recent}: 31-180 days, \textit{Baseline/remote}: $>$180 days \\
- Standardize units: \textit{Blood pressure}: mmHg (systolic/diastolic), \textit{Weight}: kg or oz $\rightarrow$ convert to kg {[...]}

\vspace{0.5em}
\textbf{Step 2 - Blood pressure (BP) data extraction and normalization}\\
- Extract systolic/diastolic BP values with timestamps and context (office, inpatient, ED, home, ambulatory, perioperative).\\
- For each time window (very recent, recent, baseline): compute count, mean, median, SD, min, max; flag highest recent BP\\
- Compute simple trend metrics (e.g., recent slope; BP variability via SD).\\
- Categorize BP per ACC/AHA categories using aggregated recent values: Normal ($<$120/$<$80), Elevated (120-129/$<$80) {[...]}

\vspace{0.5em}
\textbf{Step 9 - Synthesis per domain (structured fields and scoring)}\\
- For each domain, record presence, supporting data, recency, and confidence (High/Moderate/Low).\\
- {\em Domain A} - BP phenotype: last BP (date/context), mean recent BP (last 30d; 31-180d), BP category, variability flag, ambulatory BP.\\
- {\em Domain B} - Metabolic / vascular risk: Diabetes (Y/N) - last A1c ($\%$ and date), BMI and obesity category, Hyperlipidemia (Y/N) - LDL value and date, Smoking (current/former/never).
Create a simple domain scorecard: number of High/Moderate/Minor risk features {[...]}
\end{tcolorbox}
\vspace{-5pt}
\caption{Key excerpts from learned hypertension global rubric instructions.}
\label{fig:htn_rubric_excerpt}
\vspace{-5pt}
\end{figure}

\textbf{Clinically grounded BP feature construction.} \hspace{11pt}
Step 2 converts raw BP readings into structured longitudinal features. The rubric first extracts systolic/diastolic values with timestamps and clinical context. Within each temporal window, it computes summary statistics such as mean and maximum. It also derives simple trend and variability measures such as recent slope and standard deviation (SD). Finally, it maps aggregated recent BP values to ACC/AHA categories, converting irregular raw measurements into clinically interpretable BP phenotype features \citep{whelton20182017}.

\textbf{Domain-level synthesis.} \hspace{11pt}
Step 9 further compresses the extracted evidence into domain-level summaries and a scorecard. For each domain, the rubric records presence, supporting data, recency, and confidence. Domain A summarizes BP phenotype, including last BP, recent means, BP category, variability, and ambulatory BP; Domain B captures metabolic and vascular risk factors such as diabetes, BMI/obesity, hyperlipidemia, LDL, and smoking. The resulting counts of high-, moderate-, and minor-risk features provide compact task-specific signals that can be used to effectively predict the cumulative hypertension risk across BP patterns and comorbidities.

More broadly, this highlights the value of global rubrics as a bridge between raw, heterogeneous inputs and downstream prediction. By making task-relevant evidence explicit and structured, rubrics provide models with more interpretable, standardized, and prediction-aligned representations. As a complementary case study, in Appendix~\ref{app:hyponatremia_case_study} we examine the {\em learned tabular features} produced by \textcolor{methodblue}{Global-Rubric-Tabular} (\Cref{fig:rubric_creation_pipeline}, Panel F) for the hyponatremia lab task and show that they closely track the standard clinical decision tree for evaluating hyponatremia.

\section{Concluding remarks and limitations}
\label{sec:conclusion}

We proposed \emph{rubric representation learning}, where LLMs transform naive text-serializations into task-aligned representations before downstream training. Across 15 EHRSHOT tasks, rubric methods substantially outperformed baselines, particularly for small $n$. \textcolor{methodblue}{Local-Rubric} was strongest at small $n$, suggesting that LLMs' pretrained knowledge can act as a major contributor to downstream performance. Its performance was matched by \textcolor{methodblue}{Global-Rubric} at large $n$, demonstrating the advantages of having a standardized input representation. Our ablations decompose the rubric advantage into two complementary levers, an LLM-injected statistical prior that drives performance at small $n$ and a standardized template that adds representational value at scale. We further showed that automated global-rubric variants remain competitive while drastically reducing inference cost.

\paragraph{Limitations and future work.}
Our evaluation is restricted to a single benchmark and does not include richer modalities such as clinical notes or images. Currently, rubric synthesis is bounded by context length (40 patients per global rubric), and iterative refinement using additional examples, failure cases, or expert feedback is a natural next step. Further, we report a single global rubric per task. Assessing the sensitivity of downstream performance to the cohort sampled for rubric synthesis is an important but costly direction for future work. Finally, clinical deployment should require expert review, subgroup evaluation, privacy safeguards, and monitoring for errors or distribution shift.
 
\clearpage
\bibliography{ref}
\bibliographystyle{abbrvnat}

\appendix

\section{Extended related work} \label{app:related_work}
\paragraph{Data science agents.}
A complementary line of work studies ``data science (DS) agents''. Data Interpreter targets end-to-end problem solving, using code generation, execution, and revision to complete data analysis and mathematical tasks \citep{hong2025data_interpreter}. DS-Agent focuses on automating model development workflows such as task understanding, model selection, and training \citep{guo2024ds_agent}. DeepAnalyze and DS-STAR push further toward autonomous data science over heterogeneous files, with an emphasis on multi-step data wrangling, open-ended querying, code execution, and report generation \citep{zhang2025deepanalyze,nam2025dsstar}. This literature is closely related to our work in using LLMs as an interface to heterogeneous data. Our focus, however, is narrower and more controlled. Rather than asking LLM agents to plan and execute broad analyses, we study how LLMs can support input representation design with complex data for specific downstream tasks. This lets us isolate the role of representation choice and demonstrate its effect as a first-order driver of downstream statistical performance.

\paragraph{Medical QA benchmarks.} 
Recent work demonstrated that LLMs encode substantial clinical knowledge. \citet{singhal2023large} introduced Med-PaLM, a closed-source medical LLM from Google. They evaluated it on MultiMedQA, a benchmark combining six medical question answering datasets, and showed that instruction-tuned models could surpass prior state-of-the-art. The follow-up Med-PaLM 2 pushed further, achieving performance competitive with expert physicians \citep{singhal2025toward}. \citet{sandmann2025benchmark} evaluated open-source DeepSeek models on clinical decision support tasks using 125 patient cases. They found that the open-source frontier models perform equally well, if not better, than proprietary models. \citet{zhang2025med} showed that even a 3B-parameter model can develop some medical reasoning skills when trained via reinforcement learning with verifiable rewards (RLVR, \citep{lambert2024tulu}) on the MedQA benchmark \citep{jin2021disease}.

\paragraph{Moving beyond medical QA benchmarks.}
While these results are impressive, high scores on medical QA do not translate to clinical impact. \citet{mehandru2024evaluating} proposed AI-based Standardized Clinical Examination (AI-SCE), modeled after the Objective Structured Clinical Examination (OSCE, \citep{zayyan2011objective}) used in medical training, to evaluate LLMs as agents in realistic, multi-step clinical scenarios rather than static question-answering benchmarks. \citet{kim2025limitations} identified that LLMs can exhibit inflexible reasoning in clinical problem-solving and struggle with unexpected {\em long-tail} situations. \citet{bedi2026holistic} introduced MedHELM, a holistic evaluation framework that organizes medical tasks into a taxonomy spanning clinical decision support, note generation, patient communication, research, and administration. Their analyses revealed that LLMs perform more variably on realistic clinical tasks than on standardized exams. \citet{jiang2025medagentbench} developed MedAgentBench, a virtual EHR environment specifically designed to benchmark LLM agents on multi-step clinical tasks.

\paragraph{LLMs in the clinic.}
Recent work probes how LLMs can be deployed to support clinicians and patients. \citet{garcia2024artificial} demonstrated that AI-generated draft replies to patient messages can reduce physician burden without sacrificing quality. \citet{yalamanchili2024quality} evaluated the quality of LLM responses to radiation oncology patient queries. \citet{unlu2024retrieval} showed that retrieval-augmented GPT-4 (RAG, \citep{lewis2020retrieval}) can assist with clinical trial screening. Randomized trials have also begun to probe whether LLMs can reliably improve clinicians' performance. \citet{goh2025gpt} conducted a trial and found that GPT-4 assistance improved physician performance, while \citet{wan2024outpatient} showed that nurse-LLM collaboration for outpatient reception resulted in ``increased satisfaction among both patients and nurses''. 

\paragraph{Conversational diagnostic models and specialist systems.}
Google's Articulate Medical Intelligence Explorer (AMIE) represents a sustained research program on LLMs in healthcare. \citet{tu2025towards} introduced AMIE, a conversational and diagnostic AI system, trained via self-play in a simulated environment to conduct multi-turn clinical conversations. They showed that it outperformed primary care physicians on 30 of 32 evaluation axes in a randomized, blinded study using standardized patient actors. \citet{mcduff2025towards} evaluated AMIE's capacity for differential diagnosis, demonstrating that it generates diagnostic lists that exceed GPT-4's quality and improve clinicians' diagnostic accuracy when used as an assistive tool. Subsequent work extended AMIE to specialist domains. \citet{palepu2025exploring} evaluated its performance in oncology care, and \citet{o2026large} reported results for complex cardiology cases. 

\paragraph{LLMs with EHR data.}
There is a growing literature developing medical foundation models pretrained on large EHR or claims data for risk prediction and trajectory modeling \citep{steinberg2023motor, renc2024ethos, waxler2025generative}. Recent work shows general-purpose LLMs can match domain-specific models on clinical tasks \citep{hegselmann2025large}, which we reproduce and strengthen with our LLM-derived rubrics.

\citet{agrawal2022large} showed that LLMs are effective clinical information extractors, pulling structured data from unstructured clinical text with few examples. \citet{fleming2024medalign} released MedAlign, a clinician-generated dataset for instruction following, targeting realistic EHR-grounded tasks. \citet{shi2024ehragent} proposed EHRAgent, an LLM agent that generates and executes code to answer clinician queries. 
\citet{lin2025training} combined supervised fine-tuning with reinforcement learning, demonstrating gains across medical calculation, patient-trial matching, and disease diagnosis in EHRSHOT benchmark. \citet{liao2025ehr} developed EHR-R1, a reasoning-enhanced model for EHR analysis using reinforcement learning. \citet{kirchler2026large} demonstrated that LLM-based clinical prediction models can have improved cross-country and system transferability. \citet{yoon2026paregta} proposed an encoding approach for EHR data using LLMs to better emphasize temporal information. 

\paragraph{Connections to AI for science and research.}
There is a growing literature on using LLMs to automate parts of the scientific process. Systems such as the AI Scientist \citep{lu2024ai, yamada2025ai} generate hypotheses and iteratively design experiments. Other work emphasizes execution-grounded research pipelines, where LLM-generated plans are validated through execution and feedback \citep{si2026towards}. \citet{shao2025dr} explore evolving rubrics to guide multi-step research. These directions are conceptually related to our setting: LLMs are used to generate intermediate representations and pipelines for complex tasks and datasets, that are subsequently executed and evaluated for downstream objectives.

\clearpage
\section{EHRSHOT subsampling procedure and dataset statistics} \label{app:ehrshot}
\label{app:ehrshot_subsampling}

\begin{table}[ht]
\centering
\caption{Number of samples (positive cases) per task and split. All splits for Lab results and Chest X-ray tasks are subsampled from the original dataset. Validation splits are subsampled for all tasks.}
\label{tab:sample_counts}
\begin{tabular}{p{3.6cm} l r r r}
\toprule
Category & Task & Train & Val & Test \\
\midrule

\multirow{3}{=}{Operational Outcomes (3)}
& ICU transfer & 2402 (113) & 100 (50) & 2037 (85) \\
& Length of stay $>$7 days & 2569 (681) & 100 (50) & 2195 (552) \\
& 30-day readmission & 2608 (370) & 100 (50) & 2189 (260) \\

\midrule

\multirow{6}{=}{Assignment of New Diagnosis (6)}
& Hypertension & 1259 (182) & 100 (50) & 1258 (159) \\
& Hyperlipidemia & 1684 (205) & 100 (50) & 1317 (172) \\
& Pancreatic cancer & 2576 (155) & 100 (50) & 2220 (56) \\
& Celiac disease & 2623 (62) & 22 (11) & 2222 (21) \\
& Lupus & 2570 (104) & 66 (33) & 2243 (20) \\
& Acute MI & 2534 (175) & 100 (50) & 2127 (144) \\

\midrule

\multirow{5}{=}{Anticipating Labs (5)}
& Thrombocytopenia & 2000 (1000) & 100 (50) & 2000 (1000) \\
& Hyperkalemia & 2000 (1000) & 100 (50) & 1896 (948) \\
& Hypoglycemia & 2000 (1000) & 100 (50) & 1566 (783) \\
& Hyponatremia & 2000 (1000) & 100 (50) & 2000 (1000) \\
& Anemia & 2000 (1000) & 100 (50) & 2000 (1000) \\

\midrule

\multirow{1}{=}{Chest X-ray Findings (1)}
& Chest X-ray abnormality & 2000 (1000) & 100 (50) & 2000 (1000) \\

\bottomrule
\end{tabular}
\end{table}

We evaluate on the 15 binary prediction tasks from EHRSHOT \citep{wornow2023ehrshot} listed in \Cref{sec:ehrshot}. Several of our methods invoke an LLM call per example, so we subsample some splits for budget reasons. Final per-split sample counts are reported in \Cref{tab:sample_counts}.

Let $n_+$ denote the number of positive examples in the corresponding original split. Validation sets are label-balanced with up to $\min(50,n_+)$ positives and the same number of negatives, so most validation sets contain $50$/$50$. For operational and diagnosis tasks we retain the original EHRSHOT training and test splits, which are moderate in size. For lab and chest X-ray tasks the original splits are substantially larger, and we subsample training and test splits to up to $\min(1000,n_+)$ positives and the same number of negatives, yielding balanced subsets of up to $2{,}000$ examples per split; when fewer positives are available, negatives are matched to the available positives.

\section{Reproduction and compute costs} \label{app:compute_costs}
We summarize the practical resource requirements for reproducing our experiments on the subsampled EHRSHOT splits described above (\Cref{app:ehrshot}, \Cref{tab:sample_counts}).

\textbf{NaiveText truncation.}\hspace{11pt} Following \citet{hegselmann2025large}, we clip \textcolor{methodblue}{NaiveText} serializations at $8{,}192$ tokens (Qwen3-8B tokenizer).

\textbf{LLM API costs.}\hspace{11pt} Methods with a per-example LLM call---\textcolor{methodblue}{Global-Rubric}, \textcolor{methodblue}{Global-Rubric-Blind}, \textcolor{methodblue}{Local-Rubric}, and \textcolor{methodblue}{Local-Rubric-Basic}---each cost approximately \$50 per task in GPT5-Mini API usage on our subsampled splits, totaling roughly \$3{,}000 across 4 methods and 15 tasks. Global-rubric synthesis and the one-time parser/tabularization scripts used by \textcolor{methodblue}{Global-Rubric-Auto} and \textcolor{methodblue}{Global-Rubric-Tabular} (\Cref{fig:rubric_creation_pipeline}, Panels E--F) add a small one-off cost; once those scripts are produced, applying them to new examples at deployment time is essentially free.

\textbf{Embedding compute.}\hspace{11pt} Text embeddings (\texttt{Qwen3-Embedding-8B} for all main results, plus the open-weights backbones used in \Cref{app:smaller_embedding}) are computed on a node with 4 NVIDIA A100 GPUs. Embedding the full dataset for any given representation method (\textcolor{methodblue}{NaiveText} or any rubric variant) takes a few hours per method.

\textbf{Downstream training.}\hspace{11pt} The downstream classifiers (logistic regression on frozen embeddings; LightGBM for \textcolor{methodblue}{Count-GBM}; XGBoost for \textcolor{methodblue}{Global-Rubric-Tabular}) are lightweight and run on commodity CPU hardware in seconds to minutes per task; their cost is negligible compared to the embedding and LLM-call steps above.

\clearpage
\section{Case study: learned tabular features for prediction of hyponatremia}
\label{app:hyponatremia_case_study}

As a complement to the qualitative analysis in \Cref{sec:qual_results}, we examine the learned tabular features (\Cref{fig:rubric_creation_pipeline}, Panel F) for prediction of hyponatremia abnormality. Across the 15 tasks, the auto-generated rubric feature schemas range from 147 to 450 features per task and cluster around 200--250 features. They are predominantly binary (72\%), followed by numeric (19\%) and categorical (9\%). The high binary share reflects pervasive one-hot encoding of categoricals and the inclusion of a \texttt{\_missing} indicator for nearly every field. Numeric features capture lab values, vitals, and counts.

We focus on learned tabular features for the hyponatremia lab task, for which the full global rubric is given in Appendix~\ref{app:hyponatremia_rubric}. We make several observations. The feature structure closely mirrors the diagnostic decision tree used clinically when evaluating hyponatremia. A first step in clinical reasoning is determining whether apparent hyponatremia is physiologic or artificially low due to hyperglycemia or other osmotic effects; accordingly, the rubric extracts recent glucose measurements ($\texttt{Glucose-Last3}$) and serum osmolality values, while the tabular features include indicators reflecting level of glucose in the blood, {\em e.g.} $\texttt{glucose-type-blood-present}$. If true hypotonic hyponatremia is present, clinicians next evaluate urine osmolality and urine sodium to distinguish between states of antidiuretic hormone (ADH) activity and renal sodium handling, which helps identify etiologies such as SIADH or hypovolemia~\citep{spasovski2014clinical}. Consistent with this framework, the rubric extracts urine sodium, urine osmolality, and conditions associated with SIADH ({\em e.g.}, pulmonary infections, CNS disorders, malignancy). The resulting features include indicators that reflect such conditions, {\em e.g.}, $\texttt{acute-cond-Pulmonary infection / pneumonia / pulmonary disease}$ and $\texttt{acute-cond-any}$. Thus, the learned features directly operationalize the same diagnostic flow in clinical practice.

A second group of features captures baseline risk factors and comorbidities that predispose patients to hyponatremia. The rubric explicitly extracts conditions such as chronic kidney disease (CKD), dialysis history, malignancy, and medications known to induce hyponatremia ({\em e.g.,} thiazide diuretics). These signals appear directly in the tabularized representation through features such as \texttt{dialysis-history-Yes}, \texttt{procedure-Hemodialysis}, and \texttt{med-class-count-thiazide-diuretic}. These variables correspond to well-known clinical risk factors for hyponatremia, including impaired renal free-water handling and medication-induced sodium loss~\citep{verbalis2013diagnosis}.

Finally, the most predictive signals arise from the acuity and trajectory of prior sodium measurements. The rubric explicitly extracts the three most recent sodium values and the lowest sodium in the prior 90 days, along with contextual metadata such as the setting of the measurement ({\em e.g.,} inpatient vs outpatient). Correspondingly, the largest-magnitude coefficients in the tabular feature set correspond to prior sodium measurements, including \texttt{serum-na-recent-le-134} and \texttt{prior-documented-hyponatremia-Yes}. Clinically, this is expected, as patients with a history of chronic or recurrent hyponatremia ({\em e.g.}, due to heart failure or cirrhosis) are substantially more likely to have abnormal sodium levels on subsequent laboratory testing~\citep{upadhyay2006incidence}.

This example illustrates how rubric representations surface task-relevant information that would otherwise be buried in a long, heterogeneous text-serialization of the patient record. In the naive text format, prior sodium measurements appear scattered across multiple visits and lab panels, interleaved with unrelated clinical events. The rubric reorganizes this information into a compact set of fields that explicitly capture the recent trajectory of the lab. In doing so, it converts diffuse signals in language space into structured features that a simple downstream model can use efficiently.
\clearpage
\section{Additional quantitative results} \label{app:additional results}
\subsection{Average numerical results for $n=10$ and $n=\text{All}$ regimes} \label{app:emb_table}
\providecolor{localrubric}{RGB}{190, 100, 25}
\providecolor{globalrubric}{RGB}{45, 90, 165}
\renewcommand{\arraystretch}{0.94}
\setlength{\tabcolsep}{3pt}
\begin{table}[htbp]
\centering
\small
\caption{AUROC and AUPRC across all methods for $n=10$ and $n=\text{All}$. \textbf{Green}: best in entire column. \textbf{Blue}: best within a method across backbones. \textbf{Oper.:} Operational Outcomes, \textbf{New Dx:} Assignment of New Diagnosis, \textbf{Labs:} Anticipating Lab Results, \textbf{CXR:} Chest X-ray Findings}
\label{tab:embedding_8b}
\vspace{0.5em}
\textbf{(a) AUROC} \\[0.2em]
\begin{tabular}{llccccc}
\toprule
 & & \makecell{\textbf{Overall}\\\textbf{(15)}} & \makecell{\textbf{Oper.}\\\textbf{(3)}} & \makecell{\textbf{New Dx}\\\textbf{(6)}} & \makecell{\textbf{Labs}\\\textbf{(5)}} & \makecell{\textbf{CXR}\\\textbf{(1)}} \\
\midrule
\multirow{2}{*}{$n=0$} & Qwen3-8B-Zeroshot & .610$_{.601-.618}$ & .639$_{.616-.660}$ & .557$_{.541-.575}$ & .667$_{.658-.677}$ & \cellcolor{bestblue}.546$_{.519-.573}$ \\
 & GPT5-Mini-Zeroshot & \cellcolor{bestblue}.644$_{.635-.653}$ & \cellcolor{bestblue}.680$_{.658-.700}$ & \cellcolor{bestblue}.613$_{.594-.632}$ & \cellcolor{bestblue}.684$_{.675-.694}$ & .520$_{.498-.541}$ \\
\midrule
\multirow{10}{*}{$n=10$} & Count-GBM & .594$_{.579-.609}$ & .635$_{.610-.658}$ & .619$_{.583-.652}$ & .544$_{.532-.555}$ & \cellcolor{bestblue}.571$_{.541-.602}$ \\
 & CLMBR-T & .597$_{.582-.613}$ & .687$_{.662-.709}$ & .593$_{.559-.629}$ & .578$_{.567-.588}$ & .453$_{.421-.484}$ \\
 & NaiveText & .609$_{.595-.621}$ & .684$_{.664-.705}$ & .621$_{.591-.649}$ & .559$_{.547-.570}$ & .558$_{.533-.584}$ \\
 & \textcolor{localrubric}{Local-Rubric} & \cellcolor{bestblue}.701$_{.687-.715}$ & \cellcolor{bestblue}.750$_{.730-.768}$ & \cellcolor{bestblue}.713$_{.681-.743}$ & \cellcolor{bestblue}.694$_{.684-.705}$ & .517$_{.492-.543}$ \\
 & \textcolor{localrubric}{Local-Rubric-Basic} & .662$_{.648-.675}$ & .691$_{.670-.711}$ & .711$_{.682-.741}$ & .622$_{.610-.632}$ & .487$_{.457-.519}$ \\
 & \textcolor{localrubric}{Local-Rubric-NoInterp} & .687$_{.673-.701}$ & .728$_{.708-.746}$ & .711$_{.680-.741}$ & .666$_{.656-.677}$ & .525$_{.491-.556}$ \\
 & \textcolor{globalrubric}{Global-Rubric-Blind} & .632$_{.619-.646}$ & .671$_{.651-.691}$ & .643$_{.614-.672}$ & .636$_{.624-.647}$ & .437$_{.413-.461}$ \\
 & \textcolor{globalrubric}{Global-Rubric} & .649$_{.635-.662}$ & .710$_{.690-.732}$ & .657$_{.627-.688}$ & .630$_{.619-.641}$ & .513$_{.489-.538}$ \\
 & \textcolor{globalrubric}{Global-Rubric-Auto} & .652$_{.639-.666}$ & .716$_{.695-.736}$ & .653$_{.625-.683}$ & .641$_{.630-.653}$ & .503$_{.470-.534}$ \\
 & \textcolor{globalrubric}{Global-Rubric-Tabular} & .616$_{.605-.627}$ & .601$_{.575-.628}$ & .644$_{.619-.667}$ & .614$_{.603-.625}$ & .500$_{.500-.500}$ \\
\midrule
\multirow{10}{*}{$n=\text{All}$} & Count-GBM & .685$_{.670-.699}$ & .740$_{.717-.761}$ & .755$_{.722-.786}$ & .583$_{.572-.594}$ & .609$_{.576-.642}$ \\
 & CLMBR-T & .725$_{.711-.739}$ & \cellcolor{green!30}.818$_{.799-.836}$ & .697$_{.666-.728}$ & .727$_{.717-.737}$ & .609$_{.578-.640}$ \\
 & NaiveText & .699$_{.684-.714}$ & .775$_{.754-.793}$ & .709$_{.674-.744}$ & .657$_{.646-.668}$ & \cellcolor{green!30}.616$_{.592-.640}$ \\
 & \textcolor{localrubric}{Local-Rubric} & \cellcolor{green!30}.772$_{.758-.784}$ & .802$_{.786-.818}$ & .770$_{.738-.799}$ & .789$_{.780-.798}$ & .606$_{.583-.630}$ \\
 & \textcolor{localrubric}{Local-Rubric-Basic} & .737$_{.723-.752}$ & .783$_{.765-.800}$ & .725$_{.692-.758}$ & .752$_{.742-.761}$ & .599$_{.568-.628}$ \\
 & \textcolor{localrubric}{Local-Rubric-NoInterp} & .770$_{.757-.782}$ & .797$_{.779-.815}$ & \cellcolor{green!30}.776$_{.747-.803}$ & .780$_{.771-.789}$ & .596$_{.565-.625}$ \\
 & \textcolor{globalrubric}{Global-Rubric-Blind} & .751$_{.738-.764}$ & .776$_{.755-.794}$ & .745$_{.716-.775}$ & .777$_{.768-.787}$ & .585$_{.559-.608}$ \\
 & \textcolor{globalrubric}{Global-Rubric} & .763$_{.748-.777}$ & .786$_{.768-.805}$ & .756$_{.723-.789}$ & .791$_{.781-.800}$ & .594$_{.567-.619}$ \\
 & \textcolor{globalrubric}{Global-Rubric-Auto} & .756$_{.743-.769}$ & .790$_{.772-.807}$ & .752$_{.722-.782}$ & .776$_{.766-.786}$ & .575$_{.543-.605}$ \\
 & \textcolor{globalrubric}{Global-Rubric-Tabular} & .752$_{.740-.764}$ & .766$_{.748-.785}$ & .737$_{.710-.763}$ & \cellcolor{green!30}.804$_{.794-.813}$ & .540$_{.507-.572}$ \\
\bottomrule
\end{tabular}

\vspace{1em}

\textbf{(b) AUPRC} \\[0.2em]
\begin{tabular}{llccccc}
\toprule
 & & \makecell{\textbf{Overall}\\\textbf{(15)}} & \makecell{\textbf{Oper.}\\\textbf{(3)}} & \makecell{\textbf{New Dx}\\\textbf{(6)}} & \makecell{\textbf{Labs}\\\textbf{(5)}} & \makecell{\textbf{CXR}\\\textbf{(1)}} \\
\midrule
\multirow{2}{*}{$n=0$} & Qwen3-8B-Zeroshot & .316$_{.309-.324}$ & .185$_{.172-.200}$ & .080$_{.069-.093}$ & .637$_{.623-.650}$ & \cellcolor{bestblue}.526$_{.491-.560}$ \\
 & GPT5-Mini-Zeroshot & \cellcolor{bestblue}.348$_{.337-.362}$ & \cellcolor{bestblue}.221$_{.205-.240}$ & \cellcolor{bestblue}.137$_{.111-.169}$ & \cellcolor{bestblue}.647$_{.634-.659}$ & .507$_{.476-.538}$ \\
\midrule
\multirow{10}{*}{$n=10$} & Count-GBM & .298$_{.289-.307}$ & .213$_{.193-.235}$ & .098$_{.084-.116}$ & .540$_{.526-.555}$ & .543$_{.502-.583}$ \\
 & CLMBR-T & .315$_{.305-.326}$ & .265$_{.237-.294}$ & .099$_{.084-.116}$ & .575$_{.561-.590}$ & .465$_{.427-.500}$ \\
 & NaiveText & .306$_{.296-.316}$ & .225$_{.200-.253}$ & .097$_{.081-.114}$ & .557$_{.541-.572}$ & \cellcolor{bestblue}.551$_{.519-.583}$ \\
 & \textcolor{localrubric}{Local-Rubric} & \cellcolor{bestblue}.382$_{.369-.396}$ & \cellcolor{bestblue}.270$_{.244-.296}$ & .170$_{.143-.198}$ & \cellcolor{bestblue}.680$_{.664-.694}$ & .503$_{.474-.535}$ \\
 & \textcolor{localrubric}{Local-Rubric-Basic} & .345$_{.333-.359}$ & .219$_{.199-.240}$ & .163$_{.136-.193}$ & .610$_{.595-.624}$ & .493$_{.454-.531}$ \\
 & \textcolor{localrubric}{Local-Rubric-NoInterp} & .371$_{.356-.387}$ & .257$_{.232-.284}$ & \cellcolor{bestblue}.175$_{.144-.206}$ & .649$_{.634-.663}$ & .509$_{.469-.546}$ \\
 & \textcolor{globalrubric}{Global-Rubric-Blind} & .333$_{.322-.345}$ & .207$_{.189-.230}$ & .133$_{.112-.155}$ & .625$_{.609-.639}$ & .455$_{.427-.486}$ \\
 & \textcolor{globalrubric}{Global-Rubric} & .345$_{.332-.359}$ & .241$_{.219-.264}$ & .147$_{.120-.177}$ & .610$_{.595-.625}$ & .521$_{.489-.552}$ \\
 & \textcolor{globalrubric}{Global-Rubric-Auto} & .341$_{.331-.353}$ & .254$_{.229-.283}$ & .120$_{.104-.139}$ & .626$_{.610-.642}$ & .497$_{.460-.534}$ \\
 & \textcolor{globalrubric}{Global-Rubric-Tabular} & .303$_{.296-.310}$ & .184$_{.169-.200}$ & .091$_{.081-.102}$ & .589$_{.576-.603}$ & .497$_{.470-.523}$ \\
\midrule
\multirow{10}{*}{$n=\text{All}$} & Count-GBM & .371$_{.353-.388}$ & .274$_{.250-.302}$ & .218$_{.180-.255}$ & .570$_{.556-.584}$ & .582$_{.540-.624}$ \\
 & CLMBR-T & .430$_{.417-.444}$ & \cellcolor{green!30}.425$_{.387-.468}$ & .170$_{.145-.197}$ & .713$_{.699-.726}$ & .600$_{.559-.643}$ \\
 & NaiveText & .391$_{.377-.406}$ & .315$_{.283-.346}$ & .179$_{.151-.208}$ & .649$_{.634-.664}$ & \cellcolor{green!30}.609$_{.577-.641}$ \\
 & \textcolor{localrubric}{Local-Rubric} & .452$_{.439-.466}$ & .341$_{.310-.374}$ & .223$_{.194-.251}$ & .762$_{.748-.776}$ & .605$_{.571-.638}$ \\
 & \textcolor{localrubric}{Local-Rubric-Basic} & .431$_{.417-.445}$ & .333$_{.304-.368}$ & .203$_{.174-.234}$ & .731$_{.716-.744}$ & .594$_{.554-.633}$ \\
 & \textcolor{localrubric}{Local-Rubric-NoInterp} & .451$_{.437-.465}$ & .345$_{.315-.379}$ & .227$_{.198-.257}$ & .758$_{.744-.771}$ & .581$_{.537-.625}$ \\
 & \textcolor{globalrubric}{Global-Rubric-Blind} & .428$_{.415-.443}$ & .321$_{.292-.351}$ & .185$_{.155-.217}$ & .753$_{.739-.767}$ & .587$_{.556-.620}$ \\
 & \textcolor{globalrubric}{Global-Rubric} & \cellcolor{green!30}.459$_{.442-.478}$ & .339$_{.309-.371}$ & \cellcolor{green!30}.236$_{.200-.276}$ & .773$_{.760-.786}$ & .582$_{.550-.614}$ \\
 & \textcolor{globalrubric}{Global-Rubric-Auto} & .437$_{.422-.452}$ & .343$_{.312-.378}$ & .195$_{.164-.226}$ & .760$_{.745-.773}$ & .557$_{.518-.596}$ \\
 & \textcolor{globalrubric}{Global-Rubric-Tabular} & .445$_{.429-.461}$ & .329$_{.296-.363}$ & .208$_{.175-.243}$ & \cellcolor{green!30}.783$_{.769-.796}$ & .527$_{.486-.567}$ \\
\bottomrule
\end{tabular}

\end{table}

\clearpage

\subsection{Results using different text-embedding models as backbones in downstream training} \label{app:smaller_embedding}
\renewcommand{\arraystretch}{0.9}
\begin{table}[ht]
\centering
\small
\setlength{\tabcolsep}{3.2pt}
\caption{Performance with different text-embedding models ($n=\text{All}$). \textbf{Green}: best in entire column. \textbf{Blue}: best within a method across backbones. \textbf{Oper.:} Operational Outcomes, \textbf{New Dx:} Assignment of New Diagnosis, \textbf{Labs:} Anticipating Lab Results, \textbf{CXR:} Chest X-ray Findings.}
\vspace{5pt}
\label{tab:embedding_5backbone_auc_all_panels}

\vspace{0.35em}
\textbf{(A) AUROC} \\[0.4em]
\begin{tabular}{llccccc}
\toprule
\textbf{Method} & \textbf{Backbone} & \textbf{Overall} & \makecell{\textbf{Oper.}\\\textbf{(3)}} & \makecell{\textbf{New Dx}\\\textbf{(6)}} & \makecell{\textbf{Labs}\\\textbf{(5)}} & \makecell{\textbf{CXR}\\\textbf{(1)}} \\
\midrule
\multirow{5}{*}{NaiveText} & Mistral-7B & .612$_{.596-.626}$ & .656$_{.631-.681}$ & .647$_{.610-.681}$ & .574$_{.562-.585}$ & .453$_{.421-.483}$ \\
 & Llama3-8B & .594$_{.578-.609}$ & .660$_{.633-.687}$ & .619$_{.584-.654}$ & .552$_{.540-.563}$ & .463$_{.433-.494}$ \\
 & TE3-L & .674$_{.660-.689}$ & .741$_{.719-.761}$ & \cellcolor{bestblue}.713$_{.681-.744}$ & .610$_{.599-.621}$ & .564$_{.532-.596}$ \\
 & Qwen3-8B & \cellcolor{bestblue}.699$_{.684-.714}$ & \cellcolor{bestblue}.775$_{.754-.793}$ & .709$_{.674-.744}$ & \cellcolor{bestblue}.657$_{.646-.668}$ & \cellcolor{green!30}.616$_{.592-.640}$ \\
\cmidrule(lr){1-7}
\multirow{5}{*}{Local-Rubric} & Mistral-7B & .760$_{.746-.773}$ & \cellcolor{green!30}.804$_{.787-.820}$ & .762$_{.727-.794}$ & .768$_{.759-.778}$ & .569$_{.537-.598}$ \\
 & Llama3-8B & .761$_{.746-.775}$ & .793$_{.774-.810}$ & .764$_{.726-.796}$ & .772$_{.762-.780}$ & .591$_{.560-.621}$ \\
 & TE3-L & .762$_{.749-.774}$ & .798$_{.781-.815}$ & .762$_{.732-.790}$ & .782$_{.773-.790}$ & .553$_{.523-.585}$ \\
 & Qwen3-8B & \cellcolor{green!30}.772$_{.758-.784}$ & .802$_{.786-.818}$ & \cellcolor{green!30}.770$_{.738-.799}$ & \cellcolor{bestblue}.789$_{.780-.798}$ & \cellcolor{bestblue}.606$_{.583-.630}$ \\
\cmidrule(lr){1-7}
\multirow{5}{*}{Global-Rubric} & Mistral-7B & .725$_{.711-.739}$ & .762$_{.742-.782}$ & .692$_{.659-.724}$ & .775$_{.766-.785}$ & .557$_{.532-.583}$ \\
 & Llama3-8B & .736$_{.722-.750}$ & .764$_{.744-.784}$ & .715$_{.684-.747}$ & .777$_{.768-.787}$ & .565$_{.540-.590}$ \\
 & TE3-L & .738$_{.726-.750}$ & .764$_{.743-.784}$ & .714$_{.687-.742}$ & .788$_{.778-.796}$ & .561$_{.537-.584}$ \\
 & Qwen3-8B & \cellcolor{bestblue}.763$_{.748-.777}$ & \cellcolor{bestblue}.786$_{.768-.805}$ & \cellcolor{bestblue}.756$_{.723-.789}$ & \cellcolor{green!30}.791$_{.781-.800}$ & \cellcolor{bestblue}.594$_{.567-.619}$ \\
\bottomrule
\end{tabular}

\vspace{0.9em}

\textbf{(B) AUPRC} \\[0.4em]
\begin{tabular}{llccccc}
\toprule
\textbf{Method} & \textbf{Backbone} & \textbf{Overall} & \makecell{\textbf{Oper.}\\\textbf{(3)}} & \makecell{\textbf{New Dx}\\\textbf{(6)}} & \makecell{\textbf{Labs}\\\textbf{(5)}} & \makecell{\textbf{CXR}\\\textbf{(1)}} \\
\midrule
\multirow{5}{*}{NaiveText} & Mistral-7B & .316$_{.304-.328}$ & .244$_{.220-.274}$ & .124$_{.103-.147}$ & .561$_{.546-.576}$ & .457$_{.424-.489}$ \\
 & Llama3-8B & .310$_{.301-.321}$ & .241$_{.218-.269}$ & .116$_{.099-.133}$ & .549$_{.534-.563}$ & .494$_{.455-.532}$ \\
 & TE3-L & .351$_{.340-.364}$ & .282$_{.254-.311}$ & .145$_{.122-.173}$ & .601$_{.586-.615}$ & .546$_{.508-.584}$ \\
 & Qwen3-8B & \cellcolor{bestblue}.391$_{.377-.406}$ & \cellcolor{bestblue}.315$_{.283-.346}$ & \cellcolor{bestblue}.179$_{.151-.208}$ & \cellcolor{bestblue}.649$_{.634-.664}$ & \cellcolor{green!30}.609$_{.577-.641}$ \\
\cmidrule(lr){1-7}
\multirow{5}{*}{Local-Rubric} & Mistral-7B & .446$_{.430-.463}$ & \cellcolor{bestblue}.357$_{.326-.389}$ & .224$_{.188-.259}$ & .741$_{.727-.754}$ & .573$_{.533-.613}$ \\
 & Llama3-8B & .450$_{.433-.467}$ & .342$_{.310-.376}$ & \cellcolor{bestblue}.235$_{.199-.270}$ & .745$_{.731-.759}$ & .584$_{.546-.624}$ \\
 & TE3-L & .445$_{.431-.461}$ & .348$_{.315-.380}$ & .218$_{.187-.252}$ & .755$_{.741-.768}$ & .559$_{.518-.599}$ \\
 & Qwen3-8B & \cellcolor{bestblue}.452$_{.439-.466}$ & .341$_{.310-.374}$ & .223$_{.194-.251}$ & \cellcolor{bestblue}.762$_{.748-.776}$ & \cellcolor{bestblue}.605$_{.571-.638}$ \\
\cmidrule(lr){1-7}
\multirow{5}{*}{Global-Rubric} & Mistral-7B & .417$_{.404-.430}$ & .338$_{.307-.370}$ & .153$_{.129-.178}$ & .755$_{.742-.767}$ & .548$_{.516-.579}$ \\
 & Llama3-8B & .430$_{.417-.445}$ & .337$_{.307-.368}$ & .179$_{.153-.208}$ & .760$_{.748-.773}$ & .565$_{.533-.598}$ \\
 & TE3-L & .433$_{.419-.447}$ & \cellcolor{bestblue}.346$_{.311-.380}$ & .178$_{.153-.205}$ & .771$_{.758-.784}$ & .531$_{.502-.563}$ \\
 & Qwen3-8B & \cellcolor{green!30}.459$_{.442-.478}$ & .339$_{.309-.371}$ & \cellcolor{green!30}.236$_{.200-.276}$ & \cellcolor{green!30}.773$_{.760-.786}$ & \cellcolor{bestblue}.582$_{.550-.614}$ \\
\bottomrule
\end{tabular}
\end{table}
We evaluate four text embedding models for the downstream learning step,
keeping the classifier (logistic regression on frozen embeddings) fixed.

We compare
\textbf{Qwen3-Embedding-8B}~\citep{qwen3embedding},
two \textbf{LLM2Vec}~\citep{llm2vec} bidirectional adaptations of
\textbf{Llama-3-8B-Instruct}~\citep{llama3} and
\textbf{Mistral-7B-Instruct}~\citep{mistral7b},
and the proprietary
\textbf{text-embedding-3-large} model (\textbf{TE3-L}) accessed through the OpenAI
API~\citep{openai_te3l}. We use $n=\text{All}$ and compare the main textual representation-based methods: \textcolor{methodblue}{NaiveText}, \textcolor{methodblue}{Local-Rubric}, and \textcolor{methodblue}{Global-Rubric}.

As shown in Table~\ref{tab:embedding_5backbone_auc_all_panels}, 
Qwen3-Embedding-8B achieves the highest overall AUROC and AUPRC across methods compared. Within different task types, it continues to be either the best text-embedding model, or very close to top model. In the remainder of the paper, we report results for Qwen3-Embedding-8B model, unless otherwise stated.
\clearpage
\subsection{Sample-size sweep figures with confidence bands} \label{app:figs_with_ci}
\begin{figure}[ht]
    \centering
    \includegraphics[width=0.98\linewidth]{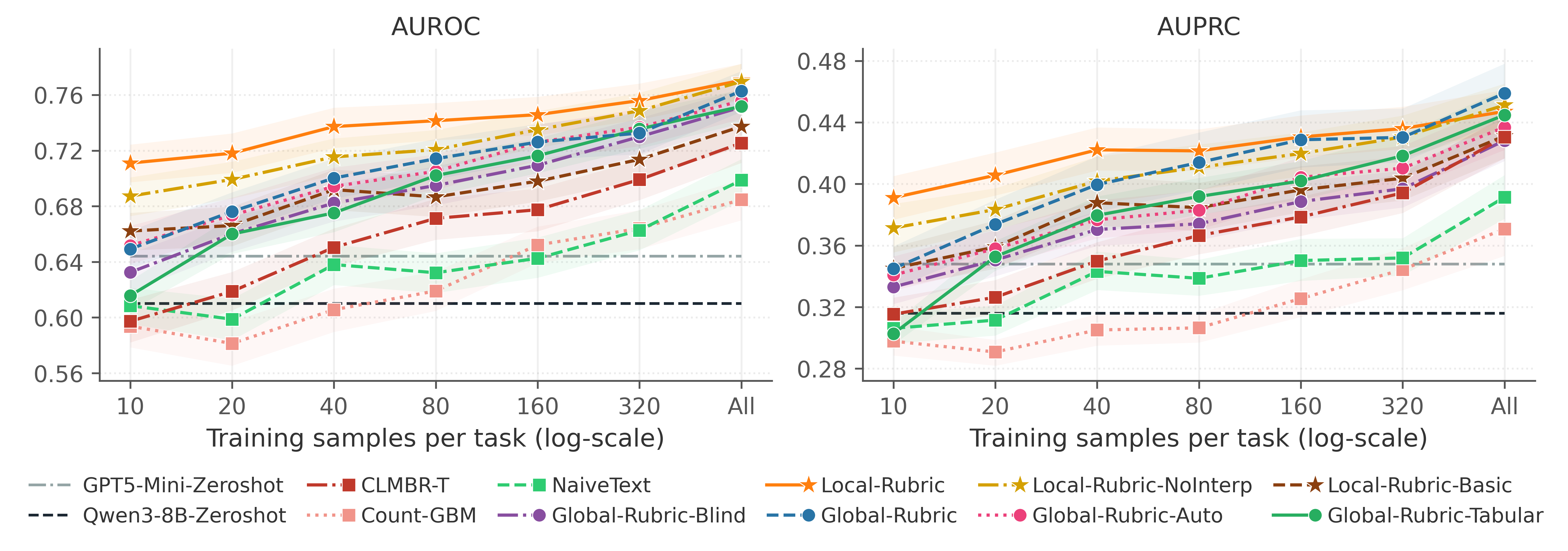}
    \caption{Average overall performance. Bands denote $95\%$ CI, bootstrapped over the test set.}
    \label{fig:fullsweep}
\end{figure}
\begin{figure}[ht]
    \centering
    \includegraphics[width=0.98\linewidth]{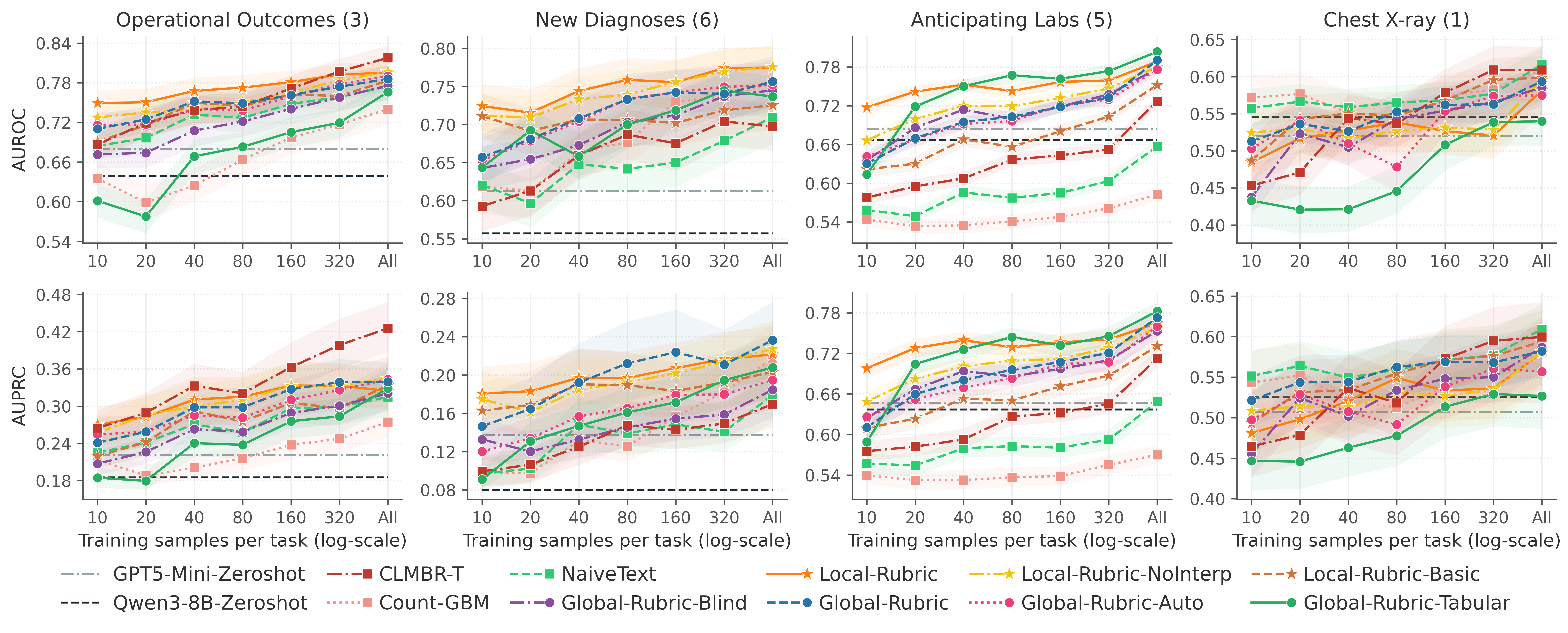}
    \caption{Average performance per task group. Bands denote $95\%$ CI, bootstrapped over the test set.}
    \label{fig:fullsweepbuckets}
\end{figure}

\subsection{Full EHRSHOT evaluation results for Global-Rubric-Tabular}
\label{app:grtab-full}

Evaluating \textcolor{methodblue}{Global-Rubric-Tabular} on the full EHRSHOT dataset without subsampling is feasible as there are no per-example LLM calls needed. \Cref{tab:grtab-full} reports results for the full dataset. The model achieves a mean AUROC of 0.770 and mean AUPRC of 0.312 across tasks. 
\begin{table*}[ht]
\centering
\small
\setlength{\tabcolsep}{3pt}
\renewcommand{\arraystretch}{0.88}
\caption{Full-dataset EHRSHOT results for \textcolor{forestgreen}{Global-Rubric-Tabular}.}
\label{tab:grtab-full}

\newcommand{\ehrci}[3]{$#1_{\scriptscriptstyle #2\!-\!#3}$}
\newcommand{\ehrbci}[3]{$\mathbf{#1}_{\scriptscriptstyle #2\!-\!#3}$}

\begin{minipage}[t]{0.49\textwidth}
\centering
\begin{tabular}{lcc}
\toprule
Task & AUROC & AUPRC \\
\midrule
\multicolumn{3}{l}{\textit{Operational Outcomes}} \\
ICU transfer & \ehrci{.800}{.752}{.846} & \ehrci{.195}{.130}{.277} \\
Long length of stay & \ehrci{.737}{.714}{.758} & \ehrci{.464}{.421}{.509} \\
30-day readmit & \ehrci{.757}{.725}{.787} & \ehrci{.357}{.300}{.419} \\
\textbf{Group Avg.} & \ehrbci{.765}{.745}{.786} & \ehrbci{.340}{.308}{.374} \\
\midrule

\multicolumn{3}{l}{\textit{Assignment of New Diagnoses}} \\
Acute MI & \ehrci{.765}{.726}{.803} & \ehrci{.183}{.140}{.225} \\
Lupus & \ehrci{.819}{.728}{.897} & \ehrci{.048}{.020}{.078} \\
Hyperlipidemia & \ehrci{.736}{.697}{.772} & \ehrci{.312}{.248}{.383} \\
Hypertension & \ehrci{.718}{.677}{.757} & \ehrci{.299}{.239}{.368} \\
Celiac disease & \ehrci{.654}{.540}{.772} & \ehrci{.098}{.016}{.190} \\
Pancreatic cancer & \ehrci{.859}{.798}{.915} & \ehrci{.378}{.239}{.518} \\
\textbf{Group Avg.} & \ehrbci{.759}{.729}{.790} & \ehrbci{.216}{.185}{.248} \\
\bottomrule
\end{tabular}
\end{minipage}
\hfill
\begin{minipage}[t]{0.49\textwidth}
\centering
\begin{tabular}{lcc}
\toprule
Task & AUROC & AUPRC \\
\midrule
\multicolumn{3}{l}{\textit{Anticipating Lab Results}} \\
Anemia & \ehrci{.810}{.806}{.815} & \ehrci{.361}{.351}{.372} \\
Hyponatremia & \ehrci{.815}{.811}{.818} & \ehrci{.560}{.552}{.568} \\
Thrombocytopenia & \ehrci{.885}{.881}{.888} & \ehrci{.571}{.558}{.582} \\
Hyperkalemia & \ehrci{.853}{.840}{.865} & \ehrci{.097}{.086}{.109} \\
Hypoglycemia & \ehrci{.751}{.733}{.768} & \ehrci{.031}{.026}{.037} \\
\textbf{Group Avg.} & \ehrbci{.823}{.818}{.827} & \ehrbci{.324}{.319}{.329} \\
\midrule

\multicolumn{3}{l}{\textit{Chest X-ray Findings}} \\
Chest X-ray & \ehrci{.584}{.572}{.597} & \ehrci{.743}{.731}{.756} \\
\textbf{Group Avg.} & \ehrbci{.584}{.571}{.597} & \ehrbci{.743}{.731}{.756} \\
\midrule

\textbf{Overall Avg. (15 tasks)}
& \ehrbci{.770}{.756}{.783}
& \ehrbci{.312}{.297}{.327} \\
\bottomrule
\end{tabular}
\end{minipage}
\end{table*}
\clearpage
\subsection{Downstream classifier hyperparameters} \label{app:downstream_hparams}
For all logistic-regression downstream classifiers (textual rubric variants, \textcolor{methodblue}{NaiveText}, and \textcolor{methodblue}{CLMBR-T}), we use scikit-learn's \texttt{LogisticRegression} with an L2 penalty. The inverse regularization strength $C$ is tuned, for each task and each training-set size $n$, on the validation split using negative log-likelihood, sweeping a log-spaced grid of 13 values, $C \in \texttt{np.logspace(-6,\,-1,\,13)}$ (i.e., $C\in\{10^{-6},\ldots,10^{-1}\}$). The maximum number of iterations is set high enough for convergence on all tasks. We enable class-balanced sample reweighting.

For \textcolor{methodblue}{Global-Rubric-Tabular}, we train an \texttt{xgboost.XGBClassifier} and tune the number of trees, maximum depth, learning rate, and minimum child weight on the validation split using negative log-likelihood. The grid is:
\begin{itemize}[leftmargin=1.5em]
    \setlength\itemsep{0pt}
    \item \texttt{n\_estimators} $\in \{50,\,100,\,300\}$
    \item \texttt{max\_depth} $\in \{2,\,3,\,5,\,7\}$
    \item \texttt{learning\_rate} $\in \{0.01,\,0.05\}$
    \item \texttt{min\_child\_weight} $\in \{1,\,5\}$
    \item \texttt{subsample} $= 0.8$ (fixed)
\end{itemize}
For \textcolor{methodblue}{Count-GBM}, we train a \texttt{lightgbm.LGBMClassifier}, and for \textcolor{methodblue}{CLMBR-T} a logistic head over embeddings, following the search procedure of \citet{wornow2023ehrshot}.
\subsection{Token counts of different textual representations} \label{app:token_counts}
\begin{table}[ht]
\centering
\caption{Token count statistics for the textual representations that are passed into the text embedding model prior to downstream learning across different methods. Statistics for each representation are computed after merging data from all 15 tasks and 3 splits in \Cref{tab:sample_counts}. Qwen3-8B tokenizer is used.}
\vspace{10pt}
\label{tab:token_stats}
\begin{tabular}{l r r r r r}
\toprule
Representation & Mean & Median & Min & Max & Std \\
\midrule
NaiveText & 5,980 & 7,769 & 131 & 8,267 & 2,648 \\
Local-Rubric & 853 & 844 & 492 & 1,430 & 116 \\
Global-Rubric & 1,283 & 1,193 & 318 & 3,136 & 452 \\
\bottomrule
\end{tabular}
\end{table}
\clearpage
\subsection{Per-task results} \label{app:pertask}
\begin{figure}[H]
    \centering
    \includegraphics[width=\linewidth]{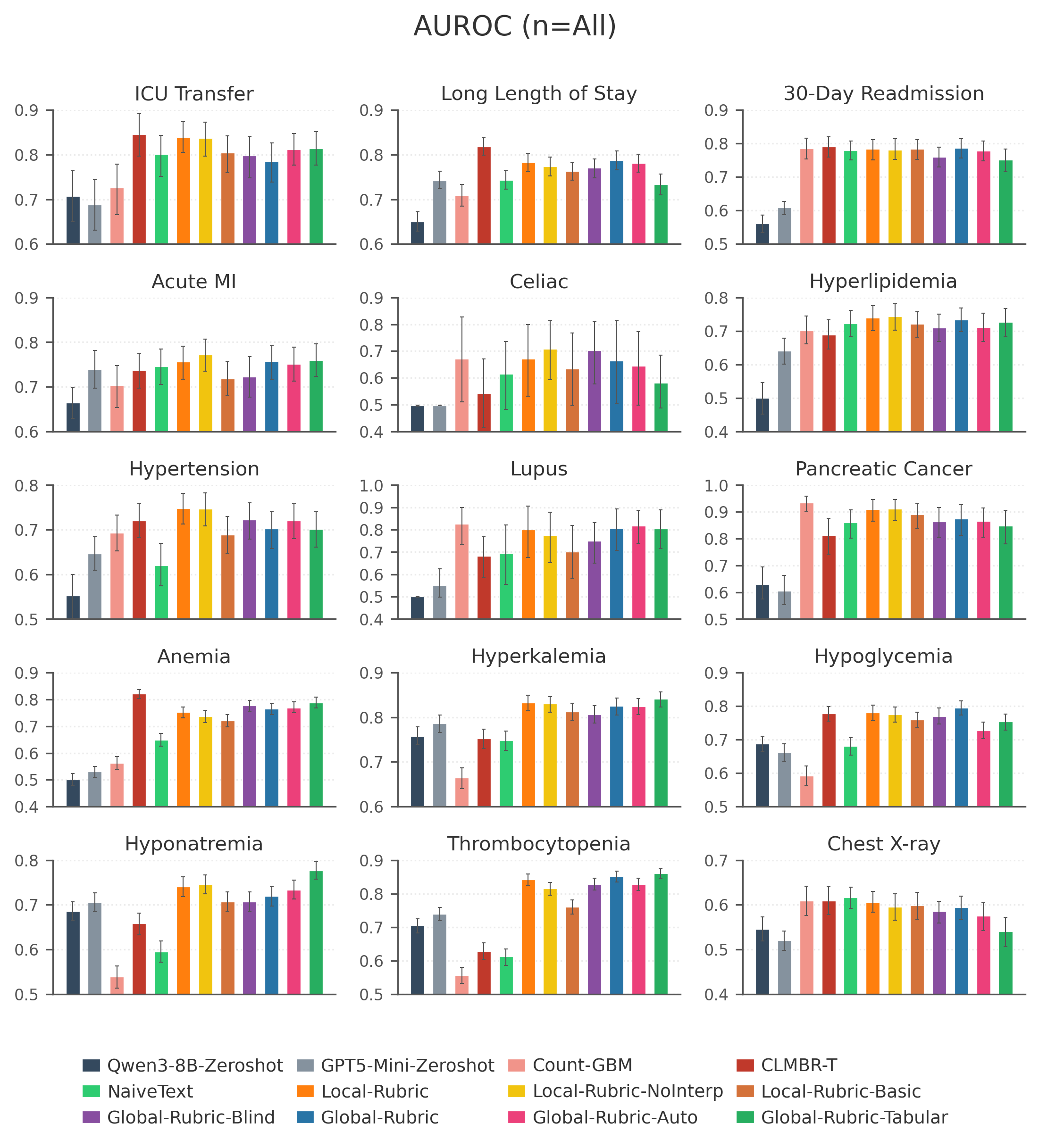}
    \caption{AUROC per-task, $n=\text{All}$}
    \label{fig:auroc_per_task}
\end{figure}

\begin{figure}[H]
    \centering
    \includegraphics[width=\linewidth]{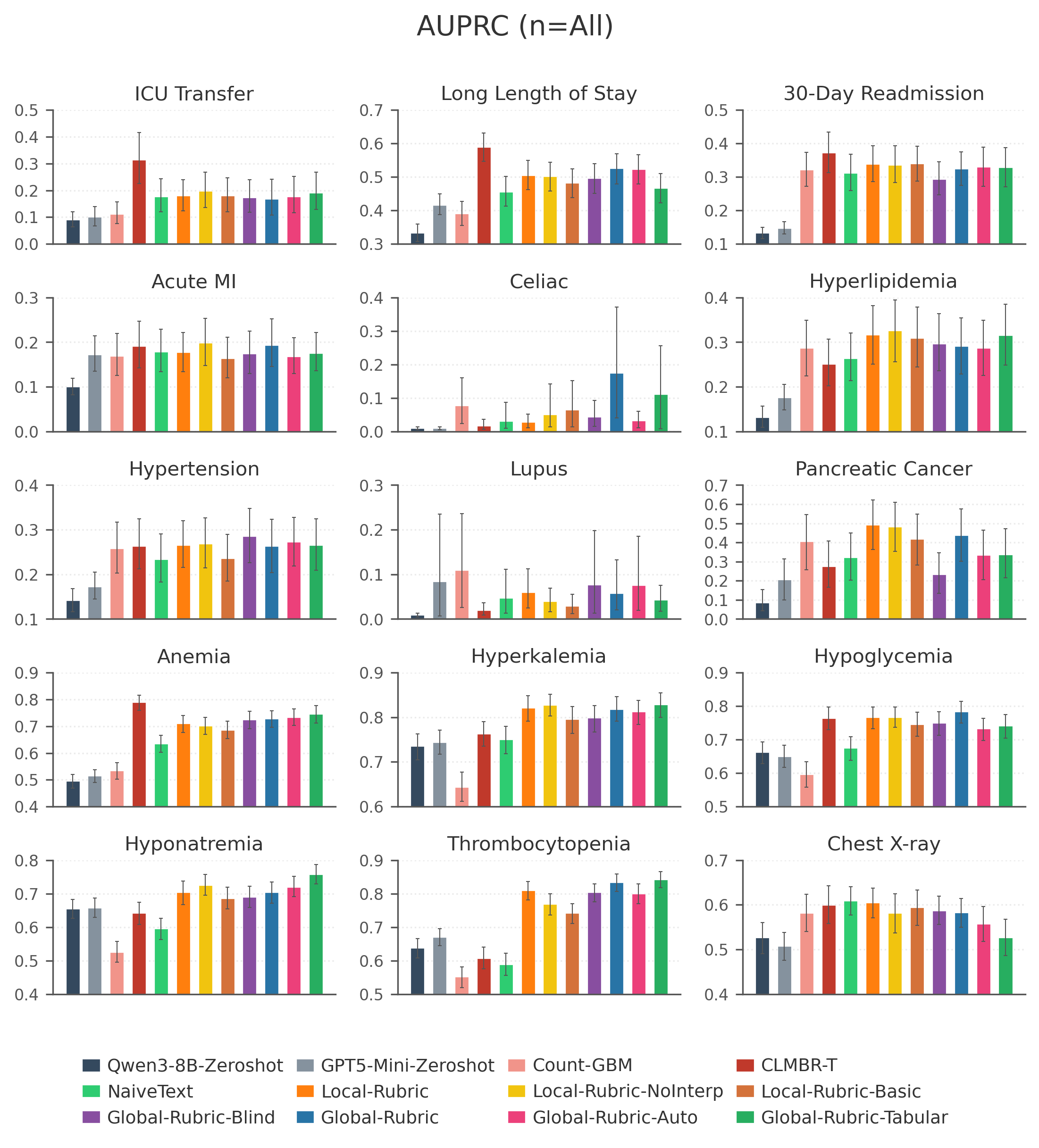}
    \caption{AUPRC per-task, $n=\text{All}$}
    \label{fig:auprc_per_task}
\end{figure}

\begin{figure}[H]
    \centering
    \includegraphics[width=\linewidth]{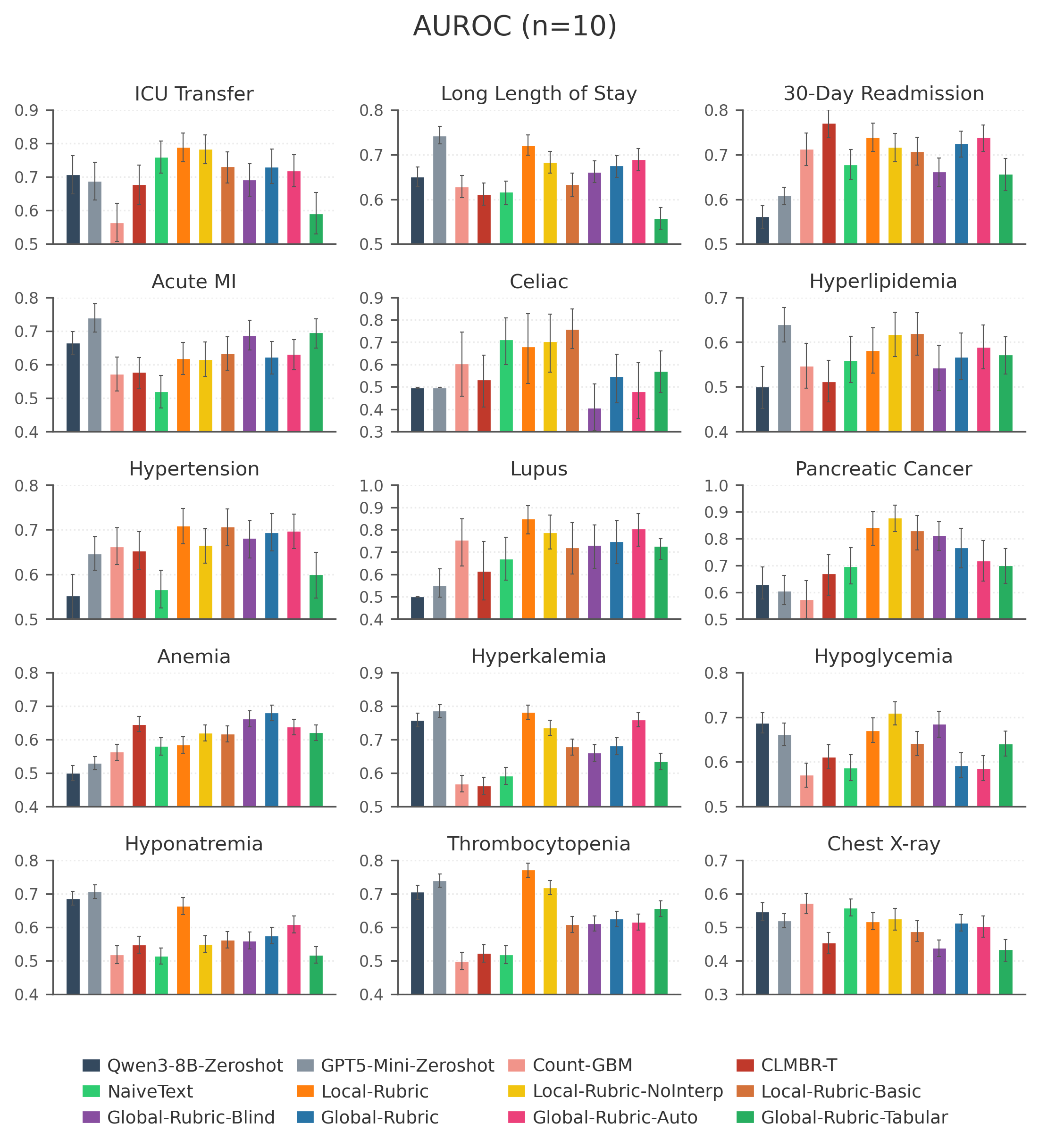}
    \caption{AUROC per-task, $n=10$}
    \label{fig:auroc_per_task_n40}
\end{figure}

\begin{figure}[H]
    \centering
    \includegraphics[width=\linewidth]{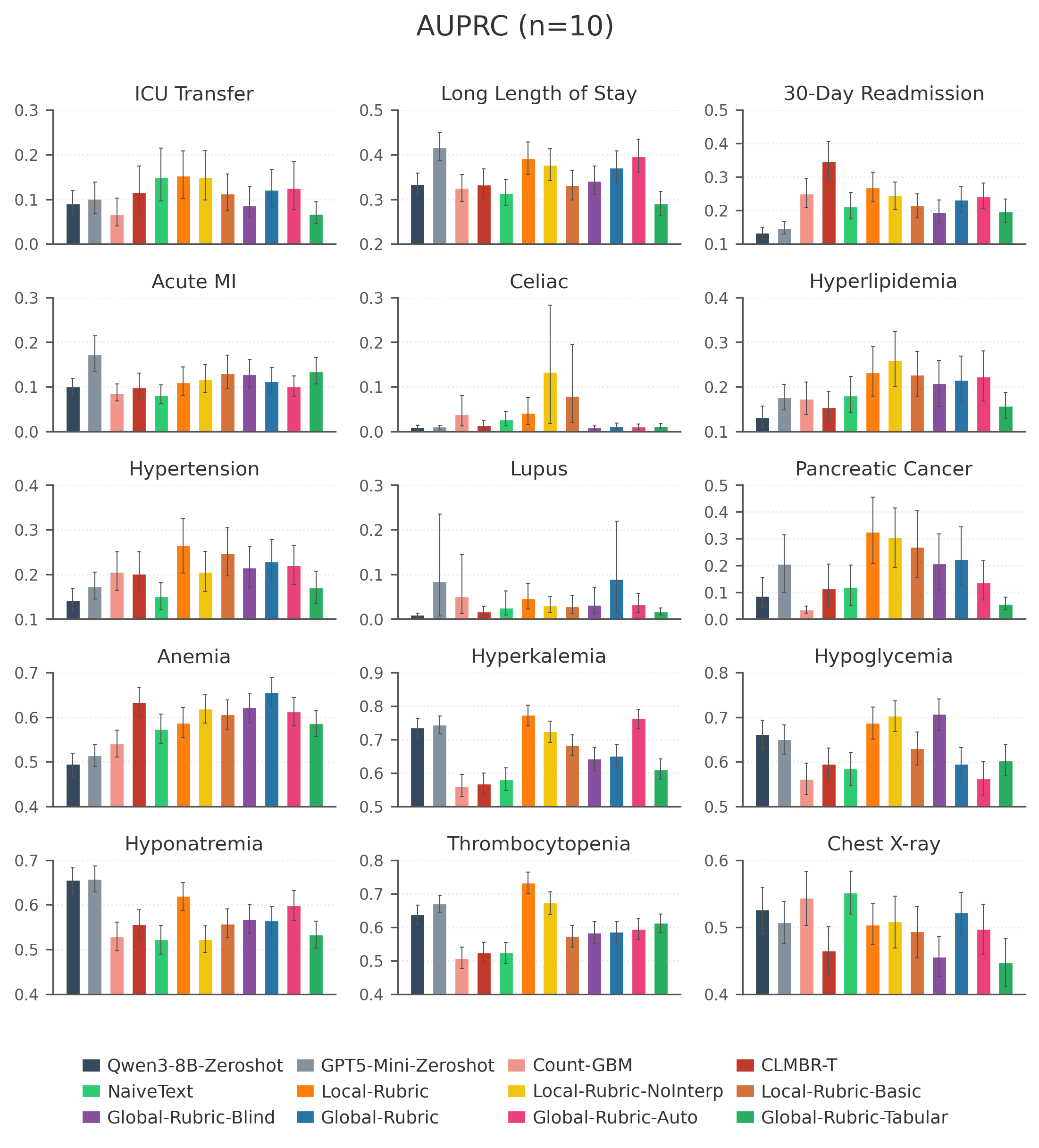}
    \caption{AUPRC per-task, $n=10$}
    \label{fig:auprc_per_task_n40}
\end{figure}

In this section, we report AUROC and AUPRC metrics separately for each task in Figures~\ref{fig:auroc_per_task}--\ref{fig:auprc_per_task_n40} and Tables~\ref{tab:embedding_task_guoicu}--\ref{tab:embedding_task_chexpert}. Qwen3-Embedding-8B is used as the text-embedding model. For some results with other embedding models, see Appendix~\ref{app:smaller_embedding}.
\newpage
\newpage
%

% --- Per-task: guo_icu ---
\providecolor{localrubric}{RGB}{190, 100, 25}
\providecolor{globalrubric}{RGB}{45, 90, 165}
\begin{table}[htbp]
\centering
\small
\renewcommand{\arraystretch}{0.85}
\caption{Guo ICU. AUROC and AUPRC with 95\% bootstrap CI. Best per column for each setting is highlighted. Embeddings are from Qwen3-8B; sample sizes refer to per-class training examples.}
\label{tab:embedding_task_guoicu}

\begin{tabular}{lcccc}
\toprule
 & \multicolumn{2}{c}{\textbf{AUROC}} & \multicolumn{2}{c}{\textbf{AUPRC}} \\
\cmidrule(lr){2-3} \cmidrule(lr){4-5}
Qwen3-8B-Zeroshot & \multicolumn{2}{c}{.707$_{.650-.764}$} & \multicolumn{2}{c}{.090$_{.064-.120}$} \\
GPT5-Mini-Zeroshot & \multicolumn{2}{c}{.688$_{.631-.744}$} & \multicolumn{2}{c}{.101$_{.068-.139}$} \\
\midrule
 & $n=10$ & $n=\text{All}$ & $n=10$ & $n=\text{All}$ \\
\cmidrule(lr){2-2} \cmidrule(lr){3-3} \cmidrule(lr){4-4} \cmidrule(lr){5-5}
Count-GBM & .564$_{.508-.622}$ & .726$_{.665-.779}$ & .066$_{.040-.103}$ & .112$_{.076-.157}$ \\
CLMBR-T & .678$_{.617-.736}$ & \cellcolor{green!30}.845$_{.797-.892}$ & .115$_{.069-.174}$ & \cellcolor{green!30}.314$_{.226-.416}$ \\
NaiveText & .759$_{.711-.807}$ & .801$_{.751-.843}$ & .149$_{.097-.214}$ & .177$_{.119-.243}$ \\
\textcolor{localrubric}{Local-Rubric} & \cellcolor{green!30}.789$_{.745-.832}$ & .839$_{.805-.873}$ & \cellcolor{green!30}.152$_{.102-.209}$ & .179$_{.124-.239}$ \\
\textcolor{localrubric}{Local-Rubric-Basic} & .731$_{.681-.775}$ & .804$_{.759-.842}$ & .113$_{.075-.156}$ & .179$_{.120-.247}$ \\
\textcolor{localrubric}{Local-Rubric-NoInterp} & .783$_{.739-.826}$ & .837$_{.797-.873}$ & .150$_{.098-.209}$ & .197$_{.136-.267}$ \\
\textcolor{globalrubric}{Global-Rubric-Blind} & .692$_{.642-.740}$ & .797$_{.748-.841}$ & .086$_{.059-.130}$ & .173$_{.118-.240}$ \\
\textcolor{globalrubric}{Global-Rubric} & .730$_{.680-.784}$ & .785$_{.738-.827}$ & .121$_{.085-.167}$ & .168$_{.108-.242}$ \\
\textcolor{globalrubric}{Global-Rubric-Auto} & .719$_{.671-.766}$ & .811$_{.777-.847}$ & .125$_{.077-.185}$ & .177$_{.116-.252}$ \\
\textcolor{globalrubric}{Global-Rubric-Tabular} & .591$_{.529-.653}$ & .814$_{.777-.851}$ & .067$_{.045-.095}$ & .191$_{.128-.268}$ \\
\bottomrule
\end{tabular}

\end{table}

% --- Per-task: guo_los ---
\providecolor{localrubric}{RGB}{190, 100, 25}
\providecolor{globalrubric}{RGB}{45, 90, 165}
\begin{table}[htbp]
\centering
\small
\renewcommand{\arraystretch}{0.85}
\caption{Guo length of stay. AUROC and AUPRC with 95\% bootstrap CI. Best per column for each setting is highlighted. Embeddings are from Qwen3-8B; sample sizes refer to per-class training examples.}
\label{tab:embedding_task_guolos}

\begin{tabular}{lcccc}
\toprule
 & \multicolumn{2}{c}{\textbf{AUROC}} & \multicolumn{2}{c}{\textbf{AUPRC}} \\
\cmidrule(lr){2-3} \cmidrule(lr){4-5}
Qwen3-8B-Zeroshot & \multicolumn{2}{c}{.650$_{.629-.672}$} & \multicolumn{2}{c}{.333$_{.308-.359}$} \\
GPT5-Mini-Zeroshot & \multicolumn{2}{c}{.742$_{.724-.763}$} & \multicolumn{2}{c}{.416$_{.387-.449}$} \\
\midrule
 & $n=10$ & $n=\text{All}$ & $n=10$ & $n=\text{All}$ \\
\cmidrule(lr){2-2} \cmidrule(lr){3-3} \cmidrule(lr){4-4} \cmidrule(lr){5-5}
Count-GBM & .628$_{.604-.654}$ & .709$_{.685-.733}$ & .325$_{.295-.355}$ & .390$_{.355-.426}$ \\
CLMBR-T & .611$_{.587-.637}$ & \cellcolor{green!30}.818$_{.799-.837}$ & .332$_{.298-.368}$ & \cellcolor{green!30}.589$_{.546-.632}$ \\
NaiveText & .616$_{.588-.641}$ & .743$_{.723-.765}$ & .313$_{.286-.344}$ & .455$_{.413-.502}$ \\
\textcolor{localrubric}{Local-Rubric} & \cellcolor{green!30}.721$_{.698-.744}$ & .783$_{.762-.803}$ & .391$_{.356-.428}$ & .505$_{.462-.550}$ \\
\textcolor{localrubric}{Local-Rubric-Basic} & .633$_{.606-.659}$ & .763$_{.743-.782}$ & .331$_{.299-.365}$ & .482$_{.438-.524}$ \\
\textcolor{localrubric}{Local-Rubric-NoInterp} & .683$_{.659-.707}$ & .773$_{.752-.794}$ & .376$_{.342-.414}$ & .502$_{.458-.544}$ \\
\textcolor{globalrubric}{Global-Rubric-Blind} & .661$_{.637-.686}$ & .770$_{.748-.791}$ & .341$_{.311-.374}$ & .496$_{.450-.540}$ \\
\textcolor{globalrubric}{Global-Rubric} & .675$_{.649-.698}$ & .787$_{.766-.809}$ & .370$_{.334-.408}$ & .526$_{.480-.570}$ \\
\textcolor{globalrubric}{Global-Rubric-Auto} & .689$_{.664-.714}$ & .781$_{.761-.801}$ & \cellcolor{green!30}.396$_{.361-.434}$ & .523$_{.479-.567}$ \\
\textcolor{globalrubric}{Global-Rubric-Tabular} & .557$_{.533-.582}$ & .734$_{.710-.757}$ & .290$_{.264-.317}$ & .467$_{.423-.510}$ \\
\bottomrule
\end{tabular}

\end{table}

% --- Per-task: guo_readmission ---
\providecolor{localrubric}{RGB}{190, 100, 25}
\providecolor{globalrubric}{RGB}{45, 90, 165}
\begin{table}[htbp]
\centering
\small
\renewcommand{\arraystretch}{0.85}
\caption{Guo readmission. AUROC and AUPRC with 95\% bootstrap CI. Best per column for each setting is highlighted. Embeddings are from Qwen3-8B; sample sizes refer to per-class training examples.}
\label{tab:embedding_task_guoreadmission}

\begin{tabular}{lcccc}
\toprule
 & \multicolumn{2}{c}{\textbf{AUROC}} & \multicolumn{2}{c}{\textbf{AUPRC}} \\
\cmidrule(lr){2-3} \cmidrule(lr){4-5}
Qwen3-8B-Zeroshot & \multicolumn{2}{c}{.561$_{.534-.586}$} & \multicolumn{2}{c}{.133$_{.116-.150}$} \\
GPT5-Mini-Zeroshot & \multicolumn{2}{c}{.609$_{.588-.627}$} & \multicolumn{2}{c}{.147$_{.130-.166}$} \\
\midrule
 & $n=10$ & $n=\text{All}$ & $n=10$ & $n=\text{All}$ \\
\cmidrule(lr){2-2} \cmidrule(lr){3-3} \cmidrule(lr){4-4} \cmidrule(lr){5-5}
Count-GBM & .712$_{.676-.748}$ & .785$_{.754-.816}$ & .249$_{.208-.295}$ & .321$_{.272-.373}$ \\
CLMBR-T & \cellcolor{green!30}.771$_{.738-.800}$ & \cellcolor{green!30}.791$_{.760-.819}$ & \cellcolor{green!30}.347$_{.289-.405}$ & \cellcolor{green!30}.373$_{.313-.434}$ \\
NaiveText & .678$_{.645-.711}$ & .780$_{.751-.808}$ & .212$_{.176-.253}$ & .312$_{.259-.368}$ \\
\textcolor{localrubric}{Local-Rubric} & .739$_{.707-.771}$ & .783$_{.752-.812}$ & .267$_{.226-.315}$ & .339$_{.286-.393}$ \\
\textcolor{localrubric}{Local-Rubric-Basic} & .707$_{.677-.739}$ & .783$_{.752-.811}$ & .214$_{.178-.250}$ & .339$_{.288-.392}$ \\
\textcolor{localrubric}{Local-Rubric-NoInterp} & .717$_{.684-.747}$ & .781$_{.752-.814}$ & .245$_{.203-.285}$ & .335$_{.283-.393}$ \\
\textcolor{globalrubric}{Global-Rubric-Blind} & .662$_{.628-.693}$ & .760$_{.729-.788}$ & .195$_{.162-.231}$ & .293$_{.246-.346}$ \\
\textcolor{globalrubric}{Global-Rubric} & .726$_{.695-.753}$ & .786$_{.757-.814}$ & .232$_{.195-.271}$ & .324$_{.275-.375}$ \\
\textcolor{globalrubric}{Global-Rubric-Auto} & .739$_{.708-.766}$ & .778$_{.748-.807}$ & .242$_{.206-.282}$ & .329$_{.272-.390}$ \\
\textcolor{globalrubric}{Global-Rubric-Tabular} & .656$_{.619-.692}$ & .751$_{.715-.783}$ & .196$_{.163-.233}$ & .328$_{.271-.387}$ \\
\bottomrule
\end{tabular}

\end{table}

% --- Per-task: new_acutemi ---
\providecolor{localrubric}{RGB}{190, 100, 25}
\providecolor{globalrubric}{RGB}{45, 90, 165}
\begin{table}[htbp]
\centering
\small
\renewcommand{\arraystretch}{0.85}
\caption{Acute MI. AUROC and AUPRC with 95\% bootstrap CI. Best per column for each setting is highlighted. Embeddings are from Qwen3-8B; sample sizes refer to per-class training examples.}
\label{tab:embedding_task_newacutemi}

\begin{tabular}{lcccc}
\toprule
 & \multicolumn{2}{c}{\textbf{AUROC}} & \multicolumn{2}{c}{\textbf{AUPRC}} \\
\cmidrule(lr){2-3} \cmidrule(lr){4-5}
Qwen3-8B-Zeroshot & \multicolumn{2}{c}{.664$_{.629-.698}$} & \multicolumn{2}{c}{.100$_{.082-.119}$} \\
GPT5-Mini-Zeroshot & \multicolumn{2}{c}{.739$_{.697-.782}$} & \multicolumn{2}{c}{.172$_{.135-.214}$} \\
\midrule
 & $n=10$ & $n=\text{All}$ & $n=10$ & $n=\text{All}$ \\
\cmidrule(lr){2-2} \cmidrule(lr){3-3} \cmidrule(lr){4-4} \cmidrule(lr){5-5}
Count-GBM & .572$_{.520-.622}$ & .704$_{.654-.748}$ & .086$_{.068-.106}$ & .169$_{.125-.219}$ \\
CLMBR-T & .577$_{.528-.621}$ & .737$_{.697-.775}$ & .099$_{.073-.131}$ & .191$_{.142-.247}$ \\
NaiveText & .520$_{.470-.568}$ & .746$_{.706-.785}$ & .081$_{.062-.105}$ & .179$_{.134-.229}$ \\
\textcolor{localrubric}{Local-Rubric} & .618$_{.570-.665}$ & .756$_{.717-.791}$ & .110$_{.081-.144}$ & .177$_{.134-.222}$ \\
\textcolor{localrubric}{Local-Rubric-Basic} & .633$_{.583-.683}$ & .718$_{.680-.757}$ & .129$_{.096-.171}$ & .163$_{.120-.211}$ \\
\textcolor{localrubric}{Local-Rubric-NoInterp} & .616$_{.564-.667}$ & \cellcolor{green!30}.772$_{.735-.807}$ & .116$_{.088-.150}$ & \cellcolor{green!30}.199$_{.148-.253}$ \\
\textcolor{globalrubric}{Global-Rubric-Blind} & .687$_{.643-.732}$ & .723$_{.677-.768}$ & .128$_{.098-.161}$ & .174$_{.130-.225}$ \\
\textcolor{globalrubric}{Global-Rubric} & .623$_{.572-.669}$ & .757$_{.717-.793}$ & .112$_{.084-.144}$ & .194$_{.146-.253}$ \\
\textcolor{globalrubric}{Global-Rubric-Auto} & .631$_{.585-.675}$ & .751$_{.713-.789}$ & .100$_{.079-.125}$ & .168$_{.130-.211}$ \\
\textcolor{globalrubric}{Global-Rubric-Tabular} & \cellcolor{green!30}.695$_{.649-.737}$ & .760$_{.723-.797}$ & \cellcolor{green!30}.134$_{.107-.165}$ & .176$_{.136-.221}$ \\
\bottomrule
\end{tabular}

\end{table}

% --- Per-task: new_celiac ---
\providecolor{localrubric}{RGB}{190, 100, 25}
\providecolor{globalrubric}{RGB}{45, 90, 165}
\begin{table}[htbp]
\centering
\small
\renewcommand{\arraystretch}{0.85}
\caption{Celiac disease. AUROC and AUPRC with 95\% bootstrap CI. Best per column for each setting is highlighted. Embeddings are from Qwen3-8B; sample sizes refer to per-class training examples.}
\label{tab:embedding_task_newceliac}

\begin{tabular}{lcccc}
\toprule
 & \multicolumn{2}{c}{\textbf{AUROC}} & \multicolumn{2}{c}{\textbf{AUPRC}} \\
\cmidrule(lr){2-3} \cmidrule(lr){4-5}
Qwen3-8B-Zeroshot & \multicolumn{2}{c}{.497$_{.495-.498}$} & \multicolumn{2}{c}{.009$_{.005-.014}$} \\
GPT5-Mini-Zeroshot & \multicolumn{2}{c}{.497$_{.496-.499}$} & \multicolumn{2}{c}{.010$_{.005-.014}$} \\
\midrule
 & $n=10$ & $n=\text{All}$ & $n=10$ & $n=\text{All}$ \\
\cmidrule(lr){2-2} \cmidrule(lr){3-3} \cmidrule(lr){4-4} \cmidrule(lr){5-5}
Count-GBM & .603$_{.457-.745}$ & .671$_{.510-.828}$ & .038$_{.012-.080}$ & .077$_{.024-.160}$ \\
CLMBR-T & .532$_{.409-.642}$ & .543$_{.415-.671}$ & .013$_{.007-.025}$ & .017$_{.007-.036}$ \\
NaiveText & .712$_{.599-.810}$ & .614$_{.482-.736}$ & .026$_{.012-.044}$ & .030$_{.009-.087}$ \\
\textcolor{localrubric}{Local-Rubric} & .680$_{.515-.828}$ & .670$_{.532-.799}$ & .041$_{.016-.076}$ & .028$_{.012-.052}$ \\
\textcolor{localrubric}{Local-Rubric-Basic} & \cellcolor{green!30}.757$_{.671-.849}$ & .634$_{.497-.768}$ & .079$_{.020-.195}$ & .064$_{.014-.152}$ \\
\textcolor{localrubric}{Local-Rubric-NoInterp} & .704$_{.567-.826}$ & \cellcolor{green!30}.708$_{.594-.813}$ & \cellcolor{green!30}.132$_{.018-.283}$ & .050$_{.014-.141}$ \\
\textcolor{globalrubric}{Global-Rubric-Blind} & .406$_{.304-.514}$ & .702$_{.578-.810}$ & .009$_{.005-.013}$ & .043$_{.015-.092}$ \\
\textcolor{globalrubric}{Global-Rubric} & .546$_{.428-.646}$ & .663$_{.505-.814}$ & .012$_{.007-.019}$ & \cellcolor{green!30}.174$_{.040-.372}$ \\
\textcolor{globalrubric}{Global-Rubric-Auto} & .480$_{.359-.609}$ & .644$_{.499-.774}$ & .011$_{.006-.017}$ & .032$_{.012-.060}$ \\
\textcolor{globalrubric}{Global-Rubric-Tabular} & .570$_{.474-.660}$ & .582$_{.487-.686}$ & .012$_{.007-.017}$ & .111$_{.008-.255}$ \\
\bottomrule
\end{tabular}

\end{table}

% --- Per-task: new_hyperlipidemia ---
\providecolor{localrubric}{RGB}{190, 100, 25}
\providecolor{globalrubric}{RGB}{45, 90, 165}
\begin{table}[htbp]
\centering
\small
\renewcommand{\arraystretch}{0.85}
\caption{Hyperlipidemia. AUROC and AUPRC with 95\% bootstrap CI. Best per column for each setting is highlighted. Embeddings are from Qwen3-8B; sample sizes refer to per-class training examples.}
\label{tab:embedding_task_newhyperlipidemia}

\begin{tabular}{lcccc}
\toprule
 & \multicolumn{2}{c}{\textbf{AUROC}} & \multicolumn{2}{c}{\textbf{AUPRC}} \\
\cmidrule(lr){2-3} \cmidrule(lr){4-5}
Qwen3-8B-Zeroshot & \multicolumn{2}{c}{.500$_{.452-.546}$} & \multicolumn{2}{c}{.132$_{.109-.157}$} \\
GPT5-Mini-Zeroshot & \multicolumn{2}{c}{.640$_{.601-.678}$} & \multicolumn{2}{c}{.176$_{.148-.205}$} \\
\midrule
 & $n=10$ & $n=\text{All}$ & $n=10$ & $n=\text{All}$ \\
\cmidrule(lr){2-2} \cmidrule(lr){3-3} \cmidrule(lr){4-4} \cmidrule(lr){5-5}
Count-GBM & .546$_{.496-.597}$ & .702$_{.662-.745}$ & .172$_{.138-.211}$ & .287$_{.225-.349}$ \\
CLMBR-T & .512$_{.466-.560}$ & .689$_{.647-.733}$ & .153$_{.122-.189}$ & .251$_{.202-.307}$ \\
NaiveText & .559$_{.510-.613}$ & .722$_{.684-.762}$ & .180$_{.142-.223}$ & .263$_{.213-.320}$ \\
\textcolor{localrubric}{Local-Rubric} & .581$_{.531-.632}$ & .740$_{.701-.776}$ & .232$_{.179-.291}$ & .316$_{.251-.382}$ \\
\textcolor{localrubric}{Local-Rubric-Basic} & \cellcolor{green!30}.619$_{.571-.666}$ & .721$_{.681-.758}$ & .226$_{.179-.280}$ & .309$_{.245-.379}$ \\
\textcolor{localrubric}{Local-Rubric-NoInterp} & .617$_{.568-.667}$ & \cellcolor{green!30}.743$_{.703-.782}$ & \cellcolor{green!30}.259$_{.200-.323}$ & \cellcolor{green!30}.326$_{.256-.394}$ \\
\textcolor{globalrubric}{Global-Rubric-Blind} & .542$_{.491-.593}$ & .710$_{.668-.750}$ & .208$_{.156-.259}$ & .297$_{.236-.364}$ \\
\textcolor{globalrubric}{Global-Rubric} & .567$_{.516-.620}$ & .734$_{.698-.769}$ & .215$_{.165-.269}$ & .291$_{.228-.355}$ \\
\textcolor{globalrubric}{Global-Rubric-Auto} & .589$_{.540-.639}$ & .711$_{.669-.753}$ & .222$_{.168-.280}$ & .286$_{.225-.349}$ \\
\textcolor{globalrubric}{Global-Rubric-Tabular} & .572$_{.529-.612}$ & .727$_{.684-.768}$ & .157$_{.129-.187}$ & .316$_{.248-.385}$ \\
\bottomrule
\end{tabular}

\end{table}

% --- Per-task: new_hypertension ---
\providecolor{localrubric}{RGB}{190, 100, 25}
\providecolor{globalrubric}{RGB}{45, 90, 165}
\begin{table}[htbp]
\centering
\small
\renewcommand{\arraystretch}{0.85}
\caption{Hypertension. AUROC and AUPRC with 95\% bootstrap CI. Best per column for each setting is highlighted. Embeddings are from Qwen3-8B; sample sizes refer to per-class training examples.}
\label{tab:embedding_task_newhypertension}

\begin{tabular}{lcccc}
\toprule
 & \multicolumn{2}{c}{\textbf{AUROC}} & \multicolumn{2}{c}{\textbf{AUPRC}} \\
\cmidrule(lr){2-3} \cmidrule(lr){4-5}
Qwen3-8B-Zeroshot & \multicolumn{2}{c}{.552$_{.502-.600}$} & \multicolumn{2}{c}{.142$_{.116-.168}$} \\
GPT5-Mini-Zeroshot & \multicolumn{2}{c}{.646$_{.609-.684}$} & \multicolumn{2}{c}{.172$_{.145-.205}$} \\
\midrule
 & $n=10$ & $n=\text{All}$ & $n=10$ & $n=\text{All}$ \\
\cmidrule(lr){2-2} \cmidrule(lr){3-3} \cmidrule(lr){4-4} \cmidrule(lr){5-5}
Count-GBM & .662$_{.622-.704}$ & .693$_{.653-.732}$ & .206$_{.164-.251}$ & .258$_{.203-.317}$ \\
CLMBR-T & .653$_{.611-.696}$ & .721$_{.682-.759}$ & .201$_{.160-.251}$ & .263$_{.213-.324}$ \\
NaiveText & .566$_{.525-.609}$ & .620$_{.574-.669}$ & .150$_{.121-.182}$ & .234$_{.183-.291}$ \\
\textcolor{localrubric}{Local-Rubric} & \cellcolor{green!30}.709$_{.669-.747}$ & \cellcolor{green!30}.747$_{.713-.782}$ & \cellcolor{green!30}.265$_{.203-.325}$ & .265$_{.215-.320}$ \\
\textcolor{localrubric}{Local-Rubric-Basic} & .707$_{.664-.747}$ & .688$_{.646-.730}$ & .247$_{.196-.305}$ & .236$_{.185-.290}$ \\
\textcolor{localrubric}{Local-Rubric-NoInterp} & .665$_{.625-.702}$ & .746$_{.709-.782}$ & .205$_{.162-.251}$ & .268$_{.215-.326}$ \\
\textcolor{globalrubric}{Global-Rubric-Blind} & .681$_{.637-.721}$ & .722$_{.679-.761}$ & .214$_{.170-.262}$ & \cellcolor{green!30}.285$_{.226-.347}$ \\
\textcolor{globalrubric}{Global-Rubric} & .693$_{.653-.736}$ & .702$_{.658-.742}$ & .228$_{.186-.278}$ & .263$_{.204-.323}$ \\
\textcolor{globalrubric}{Global-Rubric-Auto} & .696$_{.658-.735}$ & .721$_{.680-.759}$ & .220$_{.178-.265}$ & .272$_{.219-.327}$ \\
\textcolor{globalrubric}{Global-Rubric-Tabular} & .600$_{.547-.649}$ & .701$_{.661-.742}$ & .171$_{.135-.208}$ & .266$_{.209-.324}$ \\
\bottomrule
\end{tabular}

\end{table}

% --- Per-task: new_lupus ---
\providecolor{localrubric}{RGB}{190, 100, 25}
\providecolor{globalrubric}{RGB}{45, 90, 165}
\begin{table}[htbp]
\centering
\small
\renewcommand{\arraystretch}{0.85}
\caption{Lupus. AUROC and AUPRC with 95\% bootstrap CI. Best per column for each setting is highlighted. Embeddings are from Qwen3-8B; sample sizes refer to per-class training examples.}
\label{tab:embedding_task_newlupus}

\begin{tabular}{lcccc}
\toprule
 & \multicolumn{2}{c}{\textbf{AUROC}} & \multicolumn{2}{c}{\textbf{AUPRC}} \\
\cmidrule(lr){2-3} \cmidrule(lr){4-5}
Qwen3-8B-Zeroshot & \multicolumn{2}{c}{.500$_{.499-.500}$} & \multicolumn{2}{c}{.009$_{.005-.013}$} \\
GPT5-Mini-Zeroshot & \multicolumn{2}{c}{.550$_{.499-.625}$} & \multicolumn{2}{c}{.084$_{.007-.235}$} \\
\midrule
 & $n=10$ & $n=\text{All}$ & $n=10$ & $n=\text{All}$ \\
\cmidrule(lr){2-2} \cmidrule(lr){3-3} \cmidrule(lr){4-4} \cmidrule(lr){5-5}
Count-GBM & .754$_{.637-.849}$ & \cellcolor{green!30}.826$_{.734-.901}$ & .051$_{.012-.144}$ & \cellcolor{green!30}.109$_{.026-.236}$ \\
CLMBR-T & .614$_{.486-.747}$ & .681$_{.588-.768}$ & .016$_{.008-.028}$ & .020$_{.009-.037}$ \\
NaiveText & .670$_{.575-.767}$ & .695$_{.556-.822}$ & .025$_{.009-.063}$ & .047$_{.013-.112}$ \\
\textcolor{localrubric}{Local-Rubric} & \cellcolor{green!30}.849$_{.780-.907}$ & .801$_{.676-.906}$ & .046$_{.022-.079}$ & .060$_{.024-.113}$ \\
\textcolor{localrubric}{Local-Rubric-Basic} & .720$_{.602-.831}$ & .701$_{.584-.819}$ & .028$_{.012-.054}$ & .029$_{.012-.056}$ \\
\textcolor{localrubric}{Local-Rubric-NoInterp} & .788$_{.713-.865}$ & .775$_{.652-.879}$ & .030$_{.015-.051}$ & .040$_{.017-.069}$ \\
\textcolor{globalrubric}{Global-Rubric-Blind} & .731$_{.627-.821}$ & .750$_{.650-.831}$ & .031$_{.012-.071}$ & .076$_{.013-.198}$ \\
\textcolor{globalrubric}{Global-Rubric} & .747$_{.648-.840}$ & .807$_{.708-.892}$ & \cellcolor{green!30}.089$_{.016-.220}$ & .058$_{.020-.132}$ \\
\textcolor{globalrubric}{Global-Rubric-Auto} & .804$_{.727-.872}$ & .818$_{.739-.886}$ & .033$_{.015-.058}$ & .075$_{.020-.186}$ \\
\textcolor{globalrubric}{Global-Rubric-Tabular} & .727$_{.668-.760}$ & .804$_{.716-.889}$ & .017$_{.009-.025}$ & .043$_{.017-.076}$ \\
\bottomrule
\end{tabular}

\end{table}

% --- Per-task: new_pancan ---
\providecolor{localrubric}{RGB}{190, 100, 25}
\providecolor{globalrubric}{RGB}{45, 90, 165}
\begin{table}[htbp]
\centering
\small
\renewcommand{\arraystretch}{0.85}
\caption{Pancreatic cancer. AUROC and AUPRC with 95\% bootstrap CI. Best per column for each setting is highlighted. Embeddings are from Qwen3-8B; sample sizes refer to per-class training examples.}
\label{tab:embedding_task_newpancan}

\begin{tabular}{lcccc}
\toprule
 & \multicolumn{2}{c}{\textbf{AUROC}} & \multicolumn{2}{c}{\textbf{AUPRC}} \\
\cmidrule(lr){2-3} \cmidrule(lr){4-5}
Qwen3-8B-Zeroshot & \multicolumn{2}{c}{.630$_{.574-.694}$} & \multicolumn{2}{c}{.086$_{.043-.155}$} \\
GPT5-Mini-Zeroshot & \multicolumn{2}{c}{.604$_{.554-.662}$} & \multicolumn{2}{c}{.205$_{.099-.314}$} \\
\midrule
 & $n=10$ & $n=\text{All}$ & $n=10$ & $n=\text{All}$ \\
\cmidrule(lr){2-2} \cmidrule(lr){3-3} \cmidrule(lr){4-4} \cmidrule(lr){5-5}
Count-GBM & .574$_{.503-.644}$ & \cellcolor{green!30}.933$_{.902-.959}$ & .034$_{.023-.049}$ & .406$_{.258-.546}$ \\
CLMBR-T & .669$_{.589-.740}$ & .812$_{.741-.876}$ & .114$_{.046-.205}$ & .276$_{.165-.407}$ \\
NaiveText & .697$_{.630-.766}$ & .860$_{.803-.908}$ & .120$_{.051-.201}$ & .321$_{.202-.449}$ \\
\textcolor{localrubric}{Local-Rubric} & .843$_{.776-.900}$ & .909$_{.865-.947}$ & \cellcolor{green!30}.325$_{.206-.454}$ & \cellcolor{green!30}.491$_{.362-.622}$ \\
\textcolor{localrubric}{Local-Rubric-Basic} & .829$_{.757-.886}$ & .889$_{.836-.932}$ & .268$_{.154-.404}$ & .417$_{.283-.548}$ \\
\textcolor{localrubric}{Local-Rubric-NoInterp} & \cellcolor{green!30}.878$_{.826-.924}$ & .911$_{.866-.946}$ & .306$_{.192-.414}$ & .482$_{.353-.611}$ \\
\textcolor{globalrubric}{Global-Rubric-Blind} & .812$_{.755-.864}$ & .864$_{.805-.916}$ & .207$_{.109-.318}$ & .232$_{.134-.346}$ \\
\textcolor{globalrubric}{Global-Rubric} & .767$_{.691-.839}$ & .874$_{.813-.928}$ & .223$_{.127-.344}$ & .438$_{.301-.575}$ \\
\textcolor{globalrubric}{Global-Rubric-Auto} & .718$_{.641-.794}$ & .866$_{.806-.915}$ & .136$_{.071-.217}$ & .335$_{.205-.464}$ \\
\textcolor{globalrubric}{Global-Rubric-Tabular} & .699$_{.633-.763}$ & .847$_{.780-.906}$ & .056$_{.035-.081}$ & .337$_{.217-.473}$ \\
\bottomrule
\end{tabular}

\end{table}

% --- Per-task: lab_anemia ---
\providecolor{localrubric}{RGB}{190, 100, 25}
\providecolor{globalrubric}{RGB}{45, 90, 165}
\begin{table}[htbp]
\centering
\small
\renewcommand{\arraystretch}{0.85}
\caption{Anemia. AUROC and AUPRC with 95\% bootstrap CI. Best per column for each setting is highlighted. Embeddings are from Qwen3-8B; sample sizes refer to per-class training examples.}
\label{tab:embedding_task_labanemia}

\begin{tabular}{lcccc}
\toprule
 & \multicolumn{2}{c}{\textbf{AUROC}} & \multicolumn{2}{c}{\textbf{AUPRC}} \\
\cmidrule(lr){2-3} \cmidrule(lr){4-5}
Qwen3-8B-Zeroshot & \multicolumn{2}{c}{.500$_{.478-.523}$} & \multicolumn{2}{c}{.495$_{.468-.520}$} \\
GPT5-Mini-Zeroshot & \multicolumn{2}{c}{.530$_{.510-.549}$} & \multicolumn{2}{c}{.514$_{.490-.538}$} \\
\midrule
 & $n=10$ & $n=\text{All}$ & $n=10$ & $n=\text{All}$ \\
\cmidrule(lr){2-2} \cmidrule(lr){3-3} \cmidrule(lr){4-4} \cmidrule(lr){5-5}
Count-GBM & .563$_{.538-.587}$ & .562$_{.537-.587}$ & .540$_{.511-.571}$ & .533$_{.502-.563}$ \\
CLMBR-T & .646$_{.624-.669}$ & \cellcolor{green!30}.821$_{.803-.837}$ & .634$_{.602-.668}$ & \cellcolor{green!30}.788$_{.759-.816}$ \\
NaiveText & .581$_{.554-.605}$ & .649$_{.625-.673}$ & .573$_{.541-.607}$ & .634$_{.602-.667}$ \\
\textcolor{localrubric}{Local-Rubric} & .584$_{.559-.609}$ & .752$_{.731-.773}$ & .588$_{.554-.622}$ & .710$_{.677-.740}$ \\
\textcolor{localrubric}{Local-Rubric-Basic} & .618$_{.593-.641}$ & .721$_{.698-.743}$ & .606$_{.573-.639}$ & .686$_{.653-.719}$ \\
\textcolor{localrubric}{Local-Rubric-NoInterp} & .620$_{.596-.644}$ & .736$_{.714-.758}$ & .619$_{.587-.650}$ & .701$_{.669-.733}$ \\
\textcolor{globalrubric}{Global-Rubric-Blind} & .662$_{.638-.686}$ & .776$_{.756-.797}$ & .622$_{.587-.652}$ & .724$_{.691-.757}$ \\
\textcolor{globalrubric}{Global-Rubric} & \cellcolor{green!30}.680$_{.657-.702}$ & .764$_{.744-.785}$ & \cellcolor{green!30}.656$_{.623-.689}$ & .728$_{.696-.758}$ \\
\textcolor{globalrubric}{Global-Rubric-Auto} & .638$_{.615-.661}$ & .769$_{.750-.790}$ & .613$_{.582-.644}$ & .734$_{.703-.764}$ \\
\textcolor{globalrubric}{Global-Rubric-Tabular} & .621$_{.598-.644}$ & .787$_{.768-.809}$ & .586$_{.558-.614}$ & .745$_{.712-.776}$ \\
\bottomrule
\end{tabular}

\end{table}

% --- Per-task: lab_hyperkalemia ---
\providecolor{localrubric}{RGB}{190, 100, 25}
\providecolor{globalrubric}{RGB}{45, 90, 165}
\begin{table}[htbp]
\centering
\small
\renewcommand{\arraystretch}{0.85}
\caption{Hyperkalemia. AUROC and AUPRC with 95\% bootstrap CI. Best per column for each setting is highlighted. Embeddings are from Qwen3-8B; sample sizes refer to per-class training examples.}
\label{tab:embedding_task_labhyperkalemia}

\begin{tabular}{lcccc}
\toprule
 & \multicolumn{2}{c}{\textbf{AUROC}} & \multicolumn{2}{c}{\textbf{AUPRC}} \\
\cmidrule(lr){2-3} \cmidrule(lr){4-5}
Qwen3-8B-Zeroshot & \multicolumn{2}{c}{.758$_{.738-.779}$} & \multicolumn{2}{c}{.735$_{.705-.763}$} \\
GPT5-Mini-Zeroshot & \multicolumn{2}{c}{.786$_{.766-.805}$} & \multicolumn{2}{c}{.744$_{.717-.771}$} \\
\midrule
 & $n=10$ & $n=\text{All}$ & $n=10$ & $n=\text{All}$ \\
\cmidrule(lr){2-2} \cmidrule(lr){3-3} \cmidrule(lr){4-4} \cmidrule(lr){5-5}
Count-GBM & .568$_{.543-.593}$ & .665$_{.640-.687}$ & .561$_{.530-.595}$ & .644$_{.612-.677}$ \\
CLMBR-T & .562$_{.536-.588}$ & .752$_{.730-.774}$ & .568$_{.535-.601}$ & .763$_{.735-.791}$ \\
NaiveText & .592$_{.566-.617}$ & .748$_{.726-.769}$ & .581$_{.548-.616}$ & .750$_{.719-.779}$ \\
\textcolor{localrubric}{Local-Rubric} & \cellcolor{green!30}.782$_{.761-.803}$ & .832$_{.814-.850}$ & \cellcolor{green!30}.773$_{.740-.803}$ & .821$_{.791-.848}$ \\
\textcolor{localrubric}{Local-Rubric-Basic} & .679$_{.654-.702}$ & .812$_{.792-.831}$ & .683$_{.652-.715}$ & .795$_{.764-.824}$ \\
\textcolor{localrubric}{Local-Rubric-NoInterp} & .735$_{.712-.758}$ & .830$_{.812-.846}$ & .724$_{.691-.755}$ & .827$_{.803-.852}$ \\
\textcolor{globalrubric}{Global-Rubric-Blind} & .660$_{.635-.685}$ & .806$_{.787-.826}$ & .643$_{.608-.676}$ & .798$_{.767-.826}$ \\
\textcolor{globalrubric}{Global-Rubric} & .681$_{.656-.706}$ & .825$_{.805-.844}$ & .651$_{.618-.684}$ & .818$_{.791-.846}$ \\
\textcolor{globalrubric}{Global-Rubric-Auto} & .760$_{.738-.780}$ & .824$_{.806-.842}$ & .764$_{.735-.790}$ & .812$_{.784-.838}$ \\
\textcolor{globalrubric}{Global-Rubric-Tabular} & .635$_{.610-.659}$ & \cellcolor{green!30}.840$_{.823-.857}$ & .611$_{.581-.642}$ & \cellcolor{green!30}.828$_{.800-.855}$ \\
\bottomrule
\end{tabular}

\end{table}

% --- Per-task: lab_hypoglycemia ---
\providecolor{localrubric}{RGB}{190, 100, 25}
\providecolor{globalrubric}{RGB}{45, 90, 165}
\begin{table}[htbp]
\centering
\small
\renewcommand{\arraystretch}{0.85}
\caption{Hypoglycemia. AUROC and AUPRC with 95\% bootstrap CI. Best per column for each setting is highlighted. Embeddings are from Qwen3-8B; sample sizes refer to per-class training examples.}
\label{tab:embedding_task_labhypoglycemia}

\begin{tabular}{lcccc}
\toprule
 & \multicolumn{2}{c}{\textbf{AUROC}} & \multicolumn{2}{c}{\textbf{AUPRC}} \\
\cmidrule(lr){2-3} \cmidrule(lr){4-5}
Qwen3-8B-Zeroshot & \multicolumn{2}{c}{.687$_{.665-.710}$} & \multicolumn{2}{c}{.662$_{.629-.693}$} \\
GPT5-Mini-Zeroshot & \multicolumn{2}{c}{.662$_{.636-.687}$} & \multicolumn{2}{c}{.650$_{.617-.683}$} \\
\midrule
 & $n=10$ & $n=\text{All}$ & $n=10$ & $n=\text{All}$ \\
\cmidrule(lr){2-2} \cmidrule(lr){3-3} \cmidrule(lr){4-4} \cmidrule(lr){5-5}
Count-GBM & .571$_{.543-.598}$ & .591$_{.563-.621}$ & .561$_{.526-.597}$ & .596$_{.557-.634}$ \\
CLMBR-T & .611$_{.584-.639}$ & .777$_{.755-.799}$ & .595$_{.560-.631}$ & .764$_{.730-.797}$ \\
NaiveText & .587$_{.559-.616}$ & .680$_{.654-.706}$ & .584$_{.546-.621}$ & .675$_{.638-.709}$ \\
\textcolor{localrubric}{Local-Rubric} & .670$_{.644-.698}$ & .780$_{.757-.803}$ & .687$_{.651-.723}$ & .767$_{.732-.798}$ \\
\textcolor{localrubric}{Local-Rubric-Basic} & .642$_{.615-.668}$ & .759$_{.736-.782}$ & .630$_{.593-.667}$ & .745$_{.710-.781}$ \\
\textcolor{localrubric}{Local-Rubric-NoInterp} & \cellcolor{green!30}.709$_{.683-.734}$ & .775$_{.753-.797}$ & .703$_{.668-.737}$ & .767$_{.736-.798}$ \\
\textcolor{globalrubric}{Global-Rubric-Blind} & .685$_{.656-.713}$ & .769$_{.746-.795}$ & \cellcolor{green!30}.707$_{.671-.741}$ & .750$_{.713-.784}$ \\
\textcolor{globalrubric}{Global-Rubric} & .592$_{.565-.621}$ & \cellcolor{green!30}.794$_{.774-.816}$ & .596$_{.557-.632}$ & \cellcolor{green!30}.783$_{.750-.814}$ \\
\textcolor{globalrubric}{Global-Rubric-Auto} & .586$_{.558-.614}$ & .727$_{.703-.752}$ & .563$_{.525-.601}$ & .733$_{.697-.763}$ \\
\textcolor{globalrubric}{Global-Rubric-Tabular} & .641$_{.613-.669}$ & .754$_{.729-.776}$ & .603$_{.568-.638}$ & .741$_{.705-.775}$ \\
\bottomrule
\end{tabular}

\end{table}

% --- Per-task: lab_hyponatremia ---
\providecolor{localrubric}{RGB}{190, 100, 25}
\providecolor{globalrubric}{RGB}{45, 90, 165}
\begin{table}[htbp]
\centering
\small
\renewcommand{\arraystretch}{0.85}
\caption{Hyponatremia. AUROC and AUPRC with 95\% bootstrap CI. Best per column for each setting is highlighted. Embeddings are from Qwen3-8B; sample sizes refer to per-class training examples.}
\label{tab:embedding_task_labhyponatremia}

\begin{tabular}{lcccc}
\toprule
 & \multicolumn{2}{c}{\textbf{AUROC}} & \multicolumn{2}{c}{\textbf{AUPRC}} \\
\cmidrule(lr){2-3} \cmidrule(lr){4-5}
Qwen3-8B-Zeroshot & \multicolumn{2}{c}{.686$_{.666-.707}$} & \multicolumn{2}{c}{.655$_{.626-.683}$} \\
GPT5-Mini-Zeroshot & \multicolumn{2}{c}{.706$_{.685-.727}$} & \multicolumn{2}{c}{.657$_{.629-.687}$} \\
\midrule
 & $n=10$ & $n=\text{All}$ & $n=10$ & $n=\text{All}$ \\
\cmidrule(lr){2-2} \cmidrule(lr){3-3} \cmidrule(lr){4-4} \cmidrule(lr){5-5}
Count-GBM & .518$_{.491-.545}$ & .539$_{.514-.563}$ & .529$_{.497-.561}$ & .526$_{.495-.558}$ \\
CLMBR-T & .548$_{.522-.572}$ & .658$_{.633-.681}$ & .557$_{.523-.589}$ & .641$_{.608-.674}$ \\
NaiveText & .514$_{.489-.538}$ & .595$_{.571-.619}$ & .522$_{.490-.554}$ & .595$_{.563-.626}$ \\
\textcolor{localrubric}{Local-Rubric} & \cellcolor{green!30}.663$_{.638-.688}$ & .740$_{.718-.763}$ & \cellcolor{green!30}.619$_{.587-.650}$ & .703$_{.668-.737}$ \\
\textcolor{localrubric}{Local-Rubric-Basic} & .561$_{.538-.587}$ & .707$_{.684-.729}$ & .557$_{.526-.591}$ & .686$_{.655-.719}$ \\
\textcolor{localrubric}{Local-Rubric-NoInterp} & .549$_{.525-.575}$ & .745$_{.725-.767}$ & .523$_{.493-.553}$ & .725$_{.695-.758}$ \\
\textcolor{globalrubric}{Global-Rubric-Blind} & .559$_{.534-.586}$ & .706$_{.684-.729}$ & .568$_{.536-.600}$ & .690$_{.659-.722}$ \\
\textcolor{globalrubric}{Global-Rubric} & .574$_{.550-.599}$ & .719$_{.697-.740}$ & .564$_{.535-.596}$ & .703$_{.672-.735}$ \\
\textcolor{globalrubric}{Global-Rubric-Auto} & .608$_{.583-.633}$ & .733$_{.713-.756}$ & .599$_{.565-.632}$ & .720$_{.691-.751}$ \\
\textcolor{globalrubric}{Global-Rubric-Tabular} & .517$_{.492-.541}$ & \cellcolor{green!30}.776$_{.757-.797}$ & .533$_{.503-.564}$ & \cellcolor{green!30}.757$_{.730-.787}$ \\
\bottomrule
\end{tabular}

\end{table}

% --- Per-task: lab_thrombocytopenia ---
\providecolor{localrubric}{RGB}{190, 100, 25}
\providecolor{globalrubric}{RGB}{45, 90, 165}
\begin{table}[htbp]
\centering
\small
\renewcommand{\arraystretch}{0.85}
\caption{Thrombocytopenia. AUROC and AUPRC with 95\% bootstrap CI. Best per column for each setting is highlighted. Embeddings are from Qwen3-8B; sample sizes refer to per-class training examples.}
\label{tab:embedding_task_labthrombocytopenia}

\begin{tabular}{lcccc}
\toprule
 & \multicolumn{2}{c}{\textbf{AUROC}} & \multicolumn{2}{c}{\textbf{AUPRC}} \\
\cmidrule(lr){2-3} \cmidrule(lr){4-5}
Qwen3-8B-Zeroshot & \multicolumn{2}{c}{.705$_{.683-.725}$} & \multicolumn{2}{c}{.637$_{.608-.666}$} \\
GPT5-Mini-Zeroshot & \multicolumn{2}{c}{.739$_{.719-.759}$} & \multicolumn{2}{c}{.670$_{.645-.696}$} \\
\midrule
 & $n=10$ & $n=\text{All}$ & $n=10$ & $n=\text{All}$ \\
\cmidrule(lr){2-2} \cmidrule(lr){3-3} \cmidrule(lr){4-4} \cmidrule(lr){5-5}
Count-GBM & .498$_{.473-.525}$ & .556$_{.531-.580}$ & .507$_{.477-.540}$ & .551$_{.519-.582}$ \\
CLMBR-T & .522$_{.496-.548}$ & .627$_{.604-.654}$ & .523$_{.492-.554}$ & .606$_{.575-.640}$ \\
NaiveText & .518$_{.491-.545}$ & .612$_{.586-.635}$ & .524$_{.491-.555}$ & .588$_{.556-.622}$ \\
\textcolor{localrubric}{Local-Rubric} & \cellcolor{green!30}.771$_{.749-.791}$ & .841$_{.824-.859}$ & \cellcolor{green!30}.732$_{.701-.765}$ & .810$_{.781-.836}$ \\
\textcolor{localrubric}{Local-Rubric-Basic} & .608$_{.584-.632}$ & .760$_{.739-.781}$ & .572$_{.541-.606}$ & .742$_{.711-.770}$ \\
\textcolor{localrubric}{Local-Rubric-NoInterp} & .718$_{.696-.739}$ & .815$_{.796-.834}$ & .673$_{.638-.705}$ & .769$_{.736-.800}$ \\
\textcolor{globalrubric}{Global-Rubric-Blind} & .611$_{.588-.634}$ & .828$_{.810-.847}$ & .583$_{.552-.617}$ & .804$_{.776-.830}$ \\
\textcolor{globalrubric}{Global-Rubric} & .625$_{.601-.648}$ & .851$_{.834-.868}$ & .585$_{.554-.617}$ & .834$_{.807-.859}$ \\
\textcolor{globalrubric}{Global-Rubric-Auto} & .616$_{.592-.640}$ & .828$_{.809-.846}$ & .593$_{.563-.625}$ & .800$_{.770-.829}$ \\
\textcolor{globalrubric}{Global-Rubric-Tabular} & .656$_{.632-.679}$ & \cellcolor{green!30}.860$_{.844-.875}$ & .613$_{.584-.640}$ & \cellcolor{green!30}.841$_{.818-.866}$ \\
\bottomrule
\end{tabular}

\end{table}

% --- Per-task: chexpert ---
\providecolor{localrubric}{RGB}{190, 100, 25}
\providecolor{globalrubric}{RGB}{45, 90, 165}
\begin{table}[htbp]
\centering
\small
\renewcommand{\arraystretch}{0.85}
\caption{Chest X-ray. AUROC and AUPRC with 95\% bootstrap CI. Best per column for each setting is highlighted. Embeddings are from Qwen3-8B; sample sizes refer to per-class training examples.}
\label{tab:embedding_task_chexpert}

\begin{tabular}{lcccc}
\toprule
 & \multicolumn{2}{c}{\textbf{AUROC}} & \multicolumn{2}{c}{\textbf{AUPRC}} \\
\cmidrule(lr){2-3} \cmidrule(lr){4-5}
Qwen3-8B-Zeroshot & \multicolumn{2}{c}{.546$_{.519-.573}$} & \multicolumn{2}{c}{.526$_{.491-.560}$} \\
GPT5-Mini-Zeroshot & \multicolumn{2}{c}{.520$_{.498-.541}$} & \multicolumn{2}{c}{.507$_{.476-.538}$} \\
\midrule
 & $n=10$ & $n=\text{All}$ & $n=10$ & $n=\text{All}$ \\
\cmidrule(lr){2-2} \cmidrule(lr){3-3} \cmidrule(lr){4-4} \cmidrule(lr){5-5}
Count-GBM & \cellcolor{green!30}.571$_{.541-.602}$ & .609$_{.576-.642}$ & .543$_{.502-.583}$ & .582$_{.540-.624}$ \\
CLMBR-T & .453$_{.421-.484}$ & .609$_{.578-.640}$ & .465$_{.427-.500}$ & .600$_{.559-.643}$ \\
NaiveText & .558$_{.533-.584}$ & \cellcolor{green!30}.616$_{.592-.640}$ & \cellcolor{green!30}.551$_{.519-.583}$ & \cellcolor{green!30}.609$_{.577-.641}$ \\
\textcolor{localrubric}{Local-Rubric} & .517$_{.492-.543}$ & .606$_{.583-.630}$ & .503$_{.474-.535}$ & .605$_{.571-.638}$ \\
\textcolor{localrubric}{Local-Rubric-Basic} & .487$_{.457-.519}$ & .599$_{.568-.628}$ & .493$_{.454-.531}$ & .594$_{.554-.633}$ \\
\textcolor{localrubric}{Local-Rubric-NoInterp} & .525$_{.491-.556}$ & .596$_{.565-.625}$ & .509$_{.469-.546}$ & .581$_{.537-.625}$ \\
\textcolor{globalrubric}{Global-Rubric-Blind} & .437$_{.413-.461}$ & .585$_{.559-.608}$ & .455$_{.427-.486}$ & .587$_{.556-.620}$ \\
\textcolor{globalrubric}{Global-Rubric} & .513$_{.489-.538}$ & .594$_{.567-.619}$ & .521$_{.489-.552}$ & .582$_{.550-.614}$ \\
\textcolor{globalrubric}{Global-Rubric-Auto} & .503$_{.470-.534}$ & .575$_{.543-.605}$ & .497$_{.460-.534}$ & .557$_{.518-.596}$ \\
\textcolor{globalrubric}{Global-Rubric-Tabular} & .500$_{.500-.500}$ & .540$_{.507-.572}$ & .497$_{.470-.523}$ & .527$_{.486-.567}$ \\
\bottomrule
\end{tabular}

\end{table}

\clearpage
\section{Full prompts used in rubric representation learning methods} \label{app:agent_prompts}
\subsection{Prompt template for computing text-embeddings of inputs}
\label{app:task_prompt}

\definecolor{GreenBox}{HTML}{00a087}
\colorlet{GreenBoxBg}{GreenBox!6}

\definecolor{BlueBox}{HTML}{4dbbd5}
\definecolor{RedBox}{HTML}{e64b35}
\definecolor{GreenBox}{HTML}{00a087}
\definecolor{OrBox}{HTML}{FF8C00}
\definecolor{Bl2Box}{HTML}{3C5488}
\definecolor{BrBox}{HTML}{B09C85}

\colorlet{BlueBoxBg}{BlueBox!6}
\colorlet{RedBoxBg}{RedBox!6}
\colorlet{GreenBoxBg}{GreenBox!6}
\colorlet{OrBoxBg}{OrBox!6}
\colorlet{Bl2BoxBg}{Bl2Box!6}
\colorlet{BrBoxBg}{BrBox!6}

\begin{figure*}[ht]
\centering
    \begin{tcolorbox}[
        colback=gray!6,
        colframe=gray!70,
        boxrule=1pt,
        arc=5pt,
        width=0.97\linewidth,
        top=8pt, bottom=8pt, left=8pt, right=8pt
    ]
    \ttfamily\scriptsize
    \vspace{-0.6em}
    \textbf{\# Prompt template for obtaining text input embeddings for downstream training}\\
    \vspace{-0.5em}
    \hrule
    \vspace{0.5em}
    
        Based on the patient's EHR below, predict: \{\textcolor{red}{task\_query}\}\\
        
        --- Patient EHR --- \\
        \{\textcolor{red}{$x^{\text{text}}$ or $x^{\text{rubric}}$}\}\\
        --- End of EHR ---\\
        
        Based on the above EHR, predict: \{\textcolor{red}{task\_query}\}\\
        Respond with exactly one word: Yes or No.
    \end{tcolorbox}

    \caption{Prompt for converting textual inputs to embeddings. An example task query: ``Will the patient develop lupus within next year?''}
    \label{fig:task_prompt}
    \vspace{-10pt}
\end{figure*}

For every textual representation used as input to the downstream logistic regression classifier (\textcolor{methodblue}{NaiveText} and all textual rubric variants), the textual input ($x^{\text{text}}$ or $x^{\text{rubric}}$) is first wrapped in the task-conditioned template shown in \Cref{fig:task_prompt} before being passed to the frozen text-embedding model. The template prepends and appends a brief task query so that the resulting embedding is conditioned on the prediction target rather than reflecting only generic input content.
\clearpage
\subsection{Prompt for global rubric creation (\Cref{fig:rubric_creation_pipeline}, Panel B)}
\label{app:rubric_prompt}
\definecolor{BlueBox}{HTML}{4dbbd5}
\colorlet{BlueBoxBg}{BlueBox!6}

\begin{figure}[ht]
\begin{tcolorbox}[
    colback=BlueBoxBg,
    colframe=BlueBox,
    coltext=black,
    boxrule=1pt,
    arc=5pt,
    width=\linewidth,
]
\ttfamily\small
\textbf{\# Prompt used with GPT-5-mini for global rubric synthesis}\\
\hrule
\vspace{0.5em}
You are a medical expert designing a structured rubric for a clinical prediction task. \\

\#\# Task \\
- Name: \{\textcolor{red}{task\_name}\}\\
- Query: \{\textcolor{red}{task\_query}\} \\

\#\# Context \\
You will be given \{\textcolor{red}{40}\} labeled patient EHR examples (\{\textcolor{red}{20}\} positive, \{\textcolor{red}{20}\} negative). Another model will later use your rubric to transform new patient EHRs into structured summaries, which will then serve as input to a supervised classifier. \\

\#\# What You Must Do \\
Study the examples below. Combine what you observe in them with your medical knowledge to design a rubric template -- a set of named fields that, when filled in for any patient, produce a structured summary optimized for this prediction task. \\

The rubric should: \\
1. **Be data-driven and discriminative.** Identify which features, patterns, and interactions actually separate the positive and negative cases. The rubric should capture not just obvious indicators but also subtler or compound features you notice. At the same time, do not overfit to these 40 cases -- use your clinical knowledge to include factors that are generally relevant even if not prominent in this sample.\\

2. **Be structured and consistent.** Every rubricified output must follow the same field names and order. For each field, specify what to extract from the EHR and how to format it. Specify what to write when data is absent. \\ 

3. **Extract facts only.** The evaluator filling in the rubric must extract and organize information from the EHR. It must NOT make predictions, assign risk levels, or draw conclusions.\\

4. **Be concise.** The rubric should focus on extracting information that is relevant to the task. It should not ask the evaluator to reproduce the entire EHR. \\

\#\# Positive Examples (Ground Truth: Yes)

\{\textcolor{red}{NaiveText EHR serializations of 20 positive examples concatenated ($x_{\textnormal{text}}$ format)}\} \\

\#\# Negative Examples (Ground Truth: No)

\{\textcolor{red}{NaiveText EHR serializations of 20 negative examples concatenated ($x_{\textnormal{text}}$ format)}\} \\

\#\# Output

Output ONLY the rubric template itself -- the instructions another model will follow to transform a patient EHR. No preamble, no explanation of your reasoning. The template must be self-contained and directly usable.

\end{tcolorbox}
\caption{Prompt used with GPT-5-mini to guide global rubric creation from NaiveText serializations ($x_{\textnormal{text}}$) of EHR examples.}
\label{fig:rubric_prompt}
\vspace{-10pt}
\end{figure}

\clearpage
\subsection{Prompt for global rubric application (\Cref{fig:rubric_creation_pipeline}, Panel D)}
\label{app:rubric_application_prompt}
\definecolor{BlueBox}{HTML}{4dbbd5}
\colorlet{BlueBoxBg}{BlueBox!6}

\begin{figure}[ht]
\begin{tcolorbox}[
    colback=BlueBoxBg,
    colframe=BlueBox,
    coltext=black,
    boxrule=1pt,
    arc=5pt,
    width=\linewidth,
]
\ttfamily\small
\textbf{\# Prompt used with GPT-5-mini for global rubric application}\\
\hrule
\vspace{0.5em}

You are a medical data extraction specialist. \\

\#\# Task \\
\{\textcolor{red}{task\_query}\} \\

\#\# Rubric Template (follow this exactly) \\
\{\textcolor{red}{rubric\_instructions}\} \\

\#\# Patient EHR \\
\{\textcolor{red}{ehr\_text ($x_{\textnormal{text}}$ format)}\} \\

\#\# Instructions \\
Fill in every field of the rubric template above using ONLY information from this patient's EHR. \\

Rules:\\
- Follow the exact field order and section structure of the rubric.\\
- Be concise: use short phrases, numbers, and dates. Do not write paragraphs.\\
- If data for a field is not present in the EHR, write ``No data''.\\
- Do NOT add commentary, predictions, risk assessments, or conclusions.\\
- Do NOT include any information not found in the EHR above.\\

Rubric output:

\end{tcolorbox}
\caption{Prompt used with GPT-5-mini for transforming a naive text serialized input ($x^{\text{text}}$) into its rubric text serialization version ($x^{\text{rubric}}$).}
\label{fig:rubric_application_prompt}
\vspace{-10pt}
\end{figure}

\clearpage
\subsection{Prompt for creating a global rubric application parser (\Cref{fig:rubric_creation_pipeline}, Panel E)}
\label{app:rubric_parser_prompt}
\definecolor{BlueBox}{HTML}{4dbbd5}
\colorlet{BlueBoxBg}{BlueBox!6}

\begin{figure}[ht]
\begin{tcolorbox}[
    colback=BlueBoxBg,
    colframe=BlueBox,
    coltext=black,
    boxrule=1pt,
    arc=5pt,
    width=\linewidth,
]
\ttfamily\scriptsize
\textbf{\# Prompt used with GPT-5.2 for generating a parser script for rubric application}\\
\hrule
\vspace{0.5em}
You are an expert Python developer and medical informaticist. \\

\#\# Your Task \\
Write a complete, self-contained Python script that reads patient EHR serializations and fills in a structured clinical rubric template using **deterministic string/regex parsing only** --- no LLM API calls, no network requests.\\

\#\# Clinical Task Context\\
- Task name: \{\textcolor{red}{task\_name}\}\\
- Prediction query: \{\textcolor{red}{task\_query}\}\\

\#\# Rubric Template to Fill\\
The script must fill in every field defined in the following rubric instructions: \{\textcolor{red}{rubric\_instructions, ${\cal R}$}\} \\

\#\# EHR Serialization Format\\
Below are 40 example patient EHR serializations from the training cohort, labeled by ground-truth outcome.
For each patient you are shown BOTH:\\
  1. The raw naive text EHR serialization.\\
  2. The LLM-produced rubric fill for that exact patient --- showing you how the fields should be extracted from the raw text.\\

Use these paired examples to understand the extraction mapping precisely.\\
\{\textcolor{red}{40 paired examples of naive text serializations ($x^{\text{text}}$) and LLM-filled rubric text serializations ($x^{\text{rubric}}$).}\} \\

\#\# Required Script Interface\\
The generated script must:\\
1. Accept the following command-line arguments via argparse:\\
   - `--input\_dir`  : root directory of naivetext serializations 
   - `--output\_dir` : root directory for llmrubric-parser outputs 
   - `--task`       : task name 
   - `--splits`     : one or more of `train val test` \\

2. For each split, read `\{\{input\_dir\}\}/\{\{task\}\}/\{\{split\}\}.json` --- a JSON array where each element has:\\
   - `patient\_id` (int)
   - `prediction\_time` (ISO datetime string)
   - `task` (str)
   - `split` (str)
   - `label` (bool)
   - `serialization` (str) ← the EHR text to parse\\

3. For each patient call `fill\_rubric(serialization: str) -> str`, which:\\
   - Extracts all rubric fields from the EHR text using regex and string operations
   - Returns a filled-in rubric string that follows the exact field names, order, and format from the rubric template above
   - Writes "NA" for any field whose data is absent from the EHR\\

4. Write output to `\{\{output\_dir\}\}/\{\{task\}\}/\{\{split\}\}.json` --- a JSON array where each element has:\\
   - `patient\_id` (int)
   - `prediction\_time` (str)
   - `task` (str)
   - `split` (str)
   - `label` (bool)
   - `rubricified\_text` (str) ← output of fill\_rubric()\\

5. Create output directories as needed (parents=True, exist\_ok=True).\\

6. Print progress to stdout: total patients processed per split.\\

\#\# Constraints\\
- Use only Python standard library plus `re`, `json`, `argparse`, `pathlib`, `sys`. No third-party packages.
- No LLM API calls, network requests, external tools.
- The `fill\_rubric` function must be deterministic and handle missing data gracefully (write "NA" rather than raising exceptions).
- The script must be syntactically valid Python 3.8+.
- Do NOT hardcode file paths --- use the argparse arguments.\\

\#\# Output\\
Output ONLY the Python script, with no explanation, no preamble, and no markdown fences. The output must start with `\#!/usr/bin/env python3` and be directly writable to a .py file.

\end{tcolorbox}
\caption{Prompt used with GPT-5.2 to create a parser script for transforming a naive text serialized input ($x^{\text{text}}$) into its rubric text serialization version ($x^{\text{rubric}}$).}
\label{fig:rubric_parser_prompt}
\vspace{-10pt}
\end{figure}

\clearpage
\subsection{Prompt for creating a global rubric tabularization parser (\Cref{fig:rubric_creation_pipeline}, Panel F)}
\label{app:rubric_tabularization_prompt}
\definecolor{BlueBox}{HTML}{4dbbd5}
\colorlet{BlueBoxBg}{BlueBox!6}

\begin{center}
\begin{tcolorbox}[
    enhanced,
    breakable,
    colback=BlueBoxBg,
    colframe=BlueBox,
    boxrule=1pt,
    arc=5pt,
    width=0.95\textwidth,
    top=8pt, bottom=8pt, left=8pt, right=8pt
]
\ttfamily\scriptsize

\textbf{\# Prompt used with GPT-5.2 for generating a parser script to transform rubric serializations to tabular features}\\
\hrule
\vspace{0.5em}
You are an expert Python developer and medical informaticist.\\

\#\# Your Task \\
Write a complete, self-contained Python featurizer script that reads rubric-formatted patient EHR texts and converts each one into a **fixed-dimension numeric feature vector** using deterministic string/regex parsing — no LLM calls, no network requests.\\

\#\# Clinical Task Context\\
- Task name: \{\textcolor{red}{task\_name}\}\\
- Prediction query: \{\textcolor{red}{task\_query}\}\\

\#\# Rubric Parser Source (shows all rubric field names and their text formats)\\
The following is the parser that generates the rubric text. Study it to understand which fields exist and how their values are formatted in the text. This is the **ground truth** for what fields can appear in a rubric text and how their values are formatted.\\

```python \\
\{\textcolor{red}{task-specific rubric parser generated via prompt in \Cref{app:rubric_parser_prompt}}\} \\
```\\

\#\# Reference Rubric Texts (\{\textcolor{red}{20}\} positive, \{\textcolor{red}{20}\} negative) \\

**Important context:** These \{\textcolor{red}{40}\} patients are the cohort that was used to *design* the rubric itself. They are provided as examples so you can calibrate your regex patterns against actual data.\\

**However**, the featurizer you write will be applied to a **much larger dataset** (thousands of patients). Your feature extraction logic must therefore be:\\

- **General**: handle any value the rubric parser could plausibly produce, not just the values seen in these 40 patients\\

- **Robust**: gracefully handle missing, NA, or unexpected values for every field\\

- **Comprehensive**: derive features from every field in the rubric, even if that field happens to be NA for all \{\textcolor{red}{40}\} examples shown here\\

Use the parser source above as the authoritative specification of fields and value formats; use the examples below to validate and calibrate your regex patterns.\\

\{\textcolor{red}{40 example text serialization in $x^{\text{rubric}}$ format}\} \\

\#\# Required Script Interface\\ 

The generated script must:\\
1. Accept CLI arguments via argparse:
   - `--input\_dir`  \\
   - `--output\_dir` \\ 
   - `--task`        \\ 
   - `--splits`      \\ 

2. For each split, read `\{\{input\_dir\}\}/\{\{split\}\}/\{\{task\}\}.json` — a JSON array where each element has: \\
   - `patient\_id` (int)\\
   - `label\_time` (ISO datetime string)\\
   - `label\_value` (bool)\\
   - `conversations` (list) — rubric text is in `conversations[1]["content"]` between `--- Patient EHR ---` and `--- End of EHR ---`\\

3. Implement `def extract\_features(rubric\_text: str) -> dict[str, float]`:\\
   - Parse every rubric field from the text\\
   - Return a flat dict mapping feature name → float value\\
   - For **numeric fields**: extract the number; if missing/NA write `0.0` and set `\{\{field\}\}\_missing = 1.0`\\
   - For **categorical / Yes/No fields**: one-hot encode all known values; unknown/NA → all zeros plus a `\{\{field\}\}\_missing = 1.0` indicator\\
   - All returned values must be float (0.0 or 1.0 for binary, numeric otherwise)\\
   - The dict must have the **same keys in the same order** for every call (fixed schema)\\

4. Define `SCHEMA: list[dict]` at module level — one entry per feature with keys:\\
   - `"name"`: feature name (matches key in extract\_features output)\\
   - `"type"`: `"numeric"`, `"binary"`, or `"categorical"`\\
   - `"description"`: short human-readable description\\
   - `"possible\_values"`: list of string values for categorical/binary fields, omit for numeric\\

5. For each split, build an N×F float32 matrix from `extract\_features`, save as:\\
   - `\{\{output\_dir\}\}/\{\{task\}\}/\{\{split\}\}.npz` with numpy keys:\\
     - `embeddings`: shape (N, F) float32\\
     - `labels`: shape (N,) int32\\
     - `patient\_ids`: shape (N,) int64\\
     - `prediction\_times`: shape (N,) object (strings)\\

6. Save `\{\{output\_dir\}\}/\{\{task\}\}/feature\_schema.json` once (after processing the first split):\\
   ```json\\
   \{\{
     "task": "\{task\}",\\
     "task\_query": "\{task\_query\}",\\
     "num\_features": <F>,\\
     "features": <SCHEMA list>\\
   \}\}\\
   ```\\

7. Create output directories as needed. Print progress to stdout.\\

\#\# Constraints\\
- Use only Python standard library plus `re`, `json`, `numpy`, `argparse`, `pathlib`, `sys`. No third-party packages beyond numpy.\\
- No LLM API calls, no network requests.\\
- `extract\_features` must be deterministic and never raise exceptions on any input (catch all errors, default to 0.0).\\
- The script must be syntactically valid Python 3.8+.\\
- Do NOT hardcode file paths — use the argparse arguments.\\
- Aim for **at least 30 features** to capture the richness of the rubric. \\Include all numeric fields, all categorical fields (one-hot), and Yes/No procedure/comorbidity flags.\\

\#\# Output\\
Output ONLY the Python script, with no explanation, no preamble, and no markdown fences. Start with `\#!/usr/bin/env python3`.

\end{tcolorbox}

\captionof{figure}{Prompt used with GPT-5.2 to create a parser script that transforms rubric-transformed inputs ($x^{\text{rubric}}$) into a fixed-dimensional tabular feature vector.}
\label{fig:rubric_tabularization_prompt}
\end{center}

\clearpage
\subsection{Prompt for stripping away interpretation of evidence from a local rubric representation (\Cref{fig:sum_rub_creation_prompts}, Left)}
\label{app:local-interp}
\definecolor{BlueBox}{HTML}{4dbbd5}
\colorlet{BlueBoxBg}{BlueBox!6}

\begin{figure}[ht]
\begin{tcolorbox}[
    colback=BlueBoxBg,
    colframe=BlueBox,
    coltext=black,
    boxrule=1pt,
    arc=5pt,
    width=\linewidth,
]
\ttfamily\small
\textbf{\# Prompt used with GPT-5-mini for stripping away interpretive language from Local-Rubrics}\\
\hrule
\vspace{0.5em}

Below is a clinical summary of a patient's EHR. Your task is to produce a MINIMALLY EDITED version that removes ONLY interpretive and predictive language.

\vspace{1.5em}
--- START OF EHR DATA ---\par
\{\textcolor{red}{NaiveText\_Serialization ($x^{\text{text}}$)}\}\par
--- END OF EHR DATA ---

\vspace{1.5em}
REMOVE (by deleting or replacing with neutral phrasing):\\

- Causal/risk language ("increases risk", "raises concern", "associated with worse outcomes") \\
- Protective framing ("protective", "favorable", "reassuring", "reduces risk") \\
- Interpretive connectives ("suggests", "indicates", "consistent with", "likely", "concerning for") \\
- Weighing/aggregating/conclusion statements that assess overall risk or probability \\
- Predictive statements or probability assessments

\vspace{1.5em}
KEEP UNCHANGED as much as possible: \\

- The overall structure, section ordering, and formatting (including think tags and \#\#\# headers) \\
- All factual clinical content: demographics, diagnoses, lab values with numbers, vitals, medications, procedures, dates, and other concrete data points \\
- The exact wording of factual statements — do NOT rephrase facts that contain no interpretation \\
- Section headers — keep them but strip interpretive framing from bullet points beneath them \\
- Bullet point structure and ordering

\vspace{1.5em}
IMPORTANT:\\

- Make the FEWEST changes necessary. If a bullet point is purely factual, leave it verbatim. \\
- If a bullet mixes fact + interpretation, keep the fact and delete only the interpretive clause. Example: "Hyperglycemia (glucose 202 mg/dL) indicating diabetes or stress hyperglycemia (associated with worse outcomes)" -> "Hyperglycemia (glucose 202 mg/dL)." \\
- If a bullet is purely interpretive with no factual content, delete it entirely. \\
- Do NOT add any new information, reorganize sections, or change the summary's structure. \\
- The section "WEIGHING AND AGGREGATING THE EVIDENCE" (section 5) should be ENTIRELY removed — delete the header and all its content. It is purely interpretive.
\end{tcolorbox}
\caption{Prompt used with GPT-5-mini for stripping away interpretive language from Local-Rubric representation ({\em i.e.,} modifying $x^{\text{rubric}}$).}
\label{fig:local-interp}
\vspace{-10pt}
\end{figure}

\clearpage
\section{Full global rubric examples} \label{app:fullrubrics}
\subsection{Full global rubric for the hypertension onset prediction task}
\label{app:htn_rubric}

\definecolor{OrBox}{HTML}{FF8C00}
\colorlet{OrBoxBg}{OrBox!6}

\begin{center}
\begin{tcolorbox}[
    enhanced,
    breakable,
    colback=OrBoxBg,
    colframe=OrBox,
    boxrule=1pt,
    arc=5pt,
    width=0.95\textwidth,
    top=8pt, bottom=8pt, left=8pt, right=8pt
]
\ttfamily\scriptsize

\textbf{RUBRIC INSTRUCTIONS FOR TASK: HYPERTENSION}\\
\hrule
\vspace{0.5em}

\textbf{Rubric purpose}\\
- Provide a reproducible, stepwise process to transform any EHR into a structured, clinical-evidence summary useful for assessing the likelihood that a patient will develop hypertension in the next year.\\
- The rubric standardizes what to extract, how to summarize trends and risk factors, and how to record uncertainty and provenance so downstream models or clinicians can apply consistent reasoning.\\

\vspace{0.5em}
\textbf{How to use this rubric}\\
- Follow the numbered extraction and analysis steps for each new patient.\\
- Populate the structured template fields exactly (use units shown). If data are missing, enter ``missing'' and note time windows attempted.\\
- Do NOT make a final yes/no prediction inside the form. Instead, produce the structured summary and quantitative or qualitative risk-domain scores for downstream modeling.\\

\vspace{0.5em}
\textbf{A. Preparation (before extracting)}\\
1. Define the prediction window: ``next year'' relative to the EHR reference date and time.\\
2. Define time windows to extract:\\
\quad - Very recent: last 30 days\\
\quad - Recent: 31--180 days\\
\quad - Baseline/remote: $>$180 days up to available history\\
3. Standardize units and formats:\\
\quad - Blood pressure: mmHg (systolic/diastolic)\\
\quad - Weight: kg or oz $\rightarrow$ convert to kg if numeric calculations needed\\
\quad - Height: cm or in $\rightarrow$ convert to meters\\
\quad - Labs: use usual clinical units (creatinine mg/dL, A1c \%, etc.)\\
4. Log data sources (vitals, problem list, medications, laboratory, procedures, notes) and timestamp of extraction.\\

\vspace{0.5em}
\textbf{B. Step-by-step extraction \& transformation procedure}\\

\textbf{Step 1 --- Demographics and baseline context}\\
- Extract:\\
\quad - Age (years)\\
\quad - Sex / gender\\
\quad - Race / ethnicity (if available)\\
\quad - Relevant social history: tobacco (current/former/never), alcohol (heavy/regular/rare/none), illicit drug use, tobacco product types\\
\quad - Pregnancy status (current or past complications such as pre-eclampsia)\\
\quad - Baseline height and weight; calculate BMI (kg/m$^2$) and BMI category\\
- Record date of last update for each demographic item.\\

\textbf{Step 2 --- Blood pressure (BP) data extraction and normalization}\\
- Extract all systolic and diastolic BP values with timestamps and context (office, inpatient, ED, home, ambulatory, perioperative).\\
- Normalize values: remove implausible readings (document them) and ensure mmHg units.\\
- For each time window:\\
\quad - Compute count, mean, median, standard deviation, minimum, and maximum.\\
\quad - Identify last available BP and date.\\
\quad - Flag highest recent systolic and diastolic values with dates.\\
- Compute trend metrics:\\
\quad - Recent slope = (mean\_recent $-$ mean\_baseline) / time (mmHg per month); indicate direction only if clinically meaningful (e.g., $\geq$3 mmHg/year).\\
\quad - BP variability indicator: SD of systolic BP in recent window; flag high variability if SD $>$10 mmHg.\\
- Categorize BP using ACC/AHA thresholds:\\
\quad - Normal ($<$120/$<$80)\\
\quad - Elevated (120--129/$<$80)\\
\quad - Stage 1 Hypertension (130--139 or 80--89)\\
\quad - Stage 2 Hypertension ($\geq$140 or $\geq$90)\\
\quad - If mixed, note ``discordant'' and list counts per category.\\

\textbf{Step 3 --- Antihypertensive and BP-impacting medications}\\
- Extract current and recent medications with start and stop dates if available.\\
- Flag antihypertensives (ACEi, ARBs, beta-blockers, diuretics, CCBs, vasodilators).\\
- Flag BP-raising agents (systemic corticosteroids, NSAIDs, decongestants, stimulants, calcineurin inhibitors, SNRIs, MAOIs, some oral contraceptives).\\
- For each flagged medication record name, dose, dates, indication, and temporal relation to BP changes.\\

\textbf{Step 4 --- Comorbidities associated with increased HTN risk}\\
- Extract diagnoses and ICD codes with dates:\\
\quad - Major risk: CKD, diabetes, CVD, PAD, OSA, endocrine causes, pregnancy or pre-eclampsia, obesity (BMI $\geq$30), heavy alcohol use.\\
\quad - Moderate risk: hyperlipidemia, metabolic syndrome, thyroid disease, autoimmune disease with renal involvement.\\
\quad - Secondary HTN clues: resistant BP, hypokalemia, episodic symptoms.\\
- Record first documentation date, last active date, and severity when available.\\

\textbf{Step 5 --- Relevant laboratory data}\\
- Extract labs with dates grouped by time window:\\
\quad - Creatinine / eGFR\\
\quad - Electrolytes (Na, K, HCO$_3$)\\
\quad - Glucose, HbA1c\\
\quad - Lipids\\
\quad - Urine albumin or protein\\
\quad - Thyroid tests\\
\quad - Aldosterone/renin, cortisol, catecholamines if available\\
- Flag abnormal values with interpretation (e.g., eGFR $<$60 ml/min).\\

\textbf{Step 6 --- Procedures and objective testing}\\
- Extract echocardiography, renal imaging, sleep studies, ABPM.\\
- Note evidence of end-organ effects (LVH, albuminuria, renal disease).\\

\textbf{Step 7 --- Social, behavioral, and family data}\\
- Smoking status and intensity.\\
- Alcohol use severity.\\
- Family history of HTN or early CVD.\\
- Adherence or socioeconomic barriers if documented.\\

\textbf{Step 8 --- Acute confounders}\\
- Identify acute illness, pain, surgery, sepsis, AKI, or inpatient context affecting BP interpretation.\\
- Avoid using isolated inpatient readings without outpatient corroboration.\\

\textbf{Step 9 --- Domain synthesis and scoring}\\
- For each domain, record evidence, recency, and confidence (High/Moderate/Low).\\
- Domains include BP phenotype, medications, metabolic risk, kidney function, secondary HTN, end-organ disease, behavior, and acute confounders.\\
- Assign severity (Major/Moderate/Minor) and create a domain scorecard; do NOT generate a final binary label.\\

\textbf{Step 10 --- Evidence provenance and missing data}\\
- Record source, timestamp, and confidence for each major item.\\
- Explicitly flag critical missing data (e.g., no outpatient BP in 12 months).\\

\textbf{Step 11 --- Structured output}\\
- Produce a standardized summary with demographics, BP summary, medications, comorbidities, labs, procedures, lifestyle, acute confounders, domain scorecard, missing data, and a 2--4 sentence neutral text summary.\\

\textbf{Step 12 --- Guidance notes}\\
- Recommend confirmatory testing or review where appropriate (e.g., home BP, med review, nephrology referral).\\
- Do not conclude final risk.\\

\vspace{0.5em}
\textbf{Final note to the user}\\
- Use this rubric to populate the structured template for every patient. Do not record a final hypertension risk classification here; the output is intended for downstream models or clinician judgment.\\

\end{tcolorbox}

\captionof{figure}{Global rubric instructions for extracting structured patient profiles for 1-year hypertension diagnosis prediction.}
\label{fig:htn-rubric}
\end{center}

\clearpage
\subsection{Full global rubric for the hyponatremia lab result prediction task}
\label{app:hyponatremia_rubric}

 \definecolor{OrBox}{HTML}{FF8C00}
\colorlet{OrBoxBg}{OrBox!6}

\begin{center}
\begin{tcolorbox}[
    enhanced,
    breakable,
    colback=OrBoxBg,
    colframe=OrBox,
    boxrule=1pt,
    arc=5pt,
    width=0.95\textwidth,
    top=8pt, bottom=8pt, left=8pt, right=8pt
]
\ttfamily\scriptsize

\textbf{RUBRIC INSTRUCTIONS FOR TASK: HYPONATREMIA LAB RESULT}\\
\hrule
\vspace{0.5em}

PredictionDate: [Extract the 'current time' / prediction timestamp from the EHR header].\\
Format: YYYY-MM-DD. If not present write NA.\\
Patient:\\ 

- Age: [years as integer from EHR]. If not present write NA.\\ - Sex: [as documented: MALE / FEMALE / Other / Unknown]. If not present write NA.\\ - Race/Ethnicity: [as documented]. If not present write NA.\\

ProblemListFlags (presence and dates):\\

- Chronic kidney disease / End-stage renal disease (CKD/ESRD): [Yes / No]. If Yes, list documented term(s) and most recent date(s) (YYYY-MM-DD). If none write No.\\ - Dialysis history/procedure in record: [Yes / No]. If Yes, list procedure name(s) and most recent date(s). If none write No.\\ - Prior documented hyponatremia / “hypo-osmolality and or hyponatremia”: [Yes / No]. If Yes, give the documentation text and date(s). If none write No.\\ - Active malignancy listed in Problem List or current visits: [Yes / No]. If Yes, list malignancy type(s) and most recent date(s). If none write No.\\ 

SerumSodium\_Last3 (most recent first): For up to 3 most recent serum/plasma/blood sodium measurements, extract a line per measurement in this exact format:\\

- YYYY-MM-DD (days\_before\_prediction): [value] mmol/L ; Specimen=[serum/plasma/blood] ; Setting=[ED/Inpatient/Outpatient/Lab] ; Note=[any explicit result comment if present]\\ If fewer than 3 measurements exist, include those available; if none write NA.\\ SerumSodium\_Min90:\\ - Lowest documented serum/plasma/blood sodium value in the prior 90 days (value mmol/L) and date (YYYY-MM-DD). If none write NA.\\ SerumOsmolality\_Last3:\\ - For up to 3 most recent serum osmolality measurements: YYYY-MM-DD (days\_before\_prediction): [value] mOsm/kg ; Setting=[as above]\\ If none write NA.\\ 

UrineStudies\_Last3:\\
- For up to 3 most recent urine study sets, extract for each available element on one line:\\   - YYYY-MM-DD (days\_before\_prediction): UrineNa=[value] mmol/L ; UrineOsm=[value] mOsm/kg ; SpecificGravity=[value] ; Setting=[ED/Inpatient/Outpatient/Lab]\\ Only include elements that are present for that date. If no urine studies documented write NA.\\ 

RenalFunction:\\
- Most recent serum creatinine (mg/dL) and date: YYYY-MM-DD: [value] mg/dL. If none write NA.\\ - Most recent BUN (mg/dL) and date: YYYY-MM-DD: [value] mg/dL. If none write NA.\\ - Recent acute renal failure / acute kidney injury entries within 30 days: [Yes / No]. If Yes include diagnosis text and date(s). If none write No.\\ 

VolumeRelatedFindings (documented in problem lists or visit notes within past 30 days):\\ 
- Extract presence with dates for these items (list each if present as "Item: YYYY-MM-DD;"): Edema, Ascites, Hypotension (documented low BP or explicit "hypotension"), Dehydration, Vomiting, Diarrhea, Nasogastric/feeding tube, Ileostomy/colostomy, Recent large-volume paracentesis. If none of these documented in past 30 days write NA.\\ 

Medications\_PotentiallyAffectingSodium (recent administrations — extract from medication list / inpatient meds / discharge meds):\\ 
- Time window: last 14 days before PredictionDate (if EHR supports more granular times use those). For each relevant med/class present include one line:\\   - [YYYY-MM-DD last administration if available] : [Medication name] ; Class=[thiazide/loop diuretic / SSRI / SNRI / TCA / anticonvulsant (carbamazepine/oxcarbazepine) / NSAID / SSRI, etc.] ; Route=[oral/IV] ; Dose if documented=[text]\\ - If none of these medication classes documented in last 14 days write NA.\\ - Also include "Chronic diuretic use noted (Yes/No) and last documentation date" (e.g., long-term thiazide).\\ 

IVFluids\_Last72h:\\ 
- List IV fluid administrations in last 72 hours (date/time if available) in the format:\\   - YYYY-MM-DD: [fluid type as documented, e.g., D5W / D5NS / 0.9\% NaCl / hypotonic saline / LR / "glucose 50 mg/mL prefills"] ; Volume if documented.\\ - If none documented write NA.\\ AcuteConditions\_AssociatedWithSIADHorHyponatremia (documented within 30 days):\\ - For each present within 30 days, list as "Condition: YYYY-MM-DD" from problem/visit notes:\\   - Pulmonary infection / pneumonia / pulmonary disease\\   - CNS disorder (stroke, hemorrhage, encephalopathy)\\   - Sepsis / severe infection\\   - Recent major surgery / Postoperative state\\   - Pain / Severe nausea (if explicitly documented)\\   - Malignancy active (if not already in ProblemListFlags)\\ If none documented write NA.\\ 

RecentProcedures\_Chemotherapy\_Transfusion (last 30 days):\\ 
- List any of: major surgery, chemotherapy, recent blood transfusion, paracentesis, TPN (total parenteral nutrition), plasmapheresis, hemodialysis — format:\\   - YYYY-MM-DD: [procedure name / chemo agent e.g., paclitaxel] ; Notes=[if available]\\ If none write NA.\\ Glucose\_Last3:\\ - Up to 3 most recent serum/plasma or point-of-care glucose values (most recent first):\\   - YYYY-MM-DD: [value] mg/dL ; Type=[serum/plasma/glucometer] ; Setting=[ED/Inpatient/Outpatient]\\ If none write NA.\\ 

SerumProteinOrLipidExtremes:\\ 
- If very high triglycerides or abnormal total protein/albumin documented close to sodium measurement, extract:\\   - YYYY-MM-DD: Triglycerides=[value] mg/dL ; TotalProtein=[value] g/dL ; Albumin=[value] g/dL\\ If none documented write NA.\\ 

PriorHyponatremiaHistory:\\ 
- Any historical low sodium episodes before 90 days (brief): list lowest prior value and date(s) or write NA.\\ 

LabQualityNotes:\\ 
- Any documented lab-quality flags on sodium measurement (e.g., hemolysis, lipemia, “evacuated blood collection tube” note, specimen issues): extract verbatim note and date(s). If none write NA.\\ 

RelevantVitalSigns\_NearMostRecentSodium:\\ 
- From the same encounter as the most recent sodium (if identifiable), extract: systolic/diastolic BP (mmHg), heart rate (bpm), and whether on oxygen or dialysis that encounter. Format:\\   - Date: YYYY-MM-DD ; SBP=[value] ; DBP=[value] ; HR=[value] ; Oxygen=[yes/no] with O2 sat if given ; DialysisThisEncounter=[yes/no]\\ If not available write NA.\\ 

FreeText\_Findings\_Cues:\\
- Extract any verbatim phrases (short quotes) that explicitly mention hyponatremia-related language in notes or problem list (e.g., "hyponatremia", "hypo-osmolality", "SIADH", "hypotonic fluids", "low sodium") with the date and the note type. Format:\\   - YYYY-MM-DD ; Source=[ProblemList/VisitNote/LabComment] ; Text="[exact phrase]"\\ If none write NA.\\ 

DataCompleteness:\\ 
- For each of the following categories indicate [Present / Absent / Not documented]: Serum sodium labs, urine sodium/osmolality, serum osmolality, recent meds list, dialysis record, IV fluids record, creatinine/BUN. Example: SerumSodium: Present ; UrineSodium: Absent ; etc.\\ 

ExtractionRules / Formatting Rules (must follow exactly):\\ 
- Always extract facts only; do not add interpretation, risk assessment, or predictions.\\ - Dates: use YYYY-MM-DD as in EHR; if EHR provides relative days include "(N days before prediction)" after date.\\ - When multiple values on same date, include all values separated by ";".\\ - If an item not found anywhere in the EHR, write exactly "NA".\\ - Keep each field on a single line (except the repeated-measure lists which may have up to three lines as specified).\\ - Use units exactly as specified (mmol/L for Na and UrineNa; mOsm/kg for osmolality; mg/dL for glucose/BUN/creatinine; mg/dL for triglycerides).\\ - Do not synthesize or infer ranges; extract only documented numeric values and verbatim text.\\

EndOfTemplate.
\end{tcolorbox}

\captionof{figure}{Global rubric instructions for extracting structured patient profiles for hyponatremia lab result prediction (abnormal vs. normal).}
\label{fig:hypo-rubric}
\end{center}

\end{document}